%% file: acl_latex.tex
\pdfoutput=1

\documentclass[11pt]{article}

\usepackage[preprint]{acl}
\usepackage{multicol}
\usepackage{multirow}
\usepackage{array}     
\usepackage{booktabs}  
\usepackage[most]{tcolorbox}
\usepackage{xcolor}
\usepackage{times}
\usepackage{latexsym}
\usepackage{booktabs}
\usepackage{amssymb}
\usepackage{bbm}
\usepackage[T1]{fontenc}

\usepackage[utf8]{inputenc}
\usepackage{amsmath}
\usepackage{subfig}
\usepackage{microtype}

\usepackage{inconsolata}

\usepackage{graphicx}

\NewDocumentCommand{\jiayu}
{ mO{} }{\textcolor{blue}{\textsuperscript{\textit{jiayu}}\textsf{\textbf{\small[#1]}}}}

\NewDocumentCommand{\dadi}
{ mO{} }{\textcolor{green}{\textsuperscript{\textit{dadi}}\textsf{\textbf{\small[#1]}}}}

\NewDocumentCommand{\zhiyuan}
{ mO{} }{\textsuperscript{\textbf{zhiyuan}}\textcolor{red}{\textsf{\textbf{\small[#1]}}}}

\NewDocumentCommand{\yi}
{ mO{} }{\textsuperscript{\textbf{yi}}\textcolor{red}{\textsf{\textbf{\small[#1]}}}}

\NewDocumentCommand{\zhitao}
{ mO{} }{\textcolor{gray}{\textsuperscript{\textit{zhitao}}\textsf{\textbf{\small[#1]}}}}

\NewDocumentCommand{\TODO}
{ mO{} }{\textcolor{red}{\textsuperscript{\textit{TODO}}\textsf{\textbf{\small[#1]}}}}
\title{Mathematical Proof as a Litmus Test: Revealing Failure Modes \\ of Advanced Large Reasoning Models}

\author{Dadi Guo\thanks{Equal contribution.}  ~~~Jiayu Liu$^*$ ~~~Zhiyuan Fan ~~~Zhitao He ~~~\textbf{Haoran Li} ~~~\textbf{Yuxin Li}  ~~~\textbf{Yumeng Wang} \\ \textbf{Yi R. (May) Fung} \\
  Hong Kong University of Science and Technology \\
  \texttt{dguoae@connect.ust.hk} ~~~~~~\texttt{yrfung@cse.ust.hk} \\}

\begin{document}
\maketitle
\input{latex/0_abstract}
\input{latex/1_introduction}
\input{latex/2_related_work}
\input{latex/3_method}

\input{latex/4_experiment}
\input{latex/5_conclusion}

\section*{Limitations}

While \textbf{\textit{RFMDataset}} effectively uncovers reasoning deficiencies, certain limitations affect broader interpretation. These choices reflect a practical balance between resource constraints, like cost and time, and the depth of analysis sought. The dataset's selected size (200 problems) means the generalizability of identified failure modes to broader mathematical contexts may need further investigation. Our focus on natural language proofs, while practical for current models, means conclusions don't directly extend to formal mathematical systems.

\section*{Ethics Statements}
\paragraph{Offensive Content Elimination.}
Our benchmark curation pipeline does not involve generating content using LLMs. Instead, all content is manually collected and carefully validated, ensuring that the dataset remains free from offensive material. As such, we are confident that the dataset is safe and does not pose any negative societal impact.

\paragraph{Licenses.}
We will release our code under the MIT license to allow unrestricted research use. The \textbf{\textit{RFMDataset}} will be distributed under a Creative Commons (CC) license, likewise offering free access to the academic community. We take full responsibility for any potential rights violations or issues related to licensing. All licenses associated with our resources support access for research purposes and comply with their respective terms of use.

\paragraph{Annotations.}
The data annotation and verification process was conducted entirely by the six authors of this paper. The team is composed of computer science researchers (one Ph.D., two Ph.D. students, one M.Phil. student, and two senior undergraduates). A crucial qualification for this task was the team's shared, significant background in competitive mathematics, which provided the necessary domain expertise to accurately assess the logical soundness of proofs across various difficulty levels. As the annotation was an integral part of our research effort, considerations of external recruitment and compensation do not apply, in line with standard academic practice.

Our annotation process was guided by a systematic protocol. The instructions for annotation were the detailed evaluation framework and failure mode rubric presented in this paper, which were developed iteratively by the team itself. To ensure the highest quality for our ground-truth labels, we employed a collaborative review and consensus protocol rather than calculating a traditional Inter-Annotator Agreement (IAA). While IAA is vital for subjective tasks, verifying mathematical correctness is an objective process where disagreements are factual disputes to be resolved. In our protocol, each proof was annotated by a primary author. Any complex or ambiguous cases were then flagged for a second review, with the final label and error classification determined through discussion and consensus among the involved authors.

\bibliography{custom}

\clearpage

\appendix

\section{Experiment Details}
\label{app:experiment-details}

\paragraph{Models.} The models we test include: \textit{Deepseek-R1-0120}, \textit{Deepseek-R1-0528}~\cite{deepseek-r1}, \textit{Qwen3-235B-A22B}~\cite{Yang2025Qwen3TR}, \textit{Doubao-1.5-thinking-pro-0415}~\cite{seed_thinking}, \textit{Claude-3.7-Sonnet-Thinking}~\cite{anthropic2025claude}, \textit{Gemini-2.5-pro-preview-0506}, \textit{Gemini-2.5-pro-preview-0605}~\cite{deepmind2025gemini}, \textit{GPT-o1}~\cite{openai-o1}, \textit{GPT-o3-0416}, and \textit{GPT-o4-mini-0416}~\cite{openai2025o3}.

\paragraph{Prompts.} For prompting the model to generate proofs, We use four kinds of prompts in all the experiments: 
\textit{Vanilla}, \textit{reflection\_l}, \textit{reflection\_vi} and \textit{reflection\_lvi}, which is reported in Table~\ref{tab:prove-prompt-1}. 
In the evaluation stage, we unifiedly use one prompt which is presented in Figure~\ref{fig:eval-prompt}. 
Prompt permutations were not used in our baselines, as our focus on failure modes obviates the need for parallel sampling.

\section{Examples of Question}
\label{sec:examples}

This appendix presents a curated selection of example questions from the \textit{\textbf{RFMDataset}}, designed to evaluate the mathematical reasoning capabilities of large language models (LLMs) in constructing rigorous
proofs. 
The questions are drawn from the dataset described in the main document and are organized
by educational level that includes four problems of increasing difficulty (Levels 1 to 4). 
Examples of middle school, high school and university level questions are shown in Figure~\ref{fig:example-questions-ms},~\ref{fig:example-questions-hs} and~\ref{fig:example-questions-ug}.

\section{Failure case analysis}
\label{app:detail-case-analysis}

\subsection{Failure case analysis of judge model}
\label{app:case-analysis-judge}
In our observations, the judge model is mostly accurate in its assessments. However, it tends to be overly strict, often leading to discrepancies with human annotations. In Figure \ref{fig:judge-1}, we provide a case where the judge model is excessively rigid, insisting that the proof must strictly rely on the given conditions in the problem and proceed step-by-step from those premises—thus labeling it as an \textit{Incomplete Proof}. However, the problem itself does not restrict the scope of knowledge, meaning any reasonable theorem not explicitly mentioned could still be validly used. Based on this, we conclude that this is a misjudgment by the judge model.

\subsection{Failure case analysis of proof model}

\label{app:case-analysis-inference}

\begin{table}[h]
    \small
    \centering
    \begin{tabular}{p{0.50\columnwidth} p{0.30\columnwidth}}
        \toprule
        \textbf{Failure case type} & \textbf{Example}\\
        \hline
        \textit{Transformation Error} & Figure~\ref{fig:logical-violation-1} \\
        \textit{Over Generalization} & Figure~\ref{fig:over-generalization-1} \\
        \textit{Invalid Construction} & Figure~\ref{fig:invalid-construction-1}, ~\ref{fig:invalid-construction-2} \\
        \textit{Wrong Division} & Figure~\ref{fig:incomplete-proof-2}, ~\ref{fig:wrong-division-2} \\
        \textit{Circular reasoning} & Figure~\ref{fig:circular-reasoning-1},~\ref{fig:circular-reasoning-2} \\
        \textit{Logic Violation} & Figure~\ref{fig:logical-violation-1},~\ref{fig:logical-violation-2} \\
        \textit{Hidden Assumption} & Figure~\ref{fig:hidden-assumption-1}, ~\ref{fig:hidden-assumption-2} \\
        \textit{Boundary Neglect} & Figure~\ref{fig:boundary-neglect-1}, ~\ref{fig:boundary-neglect-2} \\
        \textit{Vague Argument} & Figure~\ref{fig:invalid-construction-1}, ~\ref{fig:vague-argument-2} \\
        \textit{Incomplete proof} & Figure~\ref{fig:incomplete-proof-2}, ~\ref{fig:incomplete-proof-1} \\
        \textit{Others} & Figure~\ref{fig:others} \\
        \bottomrule
    \end{tabular}
    \caption{A total of eleven failure modes are included our evaluation pipeline. The table serves as a guide to the concrete examples of corresponding failure modes.}
    \label{tab:failure-case-indexer}
\end{table}

In this appendix, we present detailed failure cases from our evaluation of large language models on mathematical proof problems in the \textbf{\textit{RFMDataset}}. 
Each entry includes the original problem statement, specifies the model that was used, identifies the type of failure mode, and provides an analysis of the resulting error.
The red italicized text indicates the erroneous parts in the proof, while the blue text in brackets provides explanations for the errors and their failure mode.
Table~\ref{tab:failure-case-indexer} serves as a guide to examples of the failure case studies.


\section{Data Processing} 

\subsection{Data collection}
\label{app:data-processing}
Upon identifying suitable problems, we preserve them via screenshots. 
We then leverage Kimi~\cite{kimi-k1.5} to extract and convert the textual content into LaTeX format, while manually annotating each problem with its respective difficulty level and knowledge domain tags. 
Note that for geometry problems that require diagrams, we reformulate the textual descriptions to ensure the scene can be fully reconstructed from natural language alone. 
This dual approach ensures that we rigorously test the models' ability to reason purely through linguistic representations.

It's particularly noteworthy that models tend to reduce the inherent difficulty of geometry problems by establishing coordinate systems or parametric equations, which circumvents our intention to test their geometric reasoning abilities. 
To counteract this and preserve the intended difficulty level, we explicitly instruct the models not to employ such methods in geometry problems, requiring them to use only the provided knowledge.

\subsection{Data annotation}
\label{app:data-annotation}

To facilitate human evaluation and establish a ground truth for assessing model performance, a human annotation process was conducted. 

For each sampled instance, human annotators were presented with the original mathematical question and the corresponding model-generated proof. A single proof was structured with a </think> tag separating the internal reasoning from the final answer.
To ensure consistent evaluation, particularly as certain models do not expose their reasoning chains through API services, judgments concerning correctness and the classification of failure modes were exclusively derived from the content after the </think> tag, which constitutes the final answer or output of the model. 


\subsection{Data Source}
\label{data_source}
We analyze the proportions of the four kinds of data sources in our dataset: Textbook, Exam, Internet Media, and Competition, as shown in Table~\ref{tab:data_source}. To highlight the distinct data sources of our dataset compared to other proof datasets, we present a comparison of the data sources for MiniF2F~\cite{miniF2F}, MiniCTX~\cite{Hu2024miniCTXNT}, and ProofNet~\cite{ProofNet} against ours, as shown in Table~\ref{tab:data_sources_comparison}.
\begin{table}[h!]
    \small
    \centering
    \begin{tabular*}{\columnwidth}{@{}c@{\extracolsep{\fill}}c@{\extracolsep{\fill}}c@{\extracolsep{\fill}}c@{\extracolsep{\fill}}c@{}}
        \toprule
        \textbf{Source} & \textbf{MS} & \textbf{HS} & \textbf{UG} & \textbf{Overall} \\
        \hline
        Textbook & 20.00 & \phantom{0}0.00 & 16.67 & \phantom{0}9.90 \\
        Exam & 24.00 & 50.00 & 15.00 & 32.67 \\
        Internet Media & 34.00 & 45.56 & 55.00 & 45.05 \\
        Competition & 26.00 & \phantom{0}2.22 & 13.33 & 11.39 \\
        \bottomrule
    \end{tabular*}
    \caption{The distribution of data sources in \textit{\textbf{RFMDataset}}. All numbers are percentages (\%).} 
    \label{tab:data_source} 
\end{table}

\begin{table}[h!]
    \small
    \centering
    \begin{tabular*}{0.95\columnwidth}{l @{\extracolsep{\fill}} >{\centering\arraybackslash}p{0.70\columnwidth}} 
        \toprule
        \textbf{Dataset} & \textbf{Data Sources} \\
        \hline
        MiniF2F & AMC, AIME, IMO, MATH dataset, textbook \\
        \hline
        MiniCTX & Lean 4 Zulip, Books: Mathematics in Lean, The Mechanics of Proof \\
        \hline
        ProofNet & Undergraduate Textbooks: Elementary Real Analysis, Introduction to Abstract Algebra, etc. \\
        \hline
        \textbf{\textit{RFMDataset}} & Textbook, Exam, Internet Media, Competition \\
        \bottomrule
    \end{tabular*}
    \caption{Comparison of Data Sources Across Different Datasets.}
    \label{tab:data_sources_comparison}
\end{table}

\section{More analysis}
\subsection{Is failure mode relevant to problem difficulty?}
\label{app:different-difficulty-same-failure-mode}
\begin{figure}[htbp]
    \centering
    \includegraphics[width=0.95\linewidth]{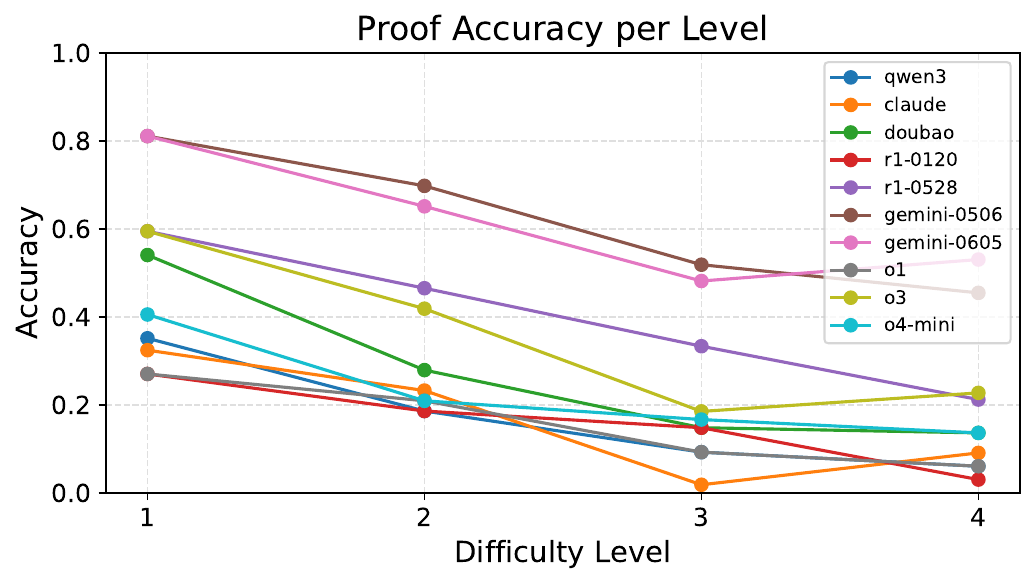}
    \caption{Accuracy of different models on each difficulty level. }
    \label{fig:accuracy_per_level}
\end{figure}
\begin{figure*}[t]
    \centering
    \subfloat[][Failure mode distribution across level-1 difficulty problems\label{fig:sub1}]{
        \includegraphics[width=0.48\linewidth]{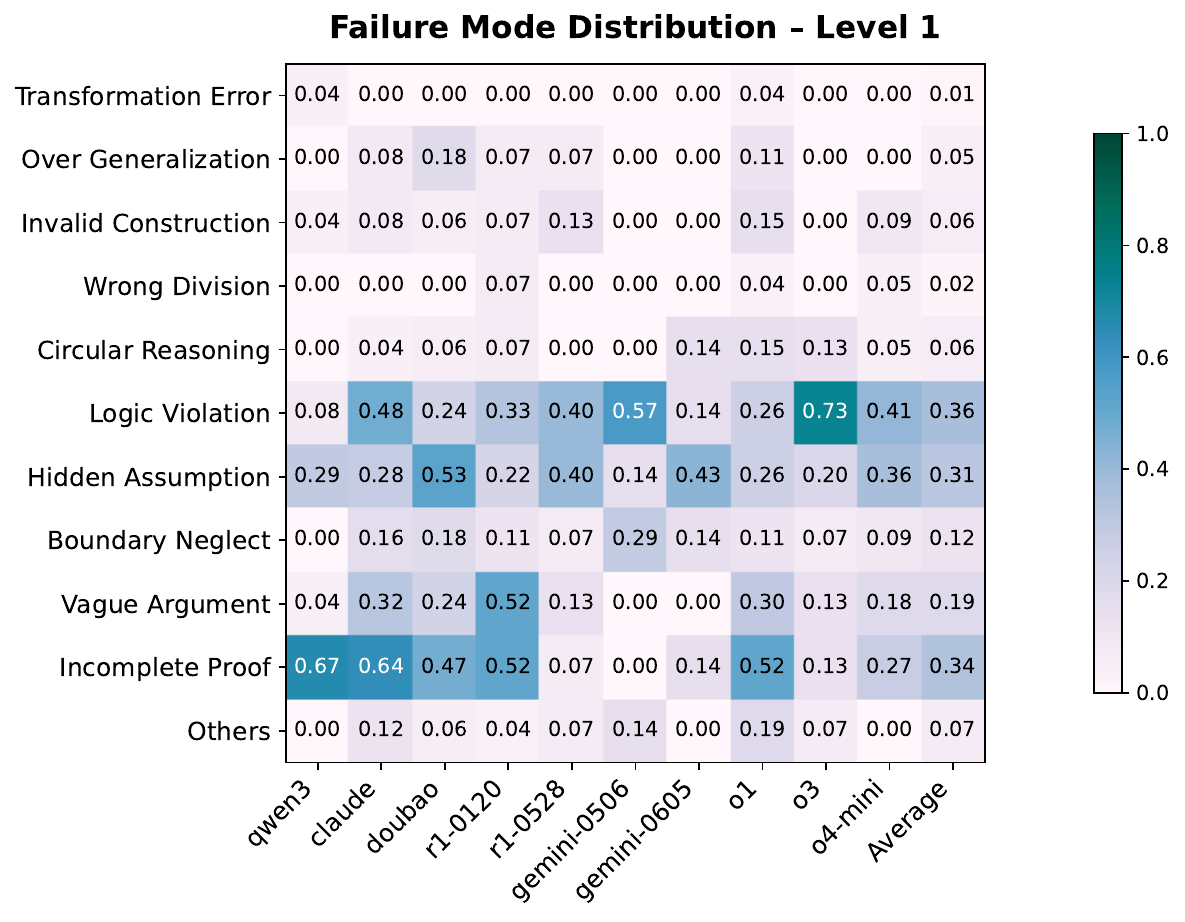}
    }
    \hfill
    \subfloat[][Failure mode distribution across level-2 difficulty problems\label{fig:sub2}]{
        \includegraphics[width=0.48\linewidth]{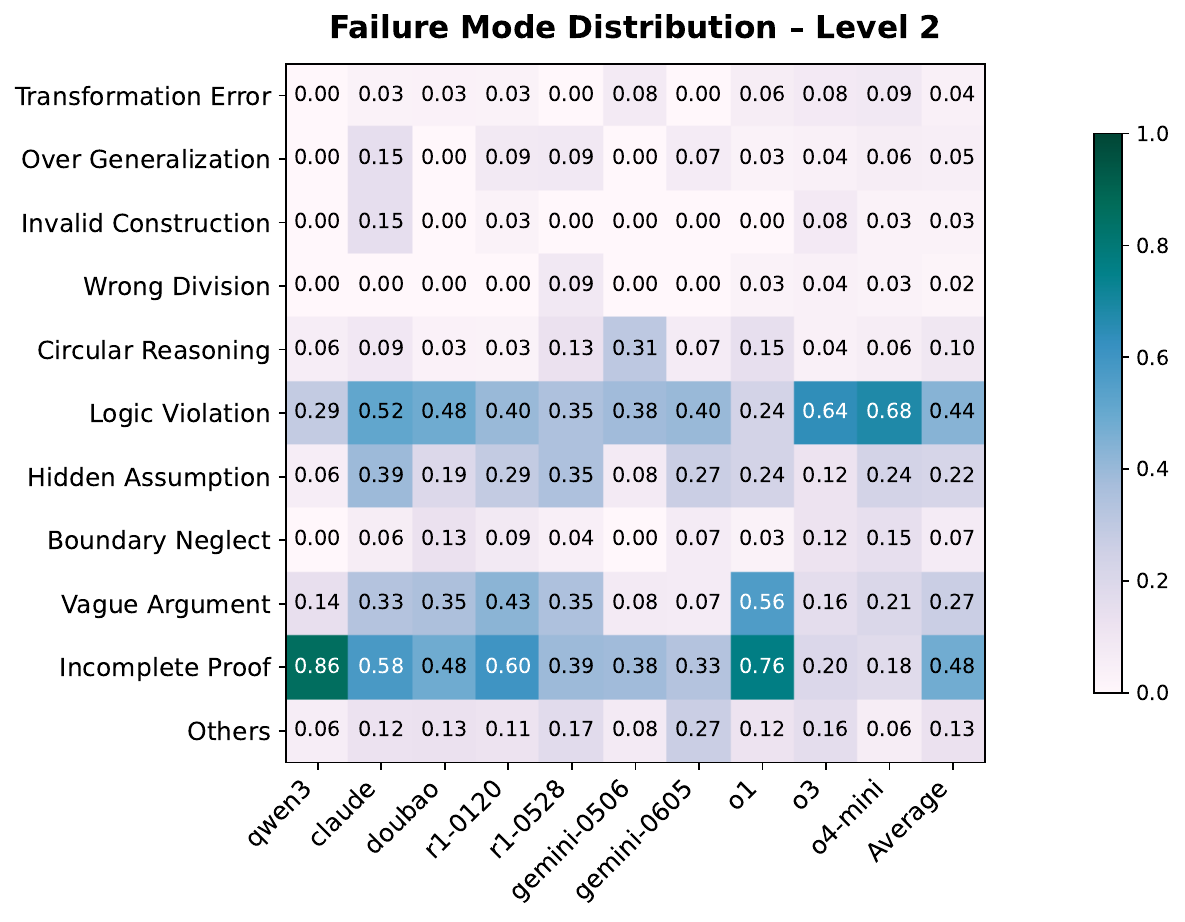}
    }
    
    \vspace{0.5cm} 
    
    \subfloat[][Failure mode distribution across level-3 difficulty problems\label{fig:sub3}]{
        \includegraphics[width=0.48\linewidth]{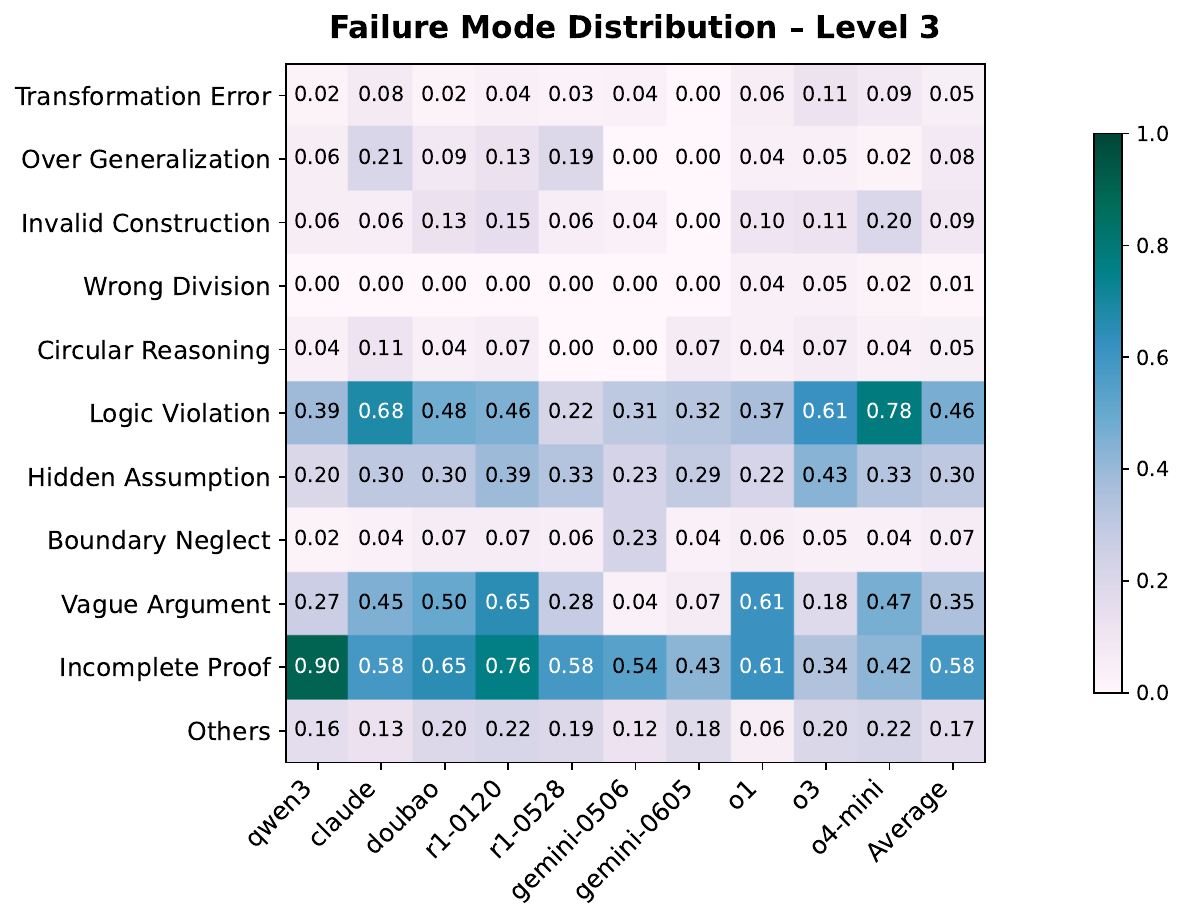}
    }
    \hfill
    \subfloat[][Failure mode distribution across level-4 difficulty problems\label{fig:sub4}]{
        \includegraphics[width=0.48\linewidth]{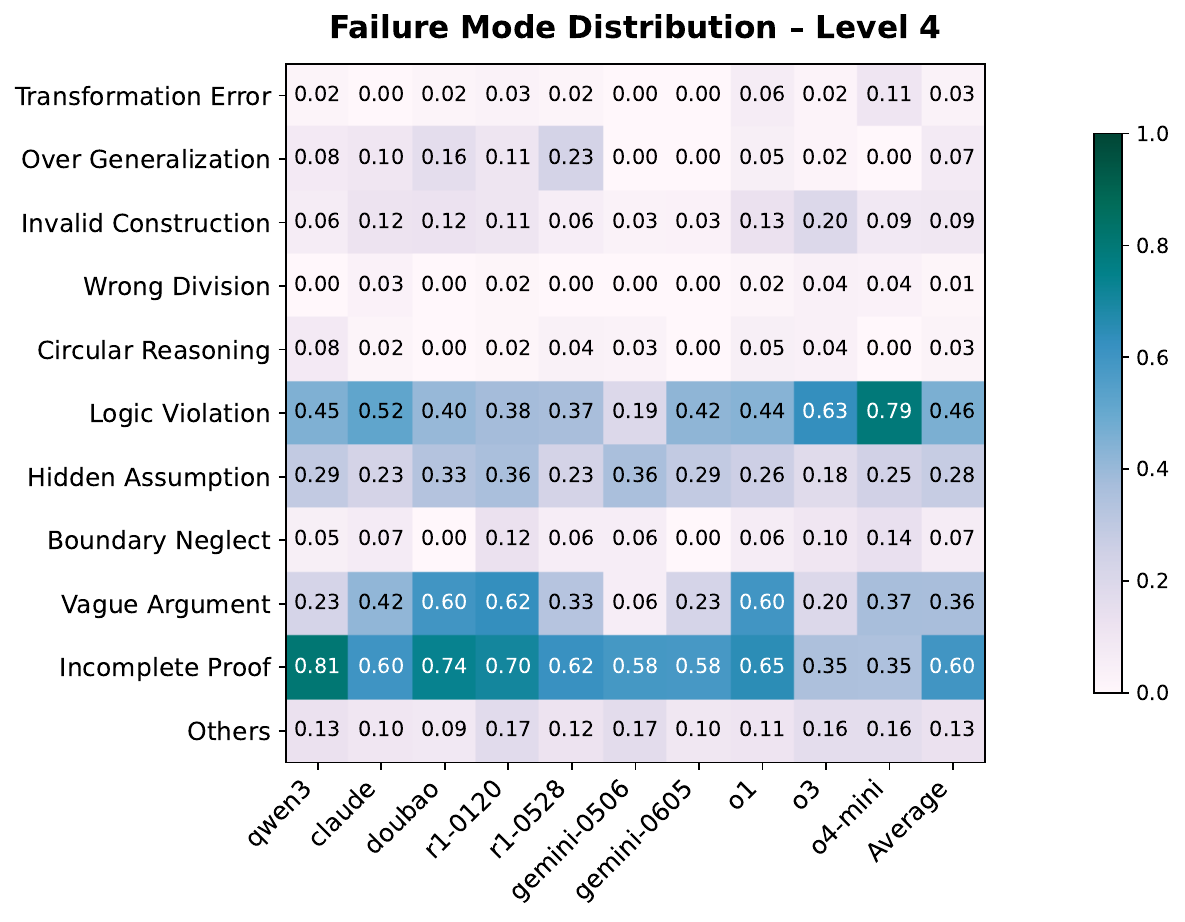}
    }
    
    \caption{Distribution of failure modes across questions of four difficulty levels. It is evident that the failure mode pattern of models is similar.}
    \label{fig:mode_level}
\end{figure*}

It is natural for models' accuracy to decline as problem difficulty increases, as illustrated in the Figure \ref{fig:accuracy_per_level}. However, how the distribution of failure modes changes with difficulty remains unknown. Figure \ref{fig:mode_level} illustrates the distribution of failure modes across questions of varying difficulty. 
It's evident that, across all four difficulty levels, model failures are predominantly concentrated in four patterns: \textit{logical violation}, \textit{vague argument}, \textit{hidden assumption}, and \textit{incomplete proof}. 
Notably, the proportions of \textit{logical violation}, \textit{vague argument}, and \textit{incomplete proof} show increasing trends with difficulty, while \textit{hidden assumption} demonstrates a slight downward trend.

\subsection{Does a longer chain-of-thought lead to better problem-solving performance?}
\label{app:different-domain-different-difficulty.}

\begin{figure*}[t]
    \centering
    \subfloat[][Comparison of accuracy and response length across domains of \textit{Claude-3.7-Sonnet-Thinking}]{
        \includegraphics[width=0.48\linewidth]{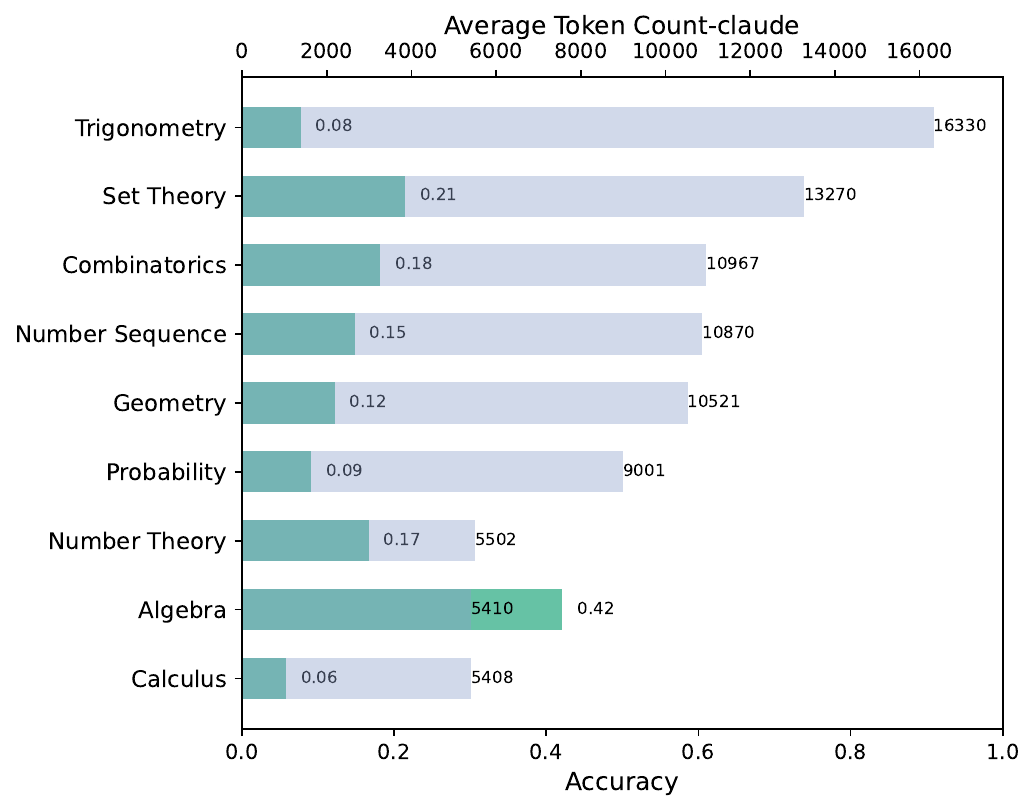}
    }
    \hfill
    \subfloat[][Comparison of accuracy and response length across domains of \textit{Doubao-1.5-thinking-pro}]{
        \includegraphics[width=0.48\linewidth]{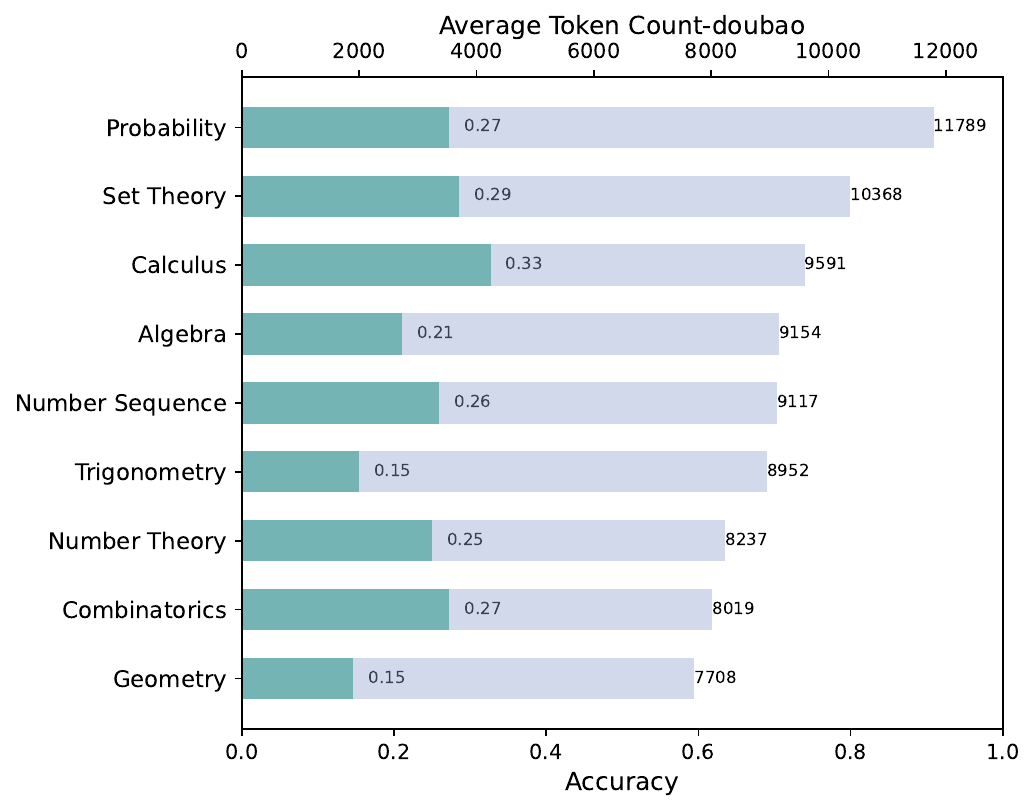}
    }
    
    \vspace{0.5cm} 
    
    \subfloat[][Comparison of accuracy and response length across domains of \textit{Qwen3-235B-A22B}]{
        \includegraphics[width=0.48\linewidth]{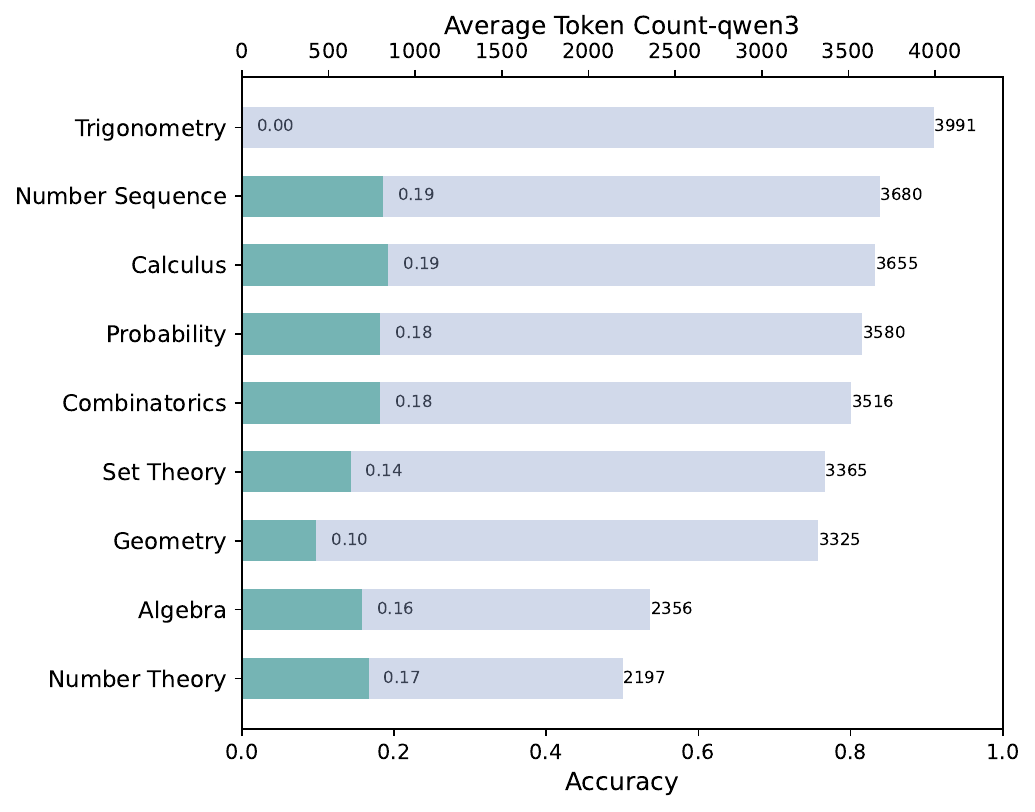}
    }
    \hfill
    \subfloat[][Comparison of accuracy and response length across domains of \textit{Deepseek-R1-0120}]{
        \includegraphics[width=0.48\linewidth]{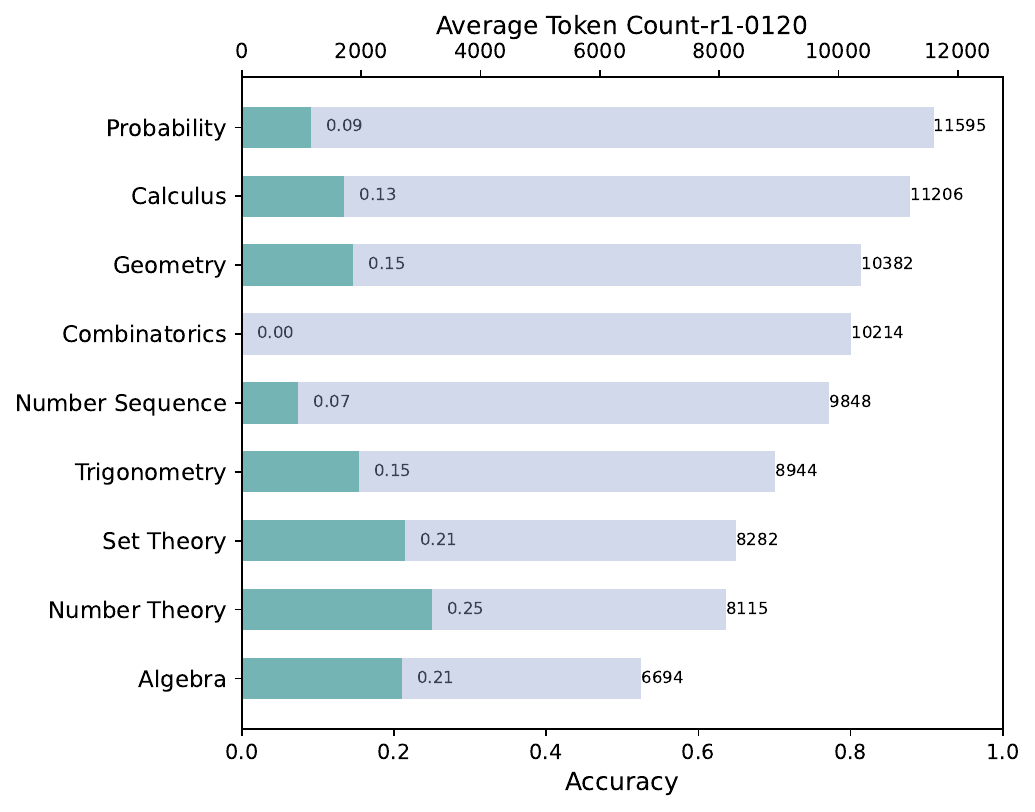}
    }
    
    \caption{The solid green bars represent average accuracy and the translucent blue bars show average answer length in tokens for each domain. 
    The domains are ordered from top to bottom by decreasing token count. 
    As the figures show, an increase in token count doesn't consistently lead to improved accuracy.}
    \label{fig:acc_token}
\end{figure*}

Figure \ref{fig:acc_token} presents an evaluation of four models (\textit{Claude-3.7-Sonnet-Thinking, Doubao-1.5-thinking-pro, Qwen3-235B-A22B}, and \textit{Deepseek-R1-0120}) analyzing their accuracy and average token counts across various domains with publicly available chain-of-thought data. 
Our findings show \textbf{no clear correlation between reasoning chain length and accuracy} across domains. In fact, longer reasoning chains often lead to lower accuracy.

Given that these models should theoretically grasp basic mathematical theorems, we hypothesize this discrepancy arises from two factors: varying levels of methodological mastery across different domains, and error accumulation during the reasoning process. 
More fundamentally, these models exhibit a limited capacity for first-principles exploration using basic mathematical theorems, performing significantly worse than mathematically proficient humans.
As Timothy Gowers \footnote{Fields Medalist (1998), Professor at the University of Cambridge and \textup{Collège de France}.} recently highlighted~\cite{gowers2025llmproofs}, empowering models to truly think like human mathematicians remains an unsolved challenge.

\begin{figure*}[h]
\centering
\vspace{-0.0em}
\subfloat{\includegraphics[trim={0cm, 0cm 0cm 0cm},height=2in]{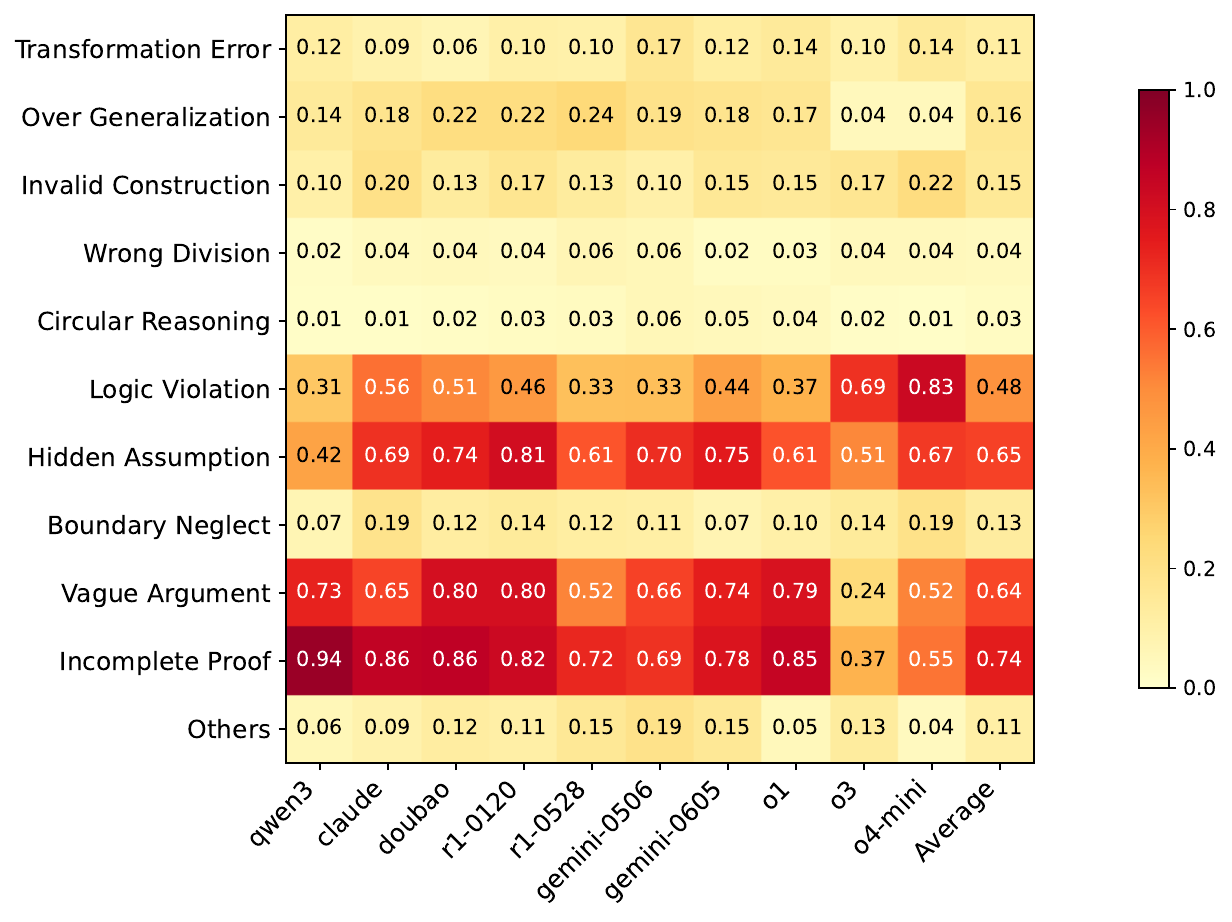}%
\label{fig:error_pattern_model_gpt}}
\hfil
\subfloat{\includegraphics[trim={0cm, 0cm 0cm 0cm},height=2.04in]{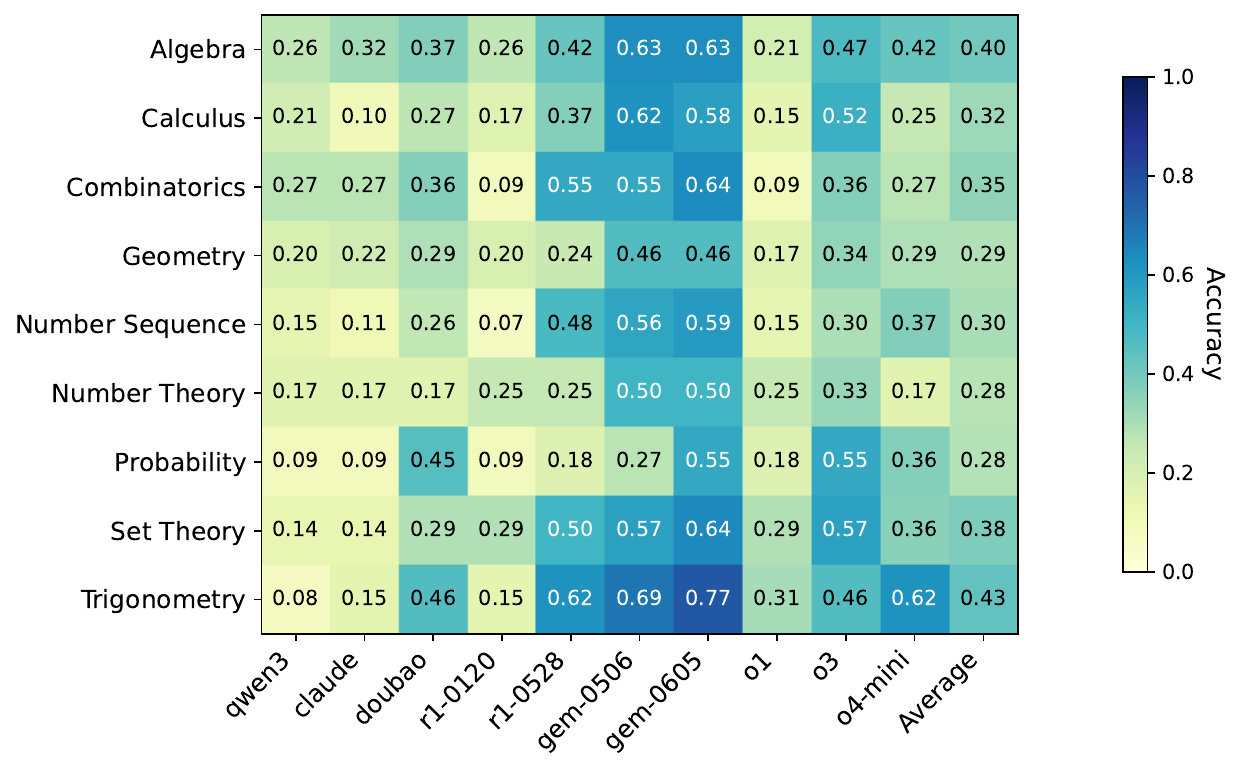}%
\label{fig:heatmap_tag_model_gpt}}
\caption{\textit{(a)} The proportion of each error pattern of each model. The \textit{Average} column summarizes the average proportion across all models for each pattern. \textit{~(b)} Accuracy of each model on every knowledge domain. The \textit{Average} column summarizes the average accuracy across all models for each domain. This failure mode distribution is from \textit{GPT-5}, which is consistent with \textit{Gemini}'s assessment of the distribution (Figure \ref{fig:error_pattern_model} and Figure \ref{fig:heatmap_tag_model}).}
\label{fig:heapmap_all_gpt}
\end{figure*}

\subsection{How do other LLM judges perform?}
\label{judge_llms}
In our primary experiments, we utilized the strong \textit{Gemini-2.5-pro-preview-0506} model as the judge to evaluate the overall correctness of proofs and the distribution of failure modes. We then demonstrate the consistency between Gemini's generated labels and human labels using correlation coefficients. However, researchers may wonder about the performance of other less powerful models when acting as judges. Therefore, we include three additional judges for supplementary experiments: \textit{DeepSeek-r1-0528}, \textit{Doubao-1.6-thinking-pro}, and notably, \textit{GPT-5}, which serves as an evaluator of comparable capability to our primary judge, \textit{Gemini-2.5-pro-preview-0506}.

The comparison of evaluation results from different judges, presented in Table \ref{tab:judge-diff}, reveals significant disparities but also a crucial pattern of consistency. Notably, the assessments from our supplementary peer judge, \textit{GPT-5}, show a high degree of agreement with our primary judge, \textit{Gemini-2.5-pro-preview-0506}. For instance, when evaluating the \textit{Gemini-2.5-pro-preview-0506} and \textit{Gemini-2.5-pro-preview-0605} models, the overall scores assigned by the Gemini judge (59.00 and 59.50) are very close to those from \textit{GPT-5} (55.00 and 57.50, respectively). This strong consistency between two state-of-the-art models validates our choice of \textit{Gemini-2.5-pro-preview-0506} as a reliable primary evaluator.

In sharp contrast, the other judges are significantly more lenient and show a large gap compared to \textit{Gemini-2.5-Pro} and \textit{GPT-5}. Specifically, \textit{Doubao-1.6-thinking-pro} consistently yields higher scores than \textit{DeepSeek-R1-0528}, and both are far more generous than \textit{Gemini} and \textit{GPT-5}. Despite these differences in absolute scores, it is important to note that the relative performance ranking of the evaluated models remains largely consistent across all judges.

Given \textit{Gemini-2.5-pro-preview-0506}'s superior performance~\cite{Huang2025Gemini2P} and its consistency with human evaluations, we consider the evaluation results from our main experiments to be reliable. We attribute the inconsistency among Judge LLMs to inherent differences in their ability to evaluate mathematical propositions and \textit{Gemini-2.5-pro-preview-0506}'s stricter judging criteria. This discrepancy not only underscores the critical importance of robust proof evaluation but also reveals significant performance gaps among current models as evaluators, marking this as a crucial area for future investigation.

Figures \ref{fig:error_pattern_model_gpt} and \ref{fig:heatmap_tag_model_gpt} illustrate the failure mode and per-domain accuracy distributions under the \textit{GPT-5} judge, respectively. Crucially, both distributions are highly consistent with those from our primary \textit{Gemini} judge, which reinforces the credibility and persuasiveness of our fine-grained analysis.

\begin{table*}[h!]
    \small 
    \centering
    \begin{tabular}{p{3.8cm} >{\centering\arraybackslash}p{1cm} >{\centering\arraybackslash}p{1cm} >{\centering\arraybackslash}p{1cm} >{\centering\arraybackslash}p{1cm} l}
        \toprule
        \textbf{Model} & \textbf{MS} & \textbf{HS} & \textbf{UG} & \textbf{Overall} & \textbf{Judge Model} \\
        \midrule
        \textit{Qwen3-235B-A22B} & 15.38 & \phantom{0}9.09 & 23.33 & 15.00 & \textit{Gemini-2.5-pro-preview-0506} \\
        \textit{Qwen3-235B-A22B} & 17.31 & 18.18 & 33.33 & 22.50 & \textit{Deepseek-R1-0528} \\
        \textit{Qwen3-235B-A22B} & 23.08 & 15.91 & 28.33 & 21.50 & \textit{Doubao-1.6-thinking-pro} \\
        \textit{Qwen3-235B-A22B} & 25.00 & 12.50 & 21.67 & 18.50 & \textit{GPT-5} \\
        \midrule
        \textit{Doubao-1.5-thinking-pro} & 19.23 & 21.59 & 33.33 & 24.50 & \textit{Gemini-2.5-pro-preview-0506} \\
        \textit{Doubao-1.5-thinking-pro} & 38.46 & 35.23 & 38.33 & 37.00 & \textit{Deepseek-R1-0528} \\
        \textit{Doubao-1.5-thinking-pro} & 48.08 & 50.00 & 46.67 & 48.50 & \textit{Doubao-1.6-thinking-pro} \\
        \textit{Doubao-1.5-thinking-pro} & 32.69 & 26.14 & 35.00 & 30.50 & \textit{GPT-5} \\
        \midrule
        \textit{Deepseek-R1-0120} & 17.31 & 10.23 & 16.67 & 14.00 & \textit{Gemini-2.5-pro-preview-0506} \\
        \textit{Deepseek-R1-0120} & 25.00 & 29.55 & 31.67 & 29.00 & \textit{Deepseek-R1-0528} \\
        \textit{Deepseek-R1-0120} & 36.54 & 25.00 & 43.33 & 33.50 & \textit{Doubao-1.6-thinking-pro} \\
        \textit{Deepseek-R1-0120} & 25.00 & 10.23 & 21.67 & 17.50 & \textit{GPT-5} \\
        \midrule
        \textit{Deepseek-R1-0528} & 26.92 & 42.05 & 38.33 & 37.00 & \textit{Gemini-2.5-pro-preview-0506} \\
        \textit{Deepseek-R1-0528} & 48.08 & 60.23 & 58.33 & 56.50 & \textit{Deepseek-r1-0528} \\
        \textit{Deepseek-R1-0528} & 51.92 & 69.32 & 71.67 & 65.50 & \textit{Doubao-1.6-thinking-pro} \\
        \textit{Deepseek-R1-0528} & 28.85 & 39.77 & 43.33 & 38.00 & \textit{GPT-5} \\
        \midrule
        \textit{Claude-3.7-Sonnet-Thinking} & 19.23 & 13.64 & 11.67 & 14.50 & \textit{Gemini-2.5-pro-preview-0506} \\
        \textit{Claude-3.7-Sonnet-Thinking} & 26.92 & 27.27 & 16.67 & 24.00 & \textit{Deepseek-R1-0528} \\
        \textit{Claude-3.7-Sonnet-Thinking} & 30.77 & 34.09 & 26.67 & 31.00 & \textit{Doubao-1.6-thinking-pro} \\
        \textit{Claude-3.7-Sonnet-Thinking} & 25.00 & 13.64 & 13.33 & 16.50 & \textit{GPT-5} \\
        \midrule
        \textit{Gemini-2.5-pro-preview-0506} & 51.92 & 53.41 & 73.33 & 59.00 & \textit{Gemini-2.5-pro-preview-0506} \\
        \textit{Gemini-2.5-pro-preview-0506} & 63.46 & 67.05 & 83.33 & 71.00 & \textit{Deepseek-R1-0528} \\
        \textit{Gemini-2.5-pro-preview-0506} & 76.92 & 72.72 & 90.00 & 79.00 & \textit{Doubao-1.6-thinking-pro} \\
        \textit{Gemini-2.5-pro-preview-0506} & 55.77 & 48.86 & 63.33 & 55.00 & \textit{GPT-5} \\
        \midrule
        \textit{Gemini-2.5-pro-preview-0605} & 50.00 & 60.23 & 66.67 & 59.50 & \textit{Gemini-2.5-pro-preview-0506} \\
        \textit{Gemini-2.5-pro-preview-0605} & 65.38 & 71.59 & 78.33 & 72.00 & \textit{Deepseek-R1-0528} \\
        \textit{Gemini-2.5-pro-preview-0605} & 78.85 & 78.41 & 83.33 & 80.00 & \textit{Doubao-1.6-thinking-pro} \\
        \textit{Gemini-2.5-pro-preview-0605} & 55.77 & 54.55 & 63.33 & 57.50 & \textit{GPT-5} \\
        \midrule
        \textit{Gemini-3-pro-preview} & 57.69 & 48.86 & 66.67 & 56.50 & \textit{Gemini-2.5-pro} \\
        \textit{Gemini-3-pro-preview} & 63.46 & 59.09 & 70.00 & 63.50 & \textit{GPT-5} \\
        \midrule
        \textit{DeepSeek-V3.2-Speciale} & 63.46 & 68.18 & 66.67 & 66.50 & \textit{Gemini-2.5-pro} \\
        \textit{DeepSeek-V3.2-Speciale} & 73.08 & 75.00 &80.00 & 76.00 & \textit{GPT-5} \\
        \midrule
        \textit{GPT-o1} & \phantom{0}7.69 & 14.77 & 18.33 & 14.00 & \textit{Gemini-2.5-pro-preview-0506} \\
        \textit{GPT-o1} & 19.23 & 34.09 & 36.67 & 31.00 & \textit{Deepseek-R1-0528} \\
        \textit{GPT-o1} & 15.38 & 46.59 & 45.00 & 38.00 & \textit{Doubao-1.6-thinking-pro} \\
        \textit{GPT-o1} & 17.31 & 17.05 & 21.67 & 18.50 & \textit{GPT-5} \\
        \midrule
        \textit{GPT-o3-0416} & 30.77 & 29.55 & 38.33 & 32.50 & \textit{Gemini-2.5-pro-preview-0506} \\
        \textit{GPT-o3-0416} & 46.15 & 44.32 & 50.00 & 46.50 & \textit{Deepseek-R1-0528} \\
        \textit{GPT-o3-0416} & 53.85 & 61.36 & 75.00 & 63.50 & \textit{Doubao-1.6-thinking-pro} \\
        \textit{GPT-o3-0416} & 40.38 & 36.36 & 55.00 & 43.00 & \textit{GPT-5} \\
        \midrule
        \textit{GPT-o4-mini-0416} & 15.38 & 22.73 & 23.33 & 21.00 & \textit{Gemini-2.5-pro-preview-0506} \\
        \textit{GPT-o4-mini-0416} & 26.92 & 39.77 & 30.00 & 33.50 & \textit{Deepseek-R1-0528} \\
        \textit{GPT-o4-mini-0416} & 34.62 & 53.41 & 41.67 & 45.00 & \textit{Doubao-1.6-thinking-pro} \\
        \textit{GPT-o4-mini-0416} & 28.85 & 37.50 & 28.33 & 32.50 & \textit{GPT-5} \\
        \bottomrule
    \end{tabular}
    \caption{The evaluation discrepancies among different judge models are notable. We observe that the correctness rates evaluated by \textit{Doubao-1.6-thinking-pro} are consistently higher than those from \textit{Deepseek-R1-0528}, which in turn are higher than \textit{Gemini-2.5-pro-preview-0506}'s. Evidently, there are substantial differences in the accuracy provided by these various judge models.}
    \label{tab:judge-diff}
    \vspace{-0.2in}
\end{table*}
\section{Possible Solutions}
\label{app:possible_solutions}
Based on the analysis of failure modes, we discuss possible solutions to improve future models. We focus on two key aspects aiming at making the model think like a mathematician: first, how to enhance the model's performance in weak areas(solutions 1), and second, how to enhance the reliability and verifiability of the models' reasoning processes (solutions 2).

\paragraph{Solution 1: Training on high-quality domain-specific data.} Our experimental results reveal significant variations in model performance across different mathematical domains. Taking geometry problems as an example, as illustrated in Figures \ref{fig:wrong-division-2} and \ref{fig:circular-reasoning-1}, models still exhibit severe errors in proving even the two most fundamental properties and theorems within this field. This discrepancy strongly suggests weakness in domain-specific knowledge. Furthermore, mathematical problems from different domains may necessitate distinct techniques and methodologies. Without training on data related to these specific methodologies, models may struggle to solve problems, even if they possess long chain-of-thought exploration capabilities~\cite{Wang2025DiversityEnhancedRF}. Consequently, a direct solution may involve training on high-quality data targeted at weak domains~\cite{onda2025leanconjecturer}. 

\paragraph{Solution 2: Agentic step-level interaction with formally verifiable environment.}Our analysis of failure modes indicates that precise single-step reasoning remains a significant challenge for current models. A crucial reason for this lies in the unreliable and often untimely nature of the models' self-generated reflections. Therefore, we propose addressing this by enabling agentic, multi-turn interactions~\cite{Huang2025Gemini2P,wang2025ma,wang2025let,jiang2025pag, Critical, CostBench, MarCon, MarPT} with formally verifiable environments, such as Lean. When a model needs to validate the correctness of its claims~\cite{Veri-R1, llm-retrieval-failure-mode}, it could translate its reasoning into Lean code and request a response from the environment. This external feedback mechanism offers far greater precision than internal self-reflection. Furthermore, as models are increasingly tasked with solving research-level mathematical problems~\cite{zhang2025realmath}, a single reasoning pass will be insufficient. This necessitates the integration of memory mechanisms from the agent domain to store current progress (e.g., already proven lemmas). Training and evaluation should also be conducted through multi-turn generation in an interactive environment.

\clearpage

\begin{table*}[h]
\centering
\begin{tabular}{p{0.18\textwidth}|p{0.82\textwidth}}
\toprule
 & \textbf{Prompt in various baselines} \\
\midrule
\textbf{Vanilla} &
\textit{Input}: \newline 
[\textit{Original Problem}]\\
\midrule
\textbf{reflection\_l} & 
\textit{Input}: \newline
You will be required to solve a mathematical proof problem with the following requirements:  \newline

1. If the problem specifies the knowledge scope or restrictions, you must endeavor to meet these requirements.  

2. When you believe the proposition can be proved, first write a draft of the proof, then check the logical correctness of each reasoning step. If any incorrectness is found, rethink and correct the logically flawed steps. Repeat this process until you can finally provide an answer that you consider error-free.\newline

The problem is: [\textit{Original Problem}]
\\
\midrule
\textbf{reflection\_vi} &
\textit{Input}: \newline
You will be required to solve a mathematical proof problem with the following requirements:\newline  

1. If the problem specifies the knowledge scope or restrictions, you must endeavor to meet these requirements.

2. Every step of your final proof process must be completely rigorous and in line with mathematical principles. It should not be a vague argument that relies on intuition, diagrams, or the word "obvious" instead of rigorous justification.  

3. When using non-trivial mathematical theorems, you must prove them based on basic knowledge rather than using them directly.\newline

The problem is: [\textit{Original Problem}]
 \\
\midrule
\textbf{reflection\_lvi} & 
\textit{Input}: \newline
You will be required to solve a mathematical proof problem with the following requirements:\newline  

1. If the problem specifies the knowledge scope or restrictions, you must endeavor to meet these requirements.  

2. Every step of your final proof process must be completely rigorous and in line with mathematical principles. It should not be a vague argument that relies on intuition, diagrams, or the word "obvious" instead of rigorous justification.  

3. When using non-trivial mathematical theorems, you must prove them based on basic knowledge rather than using them directly.

4. When you believe the proposition can be proved, first write a draft of the proof, then check the logical correctness of each reasoning step. If any incorrectness is found, rethink and correct the logically flawed steps. Repeat this process until you can finally provide an answer that you consider error-free.\newline

The problem is: [\textit{Original Problem}]
\\
\bottomrule
\end{tabular}
\caption{This is the prompt for our baseline methods: \textit{Vanilla}, \textit{reflection\_l}, \textit{reflection\_vi} and \textit{reflection\_lvi}. All terms enclosed in square brackets (``[ ]'') are placeholders, which will be dynamically replaced by relevant variables during the evaluation process.}
\label{tab:prove-prompt-1}
\end{table*}

\begin{figure*}[h!]
\begin{tcolorbox}[
    colback=blue!1!white,
    colframe=blue!70!black,
    title=Evaluation Prompt,
    fonttitle=\bfseries,
    colbacktitle=blue!5!white,
    coltitle=blue!50!black,
    boxrule=1pt,
    arc=4pt,
    boxsep=5pt,
    left=6pt,
    right=6pt,
    top=6pt,
    bottom=6pt
]

You are a math teacher highly skilled in evaluating proofs.

**Problem Statement**:
[\textit{Original Question}]

**Student’s Answer**:
[\textit{LLM generated Proof}]\newline

\textbf{Your Tasks} 

1. **Detailed Analysis**: Read the student’s answer and thought process carefully. \newline
   - If a claim is obvious in students's answer, you may accept it as valid provided the overall logic is sound. \newline
   - If the category **“Others”** is triggered, briefly describe the error in the Analysis section.\newline
   
2. **Binary Scoring**: After your analysis, output *only* the exact block below, inside triple back-ticks, nothing else:

`` \newline
Error Pattern Analysis \newline
- Transformation Error: 1|0 \newline
- Over Generalization: 1|0\newline
- Invalid Construction: 1|0\newline
- Wrong Division: 1|0\newline
- Circular Reasoning: 1|0\newline
- Logic Violation: 1|0\newline
- Hidden Assumption: 1|0\newline
- Boundary Neglect: 1|0\newline
- Vague Argument: 1|0\newline
- Vague Argument: 1|0\newline
- Others: 1|0\newline

Overall Correctness \newline
- 1|0 \newline
''\newline

\textbf{Error Pattern Rubric}\newline
Presented in Figure~\ref{fig:error-pattern-rubric}\newline

\textbf{Scoring Semantics} \newline
- In **Error Pattern Analysis**:
  ``1'' = this error pattern **is present**.
  ``0'' = this error pattern **is NOT present**. 

- In **Overall Correctness**:
  ``1'' = the proof is **completely correct** (no errors).
  ``0'' = the proof **contains at least one error**.\newline

\textbf{Consistency Rule}\newline
If **any** error pattern is ``1'', then **Overall Correctness must be ``0''**.
Only when **all** error patterns are ``0'' is Overall Correctness ``1''.\newline

Do not output anything after the code block.
Your answer is: 

\end{tcolorbox}
\caption{This unified evaluation prompt is employed within our LLM-as-a-judge evaluation pipeline. All terms enclosed in square brackets (``[ ]'') are placeholders, which will be dynamically replaced by relevant variables during the evaluation process.}
\label{fig:eval-prompt}
\end{figure*}

\begin{figure*}[h!]
\begin{tcolorbox}[
    colback=blue!1!white,
    colframe=blue!70!black,
    title=Evaluation Prompt,
    fonttitle=\bfseries,
    colbacktitle=blue!5!white,
    coltitle=blue!50!black,
    boxrule=1pt,
    arc=4pt,
    boxsep=5pt,
    left=6pt,
    right=6pt,
    top=6pt,
    bottom=6pt
]

\textbf{Error Pattern Rubric}

1.\newline Category: \textbf{Transformation Error} \newline
Definition: Recasting the target statement into a non-equivalent or strictly weaker one. \newline  Typical example: To prove convergence of $\sum a_n$, only prove $\lim a_n=0$; or replace “$A\iff B$” with “$A\!\Rightarrow\!B$”. \newline
2.\newline Category: \textbf{Over Generalization} \newline
Definition: Inferring a universal claim from a few special or hand-picked cases. \newline  Typical example: Verifying for $n=1,3,5$ then claiming the result holds for all $n\in\mathbb(N)$.  \newline
3.\newline Category: \textbf{Invalid Construction} \newline
Definition: Failing to construct an object that should exist, or constructing one that doesn’t meet requirements. \newline
Typical example: Claiming a function that is everywhere linear yet nowhere differentiable. \newline 
4.\newline Category: \textbf{Wrong Division}
Definition: Partitioning into cases that miss at least one legitimate possibility or overlap.
Typical example: When analyzing the behavior of a function, dividing cases as ``always positive,'' ``always zero,'' and ``always negative.'' \newline
5.\newline Category: \textbf{Circular Reasoning} \newline
Definition: Using the conclusion (or an equivalent reformulation) as a hidden or explicit premise. \newline 
Typical example: Assuming $B$ when trying to prove $A\!\Rightarrow\!B$.  \newline
6.\newline Category: \textbf{Logic Violation} \newline
Definition: A deduction step that contradicts logical or algebraic rules. \newline          Typical example: From $a<b$ and $c<d$ concluding $a-c<b-d$ without checking signs. \newline       
7.\newline Category: \textbf{Hidden Assumption} \newline
Definition: Applying a theorem or step whose hypotheses were neither stated nor proven. \newline 
Typical example: Differentiating a function known only to be continuous.   \newline       
8.\newline Category: \textbf{Boundary Neglect}\newline
Definition: Ignoring edge cases, endpoints, or limiting situations so the argument holds only ``in the middle.''\newline
Typical example: Proving $f(x)=\sqrt(x)$ differentiable on $[0,1]$ without checking at $x=0$. \newline
9.\newline Category: \textbf{Vague Argument} \newline      
Definition: Relying on intuition, diagrams, or ``obvious'' without formal justification. \newline Typical example: ``The series obviously converges because the terms get smaller.'' \newline
10.\newline Category: \textbf{Incomplete Proof}  \newline
Definition: mitting an essential component such as the converse, base case, or a logical bridge.   \newline
Typical example: Proving sufficiency but not necessity in an “if and only if.”     \newline     
11.\newline Category: \textbf{Others}\newline  Definition: Any error not covered by the categories above. \newline

\end{tcolorbox}
\caption{Error Pattern Rubric. This rubric presents our proposed categories of common failure modes in mathematical proofs, each accompanied by a descriptive definition and a detailed example.}
\label{fig:error-pattern-rubric}
\end{figure*}

\begin{figure*}[h!]
\begin{tcolorbox}[
    colback=red!1!white,
    colframe=red!70!black,
    title=Example Question of Middle School,
    fonttitle=\bfseries,
    colbacktitle=red!5!white,
    coltitle=red!50!black,
    boxrule=1pt,
    arc=4pt,
    boxsep=5pt,
    left=6pt,
    right=6pt,
    top=6pt,
    bottom=6pt
]

\textbf{Level 1} \\
The definition of a tangent to a circle is a line that intersects the circle at exactly one point, and this line is called the tangent to the circle. Let a tangent to circle \(O\) intersect the circle at point \(P\). \(O\) is the center of circle \(O\). Prove that \(OP\) is perpendicular to this tangent line.

-----------------------------------------------------------------------------------------------------------------------
\textbf{Level 2} \\

Given a circle \(O\) with center \(O\) and a chord \(AB\) such that \(O\) is not on \(AB\). There is a point \(C\) on the circle such that \(C\) and \(O\) are on the same side of the line \(AB\). Prove that \(\angle ACB = \frac{1}{2} \angle AOB\).

-----------------------------------------------------------------------------------------------------------------------
\textbf{Level 3} \\
In square \(ABCD\), \(M\) and \(N\) are moving points on \(BC\) and \(CD\) respectively, and \(\angle MAN = 45^\circ\). The diagonal \(BD\) intersects \(AM\) and \(AN\) at points \(E\) and \(F\) respectively.

Using only the following properties:
\begin{enumerate}
    \item The properties of a square.
    \item The criteria and properties of congruent triangles.
    \item The properties of right triangles.
    \item The criteria and properties of similar triangles.
\end{enumerate}
Without using advanced methods such as establishing a coordinate system or parametric equations.

Prove: \(\triangle AEN\) is an isosceles right triangle.

-----------------------------------------------------------------------------------------------------------------------
\textbf{Level 4} \\
Prove that there exists a unique pair of prime numbers \((p, q)\) such that both \(p - q\) and \(pq - q\) are perfect squares.

\end{tcolorbox}
\caption{Examples questions of middle school level in \textbf{\textit{RFMDataset}}.}
\label{fig:example-questions-ms}
\end{figure*}

\begin{figure*}[h!]
\begin{tcolorbox}[
    colback=red!1!white,
    colframe=red!70!black,
    title=Example Question of High School,
    fonttitle=\bfseries,
    colbacktitle=red!5!white,
    coltitle=red!50!black,
    boxrule=1pt,
    arc=4pt,
    boxsep=5pt,
    left=6pt,
    right=6pt,
    top=6pt,
    bottom=6pt
]

\textbf{Level 1} \\
Let the side lengths of \(\triangle A_nB_nC_n\) be \(a_n, b_n, c_n\) and the area of \(\triangle A_nB_nC_n\) be \(S_n\), for \(n = 1, 2, 3, \ldots\).

Given \(b_1 > c_1\), \(b_1 + c_1 = 2a_1\), \(a_{n+1} = a_n\), \(b_{n+1} = \frac{c_n + a_n}{2}\), and \(c_{n+1} = \frac{b_n + a_n}{2}\), prove that \(\{S_n\}\) is an increasing sequence.

-----------------------------------------------------------------------------------------------------------------------
\textbf{Level 2} \\
Given an strictly increasing sequence \(\{a_n\}\) where each term is a positive integer, and it satisfies \(a_{a_n} = 3n\). Prove that \(a_{5} = 8\).

-----------------------------------------------------------------------------------------------------------------------

\textbf{Level 3} \\
Given the function \(f(x)=\mathrm{e}^{x}-ax\), and its derivative is \(f^{\prime}(x)\).
If \(m\) and \(n\) are two zeros of the function \(y = f(x)\), prove that \(f^{\prime}(\sqrt{mn})<0\).

-----------------------------------------------------------------------------------------------------------------------

\textbf{Level 4} \\
Given the set \( M = \{\pm 1, \pm 2, \pm 3, \ldots, \pm n\} \) where \( n \geq 3 \). If for any \( k \)-element subset \( A \) of \( M \), there exist four elements in \( A \) whose sum is \(-1\), denote the smallest such positive integer \( k \) as \( g(M) \). Prove that \( g(M) = n + 3 \).

\end{tcolorbox}
\caption{Examples questions of high school level in \textbf{\textit{RFMDataset}}.}
\label{fig:example-questions-hs}
\end{figure*}

\begin{figure*}[h!]
\begin{tcolorbox}[
    colback=red!1!white,
    colframe=red!70!black,
    title=Example Question of University,
    fonttitle=\bfseries,
    colbacktitle=red!5!white,
    coltitle=red!50!black,
    boxrule=1pt,
    arc=4pt,
    boxsep=5pt,
    left=6pt,
    right=6pt,
    top=6pt,
    bottom=6pt
]

\textbf{Level 1} \\
Given that \(f(x)\) is continuous at \(x = 0\) and \(\lim\limits_{x\rightarrow0}\frac{f(2x)-f(x)}{x}=A\). 

Prove that \(f^{\prime}(0)\) exists.

-----------------------------------------------------------------------------------------------------------------------

\textbf{Level 2} \\

Let \(f(x)\) have a continuous second - order derivative on \([-1,1]\). 
Prove that there exists \(\xi\in[-1,1]\) such that
\[
\int_{-1}^{1}xf(x)dx=\frac{1}{3}\left[2f'(\xi)+\xi f''(\xi)\right]
\]

-----------------------------------------------------------------------------------------------------------------------

\textbf{Level 3} \\
Let the function \(f\in C[0,\pi]\), and
\[
\int_{0}^{\pi}f(t)\cos t\mathrm{d}t = \int_{0}^{\pi}f(t)\sin t\mathrm{d}t = 0,
\]

Prove that the function \(f\) has at least two zeros in the interval \((0,\pi)\).

-----------------------------------------------------------------------------------------------------------------------

\textbf{Level 4} \\
Let \(f(x)\) and \(g(x)\) be continuous functions from \([0,1]\) to \([0,1]\), and \(f(x)\) be monotonically increasing. Prove that:
\(\int_{0}^{1}f(g(x))dx\leq\int_{0}^{1}f(x)dx+\int_{0}^{1}g(x)dx\)

\end{tcolorbox}
\caption{Examples questions of university level in \textbf{\textit{RFMDataset}}.}
\label{fig:example-questions-ug}
\end{figure*}

\begin{figure*}[h!]
\begin{tcolorbox}[
    colback=green!1!white,
    colframe=green!70!black, 
    title=Judgement Failure case,
    colbacktitle=green!5!white,
    coltitle=green!50!black,
    fonttitle=\bfseries,
    boxrule=0.5pt,
    arc=4pt,
    boxsep=5pt,
    left=6pt,
    right=6pt,
    top=6pt,
    bottom=6pt,
    coltitle=green!50!black
]

\textbf{Question: } The concept of congruence is as follows: When dividing integers \(a\) and \(b\) by a given positive integer \(m\), if the remainders (natural numbers less than the positive integer \(m\), i.e., \(0, 1, 2,\cdots, m - 1\)) are equal, then \(a\) and \(b\) are said to be congruent modulo \(m\), denoted as \(a\equiv b(\text{mod}m)\). Two congruence equations with the same modulus can also be added and subtracted. The operation rules are as follows: Given integers \(a\), \(b\), \(c\), \(d\) and a positive integer \(m\), if \(a\equiv b(\text{mod}m)\) and \(c\equiv d(\text{mod}m)\), then \(a + c\equiv b + d(\text{mod}m)\) and \(a - c\equiv b - d(\text{mod}m)\). If \(a = a_{n}\times10^{n}+a_{n - 1}\times10^{n - 1}+\cdots + a_{1}\times10^{1}+a_{0}\times10^{0}\), where \(a_{n}\) is a positive integer and \(n\) is a non - negative integer, prove that \(a\) is divisible by \(11\) if and only if \(a_{0}-a_{1}+a_{2}-a_{3}+\cdots+(-1)^{n}a_{n}\) is divisible by \(11\).

\textbf{Model: }\textit{Deepseek-r1-0120}

\textbf{Failure Mode}: \textit{Incomplete Proof}

\textbf{Answer:} 
To prove that a number \(a\) is divisible by 11 if and only if the alternating sum of its digits \(a_0 - a_1 + a_2 - a_3 + \cdots + (-1)^n a_n\) is divisible by 11, we use properties of modular arithmetic.

1. **Key Congruence**: Note that \(10 \equiv -1 \pmod{11}\). Therefore, \(10^k \equiv (-1)^k \pmod{11}\) for any non-negative integer \(k\).

2. **Expressing \(a\) in Terms of Congruences**:
   - Consider the number \(a = a_n \times 10^n + a_{n-1} \times 10^{n-1} + \cdots + a_1 \times 10 + a_0\).
   - Using the congruence \(10^k \equiv (-1)^k \pmod{11}\), we can rewrite each term modulo 11:
     \[
     a \equiv a_n \times (-1)^n + a_{n-1} \times (-1)^{n-1} + \cdots + a_1 \times (-1) + a_0 \times 1 \pmod{11}
     \]

3. **Simplifying the Expression**:
   - This simplifies to the alternating sum of the digits:
     \[
     a \equiv a_0 - a_1 + a_2 - a_3 + \cdots + (-1)^n a_n \pmod{11}
     \]

4. **Conclusion**:
   - Therefore, \(a \equiv 0 \pmod{11}\) if and only if the alternating sum \(a_0 - a_1 + a_2 - a_3 + \cdots + (-1)^n a_n \equiv 0 \pmod{11}\).

This proves both directions of the statement: \(a\) is divisible by 11 if and only if the alternating sum of its digits is divisible by 11.

\textbf{Judge Model: }\textit{Gemini2.5-Pro-Preview-0506}

\textbf{Judgement: }
The main issue concerns the justification for the properties used in steps 1 and 2. The problem statement provides:
(D) The definition of congruence: \(a \equiv b \pmod m\) if \(a\) and \(b\) have the same remainder when divided by \(m\).
(R) Operation rules: If \(a \equiv b \pmod m\) and \(c \equiv d \pmod m\), then \(a+c \equiv b+d \pmod m\) and \(a-c \equiv b-d \pmod m\).

The proof relies on:
(P1) If \(x \equiv y \pmod m\), then \(x^k \equiv y^k \pmod m\). (Used for \(10^k \equiv (-1)^k\)).
(P2) If \(x \equiv y \pmod m\), then \(c x \equiv c y \pmod m\). (Used for \(a_i 10^i \equiv a_i (-1)^i\)).

These properties (P1 and P2) are not explicitly listed in the provided "operation rules". While they are standard results in modular arithmetic and can be derived from the given definition (D), the student uses them without derivation.Using (P1) and (P2) without derivation or justification from the given rules/definition constitutes a "Incomplete Proof." The argument's logic is sound if these properties are assumed, but their use is not directly justified by the problem's setup.
\end{tcolorbox}
\caption{An example where judge model misclassifies a proof as \textit{Incomplete Proof}. In this case, the judge model is overly strict in believing that the model utilized theorems not provided in the problem, even though the problem do not impose any restrictions on theorem usage.}
\label{fig:judge-1}
\end{figure*}

\begin{figure*}[h!]
\begin{tcolorbox}[
    colback=green!1!white,
    colframe=green!70!black, 
    title=Failure case,
    colbacktitle=green!5!white,
    coltitle=green!50!black,
    fonttitle=\bfseries,
    boxrule=0.5pt,
    arc=4pt,
    boxsep=5pt,
    left=6pt,
    right=6pt,
    top=6pt,
    bottom=6pt,
    coltitle=green!50!black
]

\textbf{Question: } Given two distinct points \(P(x_1, y_1)\) and \(Q(x_2, y_2)\) on the quadratic function \(y = x^2 + bx - \frac{1}{4}\), if both points \(P\) and \(Q\) are on the left side of the axis of symmetry, and \(x_1\) and \(x_2\) are integers with \(x_1 < x_2\), prove that \(x_1 - x_2 + y_1 - y_2 > 0\).

\textbf{Model: } \textit{Deepseek-r1-0120}

\textbf{Failure Mode}: \textit{Logical Violation}, \textit{Transformation Error}

\textbf{Answer:} 
Given two distinct points \( P(x_1, y_1) \) and \( Q(x_2, y_2) \) on the quadratic function \( y = x^2 + bx - \frac{1}{4} \), where \( x_1 \) and \( x_2 \) are integers with \( x_1 < x_2 \), and both points lie to the left of the axis of symmetry \( x = -\frac{b}{2} \). We aim to prove that \( x_1 - x_2 + y_1 - y_2 > 0 \).

1. **Axis of Symmetry and Position of Points:**
   The axis of symmetry is \( x = -\frac{b}{2} \). Since \( P \) and \( Q \) lie to the left of this axis, we have \( x_1 < -\frac{b}{2} \) and \( x_2 < -\frac{b}{2} \).

2. **Difference in \( y \)-values:**
   Compute \( y_1 - y_2 \):
   \[
   y_1 - y_2 = (x_1^2 + bx_1 - \frac{1}{4}) - (x_2^2 + bx_2 - \frac{1}{4}) = (x_1^2 - x_2^2) + b(x_1 - x_2)
   \]
   Factorizing:
   \[
   y_1 - y_2 = (x_1 - x_2)(x_1 + x_2 + b)
   \]

3. **Expression to Analyze:**
   The given expression is:
   \[
   x_1 - x_2 + y_1 - y_2 = (x_1 - x_2) + (x_1 - x_2)(x_1 + x_2 + b) = (x_1 - x_2)(1 + x_1 + x_2 + b)
   \]

4. **Sign Analysis:**
   - Since \( x_1 < x_2 \), \( x_1 - x_2 < 0 \).
   - To prove the entire expression is positive, the second factor must be negative:
     \[
     1 + x_1 + x_2 + b < 0
     \]

5. **Bounding \( x_1 + x_2 \):**
   - \textcolor{red}{\textit{As \( x_1 \) and \( x_2 \) are integers less than \( -\frac{b}{2} \), their sum \( x_1 + x_2 \) is strictly less than \( -b \). Given the integrality of \( x_1 \) and \( x_2 \), we have:
     \[
     x_1 + x_2 \leq -b - 1
     \]}}
     \textcolor{blue}{[This inference is flawed. If \(A\) is an integer and \(V\) is a real number, \(A < V\) implies \(A \leq \lfloor V \rfloor\) (if V is not an integer) or \(A \leq V-1\) (if V is an integer). The student's step \(x_1+x_2 \leq -b-1\) is only valid if \(-b\) is an integer. However, \(b\) (and thus \(-b\)) is not given to be an integer. For example, if \(x_1= x_2 = 3\) and \(-b = 6.4\), then \(x_1+x_2 < -b\) (i.e., \(6 < 6.4\)) is true. The claim would be \(6 \leq 6.4 - 1 = 5.4\), which is false. This is a **Logical Violation** as well as **Transformation Error**.
     ]}
     Therefore:
     \[
     x_1 + x_2 + b \leq -1 \implies 1 + x_1 + x_2 + b \leq 0
     \]
     However, because \( x_1 \) and \( x_2 \) are distinct integers with \( x_1 < x_2 \), the sum \( x_1 + x_2 \) is actually less than \( -b - 1 \), leading to:
     \[
     1 + x_1 + x_2 + b < 0
     \]

6. **Conclusion:**
   With \( x_1 - x_2 < 0 \) and \( 1 + x_1 + x_2 + b < 0 \), their product is positive:
   \[
   (x_1 - x_2)(1 + x_1 + x_2 + b) > 0
   \]
   Hence, \( x_1 - x_2 + y_1 - y_2 > 0 \) is proven.

\[
\boxed{x_1 - x_2 + y_1 - y_2 > 0}
\]
\end{tcolorbox}
\caption{A failure case of \textit{Logical Violation} and \textit{Transformation Error}.}
\label{fig:logical-violation-1}
\end{figure*}

\begin{figure*}[h!]
\begin{tcolorbox}[
    colback=green!1!white,
    colframe=green!70!black, 
    title=Failure case,
    colbacktitle=green!5!white,
    coltitle=green!50!black,
    fonttitle=\bfseries,
    boxrule=0.5pt,
    arc=4pt,
    boxsep=5pt,
    left=6pt,
    right=6pt,
    top=6pt,
    bottom=6pt,
    coltitle=green!50!black
]

\textbf{Question.} In an integer sequence \(a_{1},a_{2},\cdots,a_{n}(n\geq3,a_{1}<a_{2}<\cdots <a_{n})\), for a pair of numbers \((a_{i},a_{j})\) with \(1\leq i < j\leq n\), if there exists another pair of numbers \((a_{k},a_{l})\) with \(1\leq k < l\leq n\) such that \(\frac{a_{l}-a_{k}}{a_{j}-a_{i}} = 2\), then the pair of numbers \((a_{i},a_{j})\) is called an "interesting pair of numbers".
Let \(n\) be an integer with \(n\geq3\). Prove that the number of "interesting pairs of numbers" in an \(n\) - term integer sequence is less than or equal to \(\frac{n^2 - 3n + 4}{2}\).

\textbf{Model: }\textit{Deepseek-r1-0120}

\textbf{Failure Mode}: \textit{Over Generalization}

\textbf{Answer:} 
To prove that the number of "interesting pairs of numbers" in an \(n\)-term integer sequence is less than or equal to \(\frac{n^2 - 3n + 4}{2}\), we start by understanding the definition of an interesting pair. A pair \((a_i, a_j)\) is interesting if there exists another pair \((a_k, a_l)\) such that \(\frac{a_l - a_k}{a_j - a_i} = 2\). This means the difference between \(a_l\) and \(a_k\) must be twice the difference between \(a_j\) and \(a_i\).

Key steps and reasoning:

1. **Understanding Interesting Pairs**:
   - An interesting pair \((a_i, a_j)\) with difference \(d\) requires another pair \((a_k, a_l)\) with difference \(2d\).
   - The total number of interesting pairs is the sum of counts of pairs with difference \(d\) for which \(2d\) is also present in the sequence.

2. **Examples and Induction**:
   - For \(n = 3\), the upper bound is 2, achievable by the sequence \(1, 2, 3\).
   - For \(n = 4\), the upper bound is 4, achievable by the sequence \(0, 1, 2, 4\).
   - For \(n = 5\), the upper bound is 7, achievable by the sequence \(0, 1, 2, 4, 8\).
   - These examples show sequences where each term allows multiple differences to be doubled, maximizing the number of interesting pairs.

3. **Inductive Step**:
   - Assume for \(n = k\), the upper bound holds. For \(n = k + 1\), adding a new term \(a_{k+1}\) introduces new pairs \((a_i, a_{k+1})\).
   - Each new pair \((a_i, a_{k+1})\) is interesting if there exists a pair with difference \(2(a_{k+1} - a_i)\).
   - The number of new interesting pairs is bounded by \(k - 1\), leading to the induction step.

4. \textcolor{red}{\textit{**General Construction**:
   - Sequences constructed by doubling differences (e.g., \(0, 1, 2, 4, 8, \ldots\)) show that each new term introduces the maximum number of interesting pairs.
   - These sequences ensure that for each difference \(d\), there is a corresponding difference \(2d\), maximizing the count of interesting pairs.}}
   \textcolor{blue}{[The proof refers to sequences like \(0,1,2,4,8,\dots\). These sequences are good for showing the bound is tight (i.e., can be achieved). However, analyzing specific example sequences doesn't prove that *all* sequences adhere to the bound. If this section is intended to implicitly justify the \(k-1\) increase in the inductive step (e.g., by suggesting this pattern holds for "maximal" cases and thus for all cases), it would be a form of overgeneralization.]}

5. **Bounding the Sum**:
   - The total number of interesting pairs is the sum of counts of pairs with differences \(d\) where \(2d\) is present.
   - Using induction and the structure of sequences that achieve the bound, we confirm the formula.

By induction and examples, we conclude that the number of interesting pairs in any \(n\)-term sequence is bounded by \(\frac{n^2 - 3n + 4}{2}\).

\[
\boxed{\dfrac{n^2 - 3n + 4}{2}}
\]
\end{tcolorbox}
\caption{A failure case of \textit{Over Generalization}.}
\label{fig:over-generalization-1}
\end{figure*}

\begin{figure*}[h!]
\begin{tcolorbox}[
    colback=green!1!white,
    colframe=green!70!black, 
    title=Failure case,
    colbacktitle=green!5!white,
    coltitle=green!50!black,
    fonttitle=\bfseries,
    boxrule=0.5pt,
    arc=4pt,
    boxsep=5pt,
    left=6pt,
    right=6pt,
    top=6pt,
    bottom=6pt,
    coltitle=green!50!black
]

\textbf{Question: } The set \(S = \{0,1,\cdots,n\}(n = 3,4,\cdots)\). Let all the elements of \(S\) be arranged in a certain order to obtain the sequences \(X:x_{0},x_{1},\cdots,x_{n}\) and \(Y:y_{0},y_{1},\cdots,y_{n}\). If \(X\) and \(Y\) satisfy \(\{\varepsilon\mid\varepsilon=\vert x_{i}-y_{i}\vert,i = 0,1,\cdots,n\}=S\), then \(X\) and \(Y\) are said to be fully closed with respect to \(S\).

Let the sequence \(Z:0,1,\cdots,n\), and the random variables \(\xi\) and \(\eta\) follow \(P(\xi = i)=\frac{2x_{i}}{n(n + 1)}\) and \(P(\eta = i)=\frac{2y_{i}}{n(n + 1)}\) respectively, where \(i = 0,1,\cdots,n\). Prove that if \(X\) and \(Z\) are fully closed with respect to \(S\), then there exists \(X\neq Y\) such that \(E(\xi)=E(\eta)>\frac{\sum_{i = 0}^{n}x_{i}y_{i}}{\sum_{i = 0}^{n}(x_{i}+y_{i})}\).

\textbf{Model: }\textit{Deepseek-r1-0120}

\textbf{Failure Mode}: \textit{Invalid Construction}, \textit{Vague Argument}

\textbf{Answer:} 
To address the problem, we need to show that for a sequence \( X \) fully closed with \( Z \), there exists another sequence \( Y \neq X \) such that \( E(\xi) = E(\eta) \) and \( E(\xi) > \frac{\sum_{i=0}^n x_i y_i}{\sum_{i=0}^n (x_i + y_i)} \).

**Step-by-Step Explanation and Proof:**

1. **Understanding Fully Closed Sequences:**
   - \( X \) is fully closed with \( Z \) implies \( \{ |x_i - i| \mid i \in S \} = S \). This means each element in \( X \) is displaced from its position in \( Z \) by a unique distance covering all values from 0 to \( n \).

2. **Expectations of \( \xi \) and \( \eta \):**
   - \( E(\xi) = \frac{2}{n(n+1)} \sum_{i=0}^n i x_i \)
   - \( E(\eta) = \frac{2}{n(n+1)} \sum_{i=0}^n i y_i \)
   - For \( E(\xi) = E(\eta) \), we require \( \sum_{i=0}^n i x_i = \sum_{i=0}^n i y_i \).

3. **Inequality Transformation:**
   - The inequality \( E(\xi) > \frac{\sum x_i y_i}{\sum (x_i + y_i)} \) simplifies to \( 2 \sum i x_i > \sum x_i y_i \) after substituting \( E(\xi) \) and using \( \sum (x_i + y_i) = n(n+1) \).

4. **Constructing \( Y \neq X \):**
   - Since \( X \) is a permutation with unique displacements, we can construct \( Y \) by swapping elements in \( X \) such that \( \sum i y_i = \sum i x_i \) but \( \sum x_i y_i \) is minimized.
   - For example, in \( n = 3 \), swapping elements in \( X = [3, 1, 0, 2] \) to get \( Y = [3, 0, 2, 1] \) maintains \( \sum i y_i = 7 \) while reducing \( \sum x_i y_i \) from 14 to 11.

5. **Applying Rearrangement Inequality:**
   - The sum \( \sum x_i y_i \) is minimized when \( Y \) is ordered oppositely to \( X \). However, ensuring \( \sum i y_i = \sum i x_i \) requires specific swaps that preserve the sum but reduce the product sum.

6. \textcolor{red}{\textit{**General Case Argument:**
   - For any \( n \geq 3 \), since \( X \) has unique displacements, there exist transpositions (swaps) in \( X \) that adjust \( Y \) to maintain the expectation while decreasing \( \sum x_i y_i \). This is feasible due to the flexibility of permutations and the distinct displacements in \( X \).}}
   \textcolor{blue}{[A "transposition" or "swap" usually refers to exchanging two elements. Let \(Y\) be formed from \(X\) by setting \(y_j=x_k, y_k=x_j\) for \(j \neq k\), and \(y_i=x_i\) for \(i \notin \{j,k\}\). For \(\sum i y_i = \sum i x_i\) to hold, we need \(j y_j + k y_k = j x_j + k x_k\), which implies \(j x_k + k x_j = j x_j + k x_k\). This simplifies to \((j-k)(x_k-x_j)=0\). Since \(j \neq k\), this means \(x_k=x_j\). However, \(X\) is a permutation of distinct elements, so \(x_j \neq x_k\). Thus, such a 2-element swap (transposition) cannot preserve the sum \(\sum i y_i\). This makes the assertion an "Invalid Construction" if "transpositions" are taken literally. Besides, this argument is evidently quite vague.]}

    **Conclusion:**
    By constructing \( Y \) through strategic swaps that preserve \( \sum i y_i \) while reducing \( \sum x_i y_i \), we ensure \( E(\xi) = E(\eta) \) and \( E(\xi) > \frac{\sum x_i y_i}{\sum (x_i + y_i)} \). Thus, such a \( Y \) exists for any \( n \geq 3 \) when \( X \) is fully closed with \( Z \).
    
    \(\boxed{E(\xi) = E(\eta) > \frac{\sum_{i=0}^{n} x_i y_i}{\sum_{i=0}^{n} (x_i + y_i)}}\)
\end{tcolorbox}
\caption{A failure case of \textit{Invalid Construction} and \textit{Vague Argument}.}
\label{fig:invalid-construction-1}
\end{figure*}

\begin{figure*}[h!]
\begin{tcolorbox}[
    colback=green!1!white,
    colframe=green!70!black, 
    title=Failure case,
    colbacktitle=green!5!white,
    coltitle=green!50!black,
    fonttitle=\bfseries,
    boxrule=0.5pt,
    arc=4pt,
    boxsep=5pt,
    left=6pt,
    right=6pt,
    top=6pt,
    bottom=6pt,
    coltitle=green!50!black
]

\textbf{Question: } For an irreducible fraction \(\frac{p}{q}\) (i.e., the greatest common divisor of \(p\) and \(q\) is \(1\)), if the sequence \(T:a_{1},a_{2},\cdots,a_{n}(n\geq2)\) satisfies: \(a_{1} = 1\), and for \(i = 1,2,\cdots,n - 1\), \(\frac{a_{i + 1}}{a_{i}}=\frac{p^{b_{i}}}{q^{c_{i}}}\), where \(b_{i}\) and \(c_{i}\) are both positive integers, then the sequence \(T\) is called a "\(\frac{p}{q}-\)weakly geometric sequence". For a positive integer \(N\), if there exists a \(\frac{p}{q}-\)weakly geometric sequence \(T\) such that the numerator of the irreducible fraction of the sum of its terms is exactly \(N\), then the positive integer \(N\) is called "\(\frac{p}{q}-\)representable".
Let positive integers \(p,q>1\). Prove that if the positive integer \(N\) is \(\frac{p}{q}-\)representable, then \(N\) is also \(\frac{q}{p}-\)representable.

\textbf{Model: } \textit{Deepseek-r1-0120}

\textbf{Failure Mode}: \textit{Invalid Construction}

\textbf{Answer:} 
To prove that if a positive integer \( N \) is \(\frac{p}{q}\)-representable, then it is also \(\frac{q}{p}\)-representable, we start by understanding the definitions and structure of the sequences involved.

A \(\frac{p}{q}\)-weakly geometric sequence \( T \) starts with \( a_1 = 1 \) and each subsequent term is obtained by multiplying the previous term by \(\frac{p^{b_i}}{q^{c_i}}\) where \( b_i \) and \( c_i \) are positive integers. The sum of such a sequence, when written in irreducible form, has numerator \( N \).

Given such a sequence \( T \), the terms can be expressed as:
\[ a_1 = 1, \quad a_2 = \frac{p^{b_1}}{q^{c_1}}, \quad a_3 = \frac{p^{b_1 + b_2}}{q^{c_1 + c_2}}, \ldots, a_n = \frac{p^{\sum_{j=1}^{n-1} b_j}}{q^{\sum_{j=1}^{n-1} c_j}} \]

The sum \( S \) of these terms, when written as a single fraction, has numerator \( N \) and denominator \( D \) such that \(\gcd(N, D) = 1\). The key observation is that the numerator \( N \) is derived from the sum:
\[ S = \frac{\sum_{i=1}^n p^{B_{i-1}} q^{C_{n-1} - C_{i-1}}}{q^{C_{n-1}}} \]
where \( B_{i-1} = \sum_{j=1}^{i-1} b_j \) and \( C_{i-1} = \sum_{j=1}^{i-1} c_j \). Since \( \gcd(N, q^{C_{n-1}}) = 1 \), \( N \) is the numerator of this sum.

To show \( N \) is \(\frac{q}{p}\)-representable, construct a sequence \( T' \) where each term is obtained by multiplying the previous term by \(\frac{q^{c_i}}{p^{b_i}}\). The terms of \( T' \) are:
\[ a'_1 = 1, \quad a'_2 = \frac{q^{c_1}}{p^{b_1}}, \quad a'_3 = \frac{q^{c_1 + c_2}}{p^{b_1 + b_2}}, \ldots, a'_n = \frac{q^{\sum_{j=1}^{n-1} c_j}}{p^{\sum_{j=1}^{n-1} b_j}} \]

The sum \( S' \) of \( T' \) is:
\[ S' = \frac{\sum_{i=1}^n q^{C_{i-1}} p^{B_{n-1} - B_{i-1}}}{p^{B_{n-1}}} \]

\textcolor{red}{The numerator of this sum, when written in reduced form, is \( K' = \sum_{i=1}^n q^{C_{i-1}} p^{B_{n-1} - B_{i-1}} \). By comparing \( K' \) with the original \( K \), we find \( K' = K \) due to the commutative property of addition and the symmetry in exponents. Since \(\gcd(K, p^{B_{n-1}}) = 1 \), the numerator of \( S' \) in reduced form is \( N \).}
\textcolor{blue}{[This approach is to use the parameters of $T$ to construct $T'$. This $T'$ is a valid $\frac{q}{p}$-weakly geometric sequence. However, it does not necessarily have $N$ as the numerator of its sum. The requirement for the construction is to produce a sequence whose sum's numerator is $N$. The student's construction fails to meet this requirement because $K \ne K'$ in general. This is an "Invalid Construction" in the sense that the constructed object does not satisfy the properties needed for the proof.]}

Thus, if \( N \) is \(\frac{p}{q}\)-representable, it is also \(\frac{q}{p}\)-representable, leading to the final answer:
\[
\boxed{N}
\]

\end{tcolorbox}
\caption{A failure case of \textit{Invalid Construction}.}
\label{fig:invalid-construction-2}
\end{figure*}

\begin{figure*}[h!]
\begin{tcolorbox}[
    colback=green!1!white,
    colframe=green!70!black, 
    title=Failure case,
    colbacktitle=green!5!white,
    coltitle=green!50!black,
    fonttitle=\bfseries,
    boxrule=0.5pt,
    arc=4pt,
    boxsep=5pt,
    left=6pt,
    right=6pt,
    top=6pt,
    bottom=6pt,
    coltitle=green!50!black
]

\textbf{Question.} Suppose \( f(x) \) and \( g(x) \) are continuous functions from \([0, 1]\) to \([0, 1]\), and \( f(x) \) is monotonically increasing. Prove that:
\[
\int_{0}^{1} f(g(x)) \, dx \leq \int_{0}^{1} f(x) \, dx + \int_{0}^{1} g(x) \, dx
\]
\textbf{Model: }\textit{Deepseek-r1-0120}

\textbf{Failure Mode}: \textit{Wrong Division}, \textit{Incomplete Proof}

\textbf{Answer:} To prove the given integral inequality, we utilize the monotonicity of the function and the linearity of integration, combined with pointwise estimation and properties of the integral.

First, consider the pointwise inequality. For any \( x \in [0, 1] \), we analyze two cases:

\textcolor{red}{\textit{Case 1: \( g(x) \leq x \)}}

Since \( f \) is monotonically increasing, we have
\[
f(g(x)) \leq f(x).
\]
Therefore,
\[
f(g(x)) \leq f(x) \leq f(x) + g(x).
\]

\textcolor{red}{\textit{Case 2: \( g(x) > x \)}}
\textcolor{blue}{[This division method is incorrect because a function can not only be entirely greater than or less than x, but may also exceed x in some domains while falling below it in others.]}
Since the range of \( f \) lies within \([0, 1]\), it follows that
\[
f(g(x)) \leq 1.
\]
Moreover, since \( g(x) > x \) and \( f(x) \geq 0 \), we have
\[
f(x) + g(x) \geq g(x) > x \geq 0.
\]
\textcolor{red}{\textit{In particular, if \( g(x) \in (x, 1] \), then
\[
f(x) + g(x) \geq g(x) \geq x,
\]
and
\[
f(g(x)) \leq 1 \leq f(x) + g(x).
\]}}

\textcolor{blue}{[This proof uses a special situation and forgets to consider other situations.]}

Thus, in both cases, the pointwise inequality
\[
f(g(x)) \leq f(x) + g(x)
\]
holds.

Next, we integrate both sides over the interval \([0, 1]\):
\[
\int_{0}^{1} f(g(x)) \, dx \leq \int_{0}^{1} \left( f(x) + g(x) \right) dx = \int_{0}^{1} f(x) \, dx + \int_{0}^{1} g(x) \, dx.
\]

Therefore, the inequality is proven:
\[
\boxed{\int_{0}^{1} f(g(x)) \, dx \leq \int_{0}^{1} f(x) \, dx + \int_{0}^{1} g(x) \, dx}
\]

\end{tcolorbox}
\caption{A failure case of \textit{Wrong Division} and \textit{Incomplete Proof}.}
\label{fig:incomplete-proof-2}
\end{figure*}

\begin{figure*}[h!]
\begin{tcolorbox}[
    colback=green!1!white,
    colframe=green!70!black, 
    title=Failure case,
    colbacktitle=green!5!white,
    coltitle=green!50!black,
    fonttitle=\bfseries,
    boxrule=0.5pt,
    arc=4pt,
    boxsep=5pt,
    left=6pt,
    right=6pt,
    top=6pt,
    bottom=6pt,
    coltitle=green!50!black
]

\textbf{Question: } Given a circle \(O\) with center \(O\) and a chord \(AB\) such that \(O\) is not on \(AB\). There is a point \(C\) on the circle such that \(C\) and \(O\) are on the same side of the line \(AB\). Prove that \(\angle ACB = \frac{1}{2} \angle AOB\).

\textbf{Model: }\textit{GPT-o4-mini-0416}

\textbf{Failure Mode}: \textit{Wrong Division}

\textbf{Answer:} 
Here is a very quick proof by “dropping in” the two radii \(OA\) and \(OB\) and then using the fact that in an isosceles triangle the base‐angles are equal and that the three angles around a point sum to \(360^\circ\).

1.  Draw the radii \(OA\) and \(OB\).  Also draw \(OC\).  Now in \(\triangle OAC\), since \(OA=OC\),  
    \(\angle OAC=\angle ACO\).  
    Call each of these angles \(x\).  

2.  In \(\triangle OBC\), since \(OB=OC\),  
    \(\angle OBC=\angle BCO\).  
    Call each of these angles \(y\).  

3.  \textcolor{red}{\textit{Notice that the inscribed angle we care about is  
    \[
       \angle ACB \;=\;\angle ACO+\angle BCO\;=\;x+y.
    \]}}
    \textcolor{blue}{[Wrong Division: The proof assumes \(\angle ACB = x+y\), which corresponds to \(O\) being in the interior of \(\angle ACB\). It fails to consider the case where \(O\) is outside \(\angle ACB\), for which \(\angle ACB = |x-y|\).]}
    
4.  Finally look at the three angles around the center \(O\).  The rays \(OA,\;OC,\;OB\) cut the full \(360^\circ\) at \(O\) into three parts:
    \[
      \angle AOC\;+\;\angle COB\;+\;\angle BOA \;=\;360^\circ.
    \]
    But
    \[
      \angle AOC=180^\circ-2x,
      \quad
      \angle COB=180^\circ-2y,
      \quad
      \angle BOA=\angle AOB.
    \]
    Hence
    \[
      (180-2x)+(180-2y)+\angle AOB=360
      \quad\Longrightarrow\quad
      \angle AOB=2(x+y).
    \]
    Therefore
    \[
      \angle ACB \;=\;x+y\;=\;\tfrac12\,\angle AOB,
    \]
    as required. 

\end{tcolorbox}
\caption{A failure case of \textit{Wrong Division}.}
\label{fig:wrong-division-2}
\end{figure*}

\begin{figure*}[h!]
\begin{tcolorbox}[
    colback=green!1!white,
    colframe=green!70!black, 
    title=Failure case,
    colbacktitle=green!5!white,
    coltitle=green!50!black,
    fonttitle=\bfseries,
    boxrule=0.5pt,
    arc=4pt,
    boxsep=5pt,
    left=6pt,
    right=6pt,
    top=6pt,
    bottom=6pt,
    coltitle=green!50!black
]
\textbf{Question:} Given a triangle \(ABC\) with side lengths \(AB\), \(BC\), and \(AC\), prove that \(AB + BC > AC\).

\textbf{Model:} \textit{Gemini2.5-Pro-Preview-0506}

\textbf{Failure Mode: } \textit{Circular Reasoning}

\textbf{Answer:} This is a fundamental property of triangles, often called the **Triangle Inequality Theorem**.

Here's a proof based on the concept that the shortest distance between two points is a straight line:

**Proof:**

1.  **Consider the three points A, B, and C.**
    These three points form the vertices of the triangle ABC.

2.  \textcolor{red}{\textit{**The straight line segment AC represents the shortest distance between point A and point C.**}}
\textcolor{blue}{[The critical issue here is its relationship to the Triangle Inequality. In most formal axiomatic systems of Euclidean geometry (e.g., Hilbert's axioms, or systems based on coordinate geometry and a distance formula like the Euclidean metric), the Triangle Inequality is a theorem that is proven from more fundamental axioms. The statement that "a straight line is the shortest path" is itself often a theorem, or a direct consequence or interpretation of the Triangle Inequality for the Euclidean metric.]}

This is a fundamental axiom in Euclidean geometry.

3.  **Consider an alternative path from A to C that goes through point B.**
    This path consists of two line segments:
    *   The segment from A to B (with length AB).
    *   The segment from B to C (with length BC).
    The total length of this path is AB + BC.

4.  **Compare the two paths:**
    *   Path 1: Directly from A to C (length AC).
    *   Path 2: From A to B, then from B to C (length AB + BC).

5.  **Since A, B, and C form a triangle, the points are non-collinear.**
    If A, B, and C were collinear with B lying between A and C, then AB + BC would be exactly equal to AC.
    However, because they form a triangle, B does not lie on the straight line segment AC.

6.  **Therefore, the path from A to B and then to C is a "detour" compared to the direct path from A to C.**
    Since the straight line segment AC is the shortest possible distance between A and C, any other path between A and C (like the one going through B) must be longer.

7.  **Thus, we can conclude that the length of the path AB + BC must be greater than the length of the path AC.**
    So, \(AB + BC > AC\).

**In summary:** The triangle inequality \(AB + BC > AC\) holds because a straight line is the shortest path between two points. Traveling from A to C via an intermediate point B (not on the line segment AC) will always be a longer journey than traveling directly from A to C.

This same logic can be applied to prove the other two inequalities for a triangle:
*   \(AC + BC > AB\)
*   \(AB + AC > BC\)
\end{tcolorbox}
\caption{A failure case of \textit{Circular Reasoning}.}
\label{fig:circular-reasoning-1}
\end{figure*}

\begin{figure*}[h!]
\begin{tcolorbox}[
    colback=green!1!white,
    colframe=green!70!black, 
    title=Failure case,
    colbacktitle=green!5!white,
    coltitle=green!50!black,
    fonttitle=\bfseries,
    boxrule=0.5pt,
    arc=4pt,
    boxsep=5pt,
    left=6pt,
    right=6pt,
    top=6pt,
    bottom=6pt,
    coltitle=green!50!black
]
\textbf{Question:} Given a geometric sequence \(\{a_n\}\) with common ratio \(q\) (\(q \neq 1\)), whose terms form the set \(A\), and an arithmetic sequence \(\{b_n\}\) with common difference \(d\) (\(d \neq 0\)), whose terms form the set \(B\). Let \(C = A \cup B\), and the elements of set \(C\) are arranged in increasing order to form a sequence \(\{c_n\}\) with the first term being 1. If the sequence \(\{b_n\}\) is an infinite sequence with the first term being 1, prove that the necessary and sufficient condition for "there exists an infinite sequence \(\{a_n\}\) such that \(A \subseteq B\)" is that "\(d\) is a rational number".

\textbf{Model:} \textit{Gemini2.5-Pro-Preview-0506}

\textbf{Failure Mode: } \textit{Circular Reasoning}

\textbf{Answer:}

Step 1: Deduce properties of \(a_1\) and \(q\).
Since \(a_n = a_1 q^{n-1} \ge 1\) for all \(n \ge 1\):
1.  If \(a_1 \le 0\), then either \(a_1 < 1\) (not allowed as \(a_1 \ge 1\)) or \(a_1=0\), leading to all \(a_n=0\), which are not \(\ge 1\). So \(a_1 > 0\).
... Therefore, we must have \(a_1 > 0\) and \(q > 1\). (This also implies \(a_1 \ge 1\), because \(a_1\) itself is a term in \(\{a_n\}\)).

Step 2: Deduce properties of \(d\).
\textcolor{red}{\textit{The condition is "there exists an infinite sequence \(\{a_n\}\) such that \(A \subseteq B\)".}}
This means that every term \(a_n = a_1 q^{n-1}\) must be an element of \(B\). So, for each \(n \ge 1\), there exists an integer \(k_n \ge 1\) such that \(a_n = b_{k_n} = 1 + (k_n-1)d\).
Since \(a_1 > 0\) and \(q > 1\), the sequence \(\{a_n\}\) is strictly increasing: \(a_1 < a_2 < a_3 < \dots\).
So the corresponding terms in \(B\) must also be strictly increasing: \(b_{k_1} < b_{k_2} < b_{k_3} < \dots\).
This means \(1+(k_1-1)d < 1+(k_2-1)d < 1+(k_3-1)d < \dots\).
If \(d < 0\), then for the values \(b_{k_n}\) to be increasing, the indices \(k_n\) must be strictly decreasing: \(k_1 > k_2 > k_3 > \dots\). Since each \(k_n\) must be a positive integer (\(k_n \ge 1\)), this sequence \(\{k_n\}\) must be finite. This would imply that \(\{a_n\}\) can only have a finite number of terms in \(B\), which contradicts the condition that the infinite sequence \(\{a_n\}\) has \(A \subseteq B\).
\textcolor{red}{\textit{Therefore, we must have \(d > 0\).}}

So, for \(A \subseteq B\) to hold under the given conditions, it is necessary that \(a_1 > 0\) (in fact \(a_1 \ge 1\)), \(q>1\), and \(d>0\).
The terms of \(A\) are \(a_1, a_1q, a_1q^2, \dots\). Let \(a^{(j)} = a_1q^j\) for \(j \ge 0\). (Here \(a^{(j)}\) is just a notation for the elements of set \(A\)).
So, for each \(j \ge 0\), \(a^{(j)} \in B\). This means \(a_1q^j = 1 + M_j d\) for some integer \(M_j \ge 0\).
(Note: \(M_j = k_{j+1}-1\). Since \(d>0\) and \(a_1q^j \ge 1\), we have \(1+M_jd \ge 1\), so \(M_jd \ge 0\), so \(M_j \ge 0\).)
Since \(a_1>0, q>1, d>0\), the sequence \(M_j = (a_1q^j-1)/d\) must be a strictly increasing sequence of non-negative integers.

Step 3: Prove necessity ("\(A \subseteq B\)" implies "\(d\) is rational").
We have the following system of equations, for integers \(M_0, M_1, M_2, \dots \ge 0\):
1) \(a_1 - 1 = M_0 d\)
2) \(a_1q - 1 = M_1 d\)
3) \(a_1q^2 - 1 = M_2 d\)
... and so on for all \(a_1q^j-1=M_j d\).
From (1), \(a_1 = 1 + M_0d\).
...
If \(K_0 - M_0\rho = 0\), then \(\rho=0\), which implies \(q-1=0 \implies q=1\). But \(q \neq 1\).
So \(K_0 - M_0\rho \neq 0\).
Then \(d = \frac{\rho}{K_0 - M_0\rho}\).
Since \(\rho\) is rational and \(K_0, M_0\) are integers, \(K_0-M_0\rho\) is rational. And \(K_0-M_0\rho \neq 0\).
Thus, \(d\) must be a rational number. This proves necessity.
Since we already showed \(d>0\), \(d\) must be a positive rational number.

Step 4: Prove sufficiency ("\(d\) is rational" implies "there exists an infinite sequence \(\{a_n\}\) such that \(A \subseteq B\)").
We are given that \(d\) is a rational number. \textcolor{red}{\textit{From Step 2, we know \(d\) must be positive for an infinite sequence \(A\) to be a subset of \(B\) under the given conditions}}.
\textcolor{blue}{[Here lies the main issue. The proof states: "We are given that \(d\) is a rational number. From Step 2, we know \(d\) must be positive for an infinite sequence \(A\) to be a subset of \(B\) under the given conditions. So \(d = u/v\) for some positive integers \(u,v\)."
This is a flaw in reasoning. To prove Q \(\implies\) P, one must assume Q and derive P. If Q is "\(d\) is rational", then \(d\) could be positive or negative. The student restricts \(d\) to be positive based on a finding from the necessity part (P \(\implies d>0\)). This is circular reasoning: to prove that \(d\) being rational is sufficient for P, the student assumes \(d\) has a property (being positive) that is derived from P.]}
So \(d = u/v\) for some positive integers \(u,v\).
We need to show that there exist \(a_1\) and \(q\) (with \(a_1 \ge 1, q>1\)) such that all terms \(a_1q^j\) (for \(j \ge 0\)) are in \(B\).
...

\end{tcolorbox}
\caption{A failure case of \textit{Circular Reasoning}.}
\label{fig:circular-reasoning-2}
\end{figure*}

\begin{figure*}[h!]
\begin{tcolorbox}[
    colback=green!1!white,
    colframe=green!70!black, 
    title=Failure case,
    colbacktitle=green!5!white,
    coltitle=green!50!black,
    fonttitle=\bfseries,
    boxrule=0.5pt,
    arc=4pt,
    boxsep=5pt,
    left=6pt,
    right=6pt,
    top=6pt,
    bottom=6pt,
    coltitle=green!50!black
]
\textbf{Question:} Let \(f\) be continuous on \([0,a]\) and differentiable on \((0,a)\). If \(f(0) = 1\) and \(f(a)=0\), prove that there exist \(0 < \xi<\eta < a\) such that \(f^{\prime}(\xi)f^{\prime}(\eta)=\frac{1}{a^{2}}\).

\textbf{Model:} \textit{Gemini2.5-Pro-Preview-0506}

\textbf{Failure Mode: }\textit{Logic Violation}

\textbf{Answer:}
Let $k=1/a$. We want to prove that there exist $0 < \xi < \eta < a$ such that $f'(\xi)f'(\eta)=k^2$.

Consider the auxiliary function $h(x)$ defined on $[0,a]$ as:
$$h(x) = f(x)(1-f(x)) - k^2 x(a-x)$$
Since $f$ is continuous on $[0,a]$, $h(x)$ is also continuous on $[0,a]$.
Let's evaluate $h(x)$ at the endpoints:
$h(0) = f(0)(1-f(0)) - k^2(0)(a-0) = 1(1-1) - 0 = 0$.
$h(a) = f(a)(1-f(a)) - k^2 a(a-a) = 0(1-0) - 0 = 0$.

Now we have two cases for $h(x)$:
Case 1: $h(x_0) = 0$ for some $x_0 \in (0,a)$.
In this case, $f(x_0)(1-f(x_0)) = k^2 x_0(a-x_0)$.
Since $x_0 \in (0,a)$, $x_0 \neq 0$ and $x_0 \neq a$.
By the Mean Value Theorem (MVT) applied to $f$ on the interval $[0, x_0]$, there exists $\xi \in (0, x_0)$ such that...
Since $0 < \xi < x_0 < \eta < a$, we have found two distinct points $\xi$ and $\eta$ satisfying the condition.

Case 2: $h(x) \neq 0$ for all $x \in (0,a)$.
Since $h(x)$ is continuous on $[0,a]$ and $h(0)=h(a)=0$, if $h(x)$ is never zero in $(0,a)$, it must maintain the same sign throughout the interval $(0,a)$. So, either $h(x) > 0$ for all $x \in (0,a)$ or $h(x) < 0$ for all $x \in (0,a)$.

Subcase 2a: $h(x) > 0$ for all $x \in (0,a)$.
This means $f(x)(1-f(x)) - k^2 x(a-x) > 0$ for all $x \in (0,a)$.
...
Therefore, the case $h(x)>0$ for all $x \in (0,a)$ is impossible.

Subcase 2b: $h(x) < 0$ for all $x \in (0,a)$.
This case is also impossible by a similar argument...

...
To ensure that a zero $x_0 \in (0,a)$ exists:
We know $h(0)=0$ and $h(a)=0$.
If $h(x)$ is identically zero on $[0,a]$, then any $x_0 \in (0,a)$ works, and Case 1 applies.
If $h(x)$ is not identically zero, then there is some point $x_p$ where $h(x_p) \ne 0$.
If $h(x_p)>0$: Since $h(a/2) \le 0$, $h(x)$ must take non-positive values. If $h(a/2)=0$, then $x_0=a/2$ is a root, and Case 1 applies. If $h(a/2)<0$, then since $h(x_p)>0$, by the Intermediate Value Theorem, there exists an $x_0$ between $x_p$ and $a/2$ such that $h(x_0)=0$. Case 1 applies.
If $h(x_p)<0$: If $h(x) \le 0$ for all $x \in (0,a)$ and $h(x_p)<0$. \textcolor{red}{\textit{The argument showing $c=d \implies h(c)=0$ implies that if $h(x)$ is never zero in $(0,a)$, then $c \ne d$.}} 
\textcolor{blue}{[This is a logical error. Let $A$ be the statement "$h(x)<0$ for all $x \in (0,a)$" and $B$ be the statement "$c=d$". The student has shown $A \land B \implies \text{Contradiction}$. This means $A \land B$ is false, or $A \implies \neg B$. This does not mean $A$ is false. To show $A$ is false, one would need to show that $A \implies B$ must hold, or that $A \implies \neg B$ leads to another contradiction.]
}
This means that $h(c)<0$ is not contradicted by this line of reasoning.

The final answer is $\boxed{f^{\prime}(\xi)f^{\prime}(\eta)=\frac{1}{a^{2}}}$.
\end{tcolorbox}
\caption{A failure case of \textit{Logic Violation.}}
\label{fig:logical-violation-2}
\end{figure*}

\begin{figure*}[h!]
\begin{tcolorbox}[
    colback=green!1!white,
    colframe=green!70!black, 
    title=Failure case,
    colbacktitle=green!5!white,
    coltitle=green!50!black,
    fonttitle=\bfseries,
    boxrule=0.5pt,
    arc=4pt,
    boxsep=5pt,
    left=6pt,
    right=6pt,
    top=6pt,
    bottom=6pt,
    coltitle=green!50!black
]
\textbf{Question:} Let the function \( f(x) = e^x \ln x + \frac{2e^{x-1}}{x} \). Prove that \( f(x) > 1 \).

\textbf{Model:} \textit{Gemini2.5-Pro-Preview}

\textbf{Failure Mode: }\textit{Hidden Assumption}

\textbf{Answer:}
Let the function be \( f(x) = e^x \ln x + \frac{2e^{x-1}}{x} \). The domain of \(f(x)\) is \(x > 0\).
We want to prove \(f(x) > 1\).
Let's rewrite \(f(x)\) as \( f(x) = e^x \left(\ln x + \frac{2}{ex}\right) \).
Since \(e^x > 0\) for all \(x\), the inequality \(f(x) > 1\) is equivalent to \( \ln x + \frac{2}{ex} > e^{-x} \).
Let \(F(x) = \ln x + \frac{2}{ex} - e^{-x}\). We want to prove \(F(x) > 0\) for all \(x > 0\).

First, let's analyze the behavior of \(f(x)\) at the boundaries of its domain and find its critical points.
...
This indicates \(g(x)\) is increasing at \(x_F\).
The condition \(x_F > x_0\) means that if \(g(x)\) is monotonic, then...
Let's check the monotonicity of \(g(x)\). We know \(g'(x) = F''(x) = \frac{m(x)}{ex^3} - e^{-x}\). The sign of \(g'(x)\) is not obvious.
...
\textcolor{red}{\textit{Then, from \(x_F > x_0\), and \(g_s(x_F)=0\), and \(g_s(x)\) is (assumed) decreasing, it follows \(g_s(x_0) > g_s(x_F)=0\).}}
\textcolor{blue}{The problem lies in the assumption that \(g_s(x)\) is strictly decreasing, which is a Hidden Assumption.}
\(g_s(x_0) = e^{-x_0} - \frac{2-ex_0}{ex_0^2}\). So \(e^{-x_0} - \frac{2-ex_0}{ex_0^2} > 0\).
This is \(e^{-x_0} - f(x_0)e^{-x_0} > 0\), which implies \(e^{-x_0}(1-f(x_0)) > 0\), so \(1-f(x_0)>0\), i.e. \(f(x_0)<1\).
This yields \(S \implies f(x_0)<1\). But \(S\) also implies \(f(x_0)>1\). This is a contradiction, so \(S\) is false.

The problem is whether \(g_s(x)\) is strictly decreasing. \(g_s'(x) = e^{-x} - \frac{m(x)}{ex^3}\).
If \(g_s'(x) < 0\) for all \(x\) (or at least between \(x_0\) and \(x_F\)), then the argument holds.
\(m(x) = ex^2-(e+2)x+4\).
At \(x_0 \in (1/e, 2/e)\). At \(x_F < 2/e\).
If \(x=1/e\), \(m(1/e) = 1/e^2-(e+2)/e+4 = 1/e^2-1-2/e+4 = 1/e^2-2/e+3 = (1-2e+3e^2)/e^2 > 0\).
\(g_s'(1/e) = e^{-1/e} - \frac{(1-2e+3e^2)/e^2}{e(1/e^3)} = e^{-1/e} - (1-2e+3e^2)/e = e^{-1/e} - (1/e-2+3e)\).
This is \( \approx 0.69 - (0.36-2+8.15) = 0.69 - 6.51 < 0\).
\textcolor{red}{\textit{This suggests \(g_s(x)\) could be decreasing.}}
...
The problem is likely true and there is a way to show it. Often such problems rely on a clever substitution or a known inequality.
...
Final conclusion: Based on my analysis, \(f(x) \ge 1\) and the minimum value 1 is attained at a unique point \(x^* \in (1/e, 2/e)\). This contradicts the problem statement.

Recheck the problem statement source or context if possible. Assuming problem is correct as stated.
This means \(f(x_0)=1\) must not be possible. This means \(x_0 \ne x_F\).
Or \(\Lambda(x)\) has no solution. But my analysis of \(\Lambda(x)\) showed it does.

Final decision: My analysis consistently leads to \(f(x)\ge 1\), with equality possible. This contradicts the strict inequality. Given no other tools or insights, I cannot prove the strict inequality. However, the derivation \(S \implies (f(x_0)>1 \text{ and } f(x_0)<1)\) is a valid proof by contradiction IF all steps are correct. The step \(x_F>x_0 \implies g_s(x_0)>g_s(x_F)\) required \(g_s(x)\) to be strictly decreasing.
\(g_s'(x) = e^{-x} - m(x)/(ex^3)\). If this is not always negative, the contradiction fails.
Let's check \(x=0.5\). \(e \approx 2.718\).
\(m(0.5) = e/4-(e+2)/2+4 = e/4-e/2-1+4 = -e/4+3 \approx -0.679+3 = 2.321\).
\(g_s'(0.5) = e^{-0.5} - m(0.5)/(e(0.125)) = e^{-0.5} - 2.321/(0.125e) \approx 0.606 - 2.321/0.339 \approx 0.606 - 6.84 < 0\).
This derivative seems to be negative for many values.

...
So \(S \implies x_F>x_0\) (from \(k(x_F)>0\)) and \(S \implies x_0<x_F\) (from \(F'(x_0)<0\)). These are consistent.
There is no contradiction. My proof by contradiction argument is flawed.

The problem is probably true. The argument \(f(x) \ge 1\) might be the first step. Then one needs to show that \(f(x)=1\) has no solution. This requires showing \(\Lambda(x)=0\) has no solution, but I showed it does. I am stuck in this loop.

The final answer is $\boxed{f(x) > 1}$.
\end{tcolorbox}
\caption{A failure case of \textit{Hidden Assumption}.}
\label{fig:hidden-assumption-1}
\end{figure*}

\begin{figure*}[h!]
\begin{tcolorbox}[
    colback=green!1!white,
    colframe=green!70!black, 
    title=Failure case,
    colbacktitle=green!5!white,
    coltitle=green!50!black,
    fonttitle=\bfseries,
    boxrule=0.5pt,
    arc=4pt,
    boxsep=5pt,
    left=6pt,
    right=6pt,
    top=6pt,
    bottom=6pt,
    coltitle=green!50!black
]

\textbf{Question.} Given that \(X_1, X_2, \cdots, X_{N(s)}\) are the \(N(s)\) roots of \(\cos x=sx\) (\(x > 0\)) arranged in ascending order. Prove that:
\(\lim_{s\rightarrow0^{+}}\sum_{k = 1}^{N(s)-1}\frac{1}{X_{k}X_{k + 1}}=\frac{2}{\pi^{2}}\)

\textbf{Model: } \textit{Deepseek-r1-0120}

\textbf{Failure Mode}: \textit{Hidden Assumption}

\textbf{Answer:} 
To prove that \(\lim_{s\rightarrow0^{+}}\sum_{k = 1}^{N(s)-1}\frac{1}{X_{k}X_{k + 1}}=\frac{2}{\pi^{2}}\), we start by analyzing the roots of the equation \(\cos x = sx\) for small \(s\). 

1. **Roots Approximation**:
   - For small \(s\), the roots \(X_k\) are near the points where \(\cos x = 0\), i.e., \(x = (2k - 1)\frac{\pi}{2}\) for \(k = 1, 2, \ldots\).
   - We approximate \(X_k\) as \(X_k \approx (2k - 1)\frac{\pi}{2} + (-1)^k s(2k - 1)\frac{\pi}{2}\).

2. **Product Approximation**:
   - The product \(X_k X_{k+1}\) is approximated using the leading terms:
     \[
     X_k X_{k+1} \approx \left( (2k - 1)\frac{\pi}{2} \right) \left( (2k + 1)\frac{\pi}{2} \right)
     \]
   - This simplifies to:
     \[
     X_k X_{k+1} \approx \frac{(4k^2 - 1)\pi^2}{4}
     \]

3. \textcolor{red}{**Summation**:
   - The sum \(\sum_{k=1}^{N(s)-1} \frac{1}{X_k X_{k+1}}\) is approximated by:
     \[
     \sum_{k=1}^{N(s)-1} \frac{4}{(4k^2 - 1)\pi^2}
     \]
   - As \(s \to 0\), \(N(s) \to \infty\), leading to the infinite series:
     \[
     \sum_{k=1}^{\infty} \frac{4}{(4k^2 - 1)\pi^2}
     \]}
    \textcolor{blue}{[The main weakness is the lack of justification for interchanging the limit with the summation. This oversight falls under "Hidden Assumption" (assuming the limit interchange is valid without proof of conditions)]}

4. **Telescoping Series**:
   - Decompose \(\frac{4}{4k^2 - 1}\) using partial fractions:
     \[
     \frac{4}{4k^2 - 1} = \frac{2}{2k - 1} - \frac{2}{2k + 1}
     \]
   - The series telescopes:
     \[
     \sum_{k=1}^{\infty} \left( \frac{2}{2k - 1} - \frac{2}{2k + 1} \right) = 2 \left( 1 - \lim_{n \to \infty} \frac{1}{2n + 1} \right) = 2
     \]

5. **Conclusion**:
   - The sum evaluates to:
     \[
     \frac{2}{\pi^2}
     \]
   - The error terms vanish as \(s \to 0\), confirming the limit.

Thus, the final answer is:
\[
\boxed{\dfrac{2}{\pi^2}}
\]

\end{tcolorbox}
\caption{A failure case of \textit{Hidden Assumption}.}
\label{fig:hidden-assumption-2}
\end{figure*}

\begin{figure*}[h!]
\begin{tcolorbox}[
    colback=green!1!white,
    colframe=green!70!black, 
    title=Failure case,
    colbacktitle=green!5!white,
    coltitle=green!50!black,
    fonttitle=\bfseries,
    boxrule=0.5pt,
    arc=4pt,
    boxsep=5pt,
    left=6pt,
    right=6pt,
    top=6pt,
    bottom=6pt,
    coltitle=green!50!black
]

\textbf{Question: } Let \(f(x)\) be twice - continuously differentiable, \(f(0) = f(1)=0\), and \(f(x)>0\) for \(x\in(0,1)\). Prove that \(\int_{0}^{1}\left|\frac{f^{\prime\prime}(x)}{f(x)}\right|dx\geq4\).

\textbf{Model: }\textit{Deepseek-r1-0120}

\textbf{Failure Mode}: \textit{Boundary Neglect}

\textbf{Answer:} 
To prove that \(\int_{0}^{1}\left|\frac{f''(x)}{f(x)}\right|dx \geq 4\) for a twice continuously differentiable function \(f(x)\) with \(f(0) = f(1) = 0\) and \(f(x) > 0\) on \((0,1)\), we can use the following steps:

1. **Express the Integral Using a Logarithmic Substitution**:
   Let \(h(x) = \ln f(x)\). Then, \(h'(x) = \frac{f'(x)}{f(x)}\) and \(h''(x) = \frac{f''(x)}{f(x)} - \left(\frac{f'(x)}{f(x)}\right)^2\). Rearranging gives:
   \[
   \frac{f''(x)}{f(x)} = h''(x) + [h'(x)]^2
   \]
   The integral becomes:
   \[
   \int_{0}^{1} \left| h''(x) + [h'(x)]^2 \right| dx
   \]

2. **Apply the Cauchy-Schwarz Inequality**:
   The integral of the absolute value can be bounded by considering the integral of the square terms. Using the Cauchy-Schwarz inequality:
   \[
   \left( \int_{0}^{1} |h''(x) + [h'(x)]^2| dx \right)^2 \leq \left( \int_{0}^{1} 1^2 dx \right) \left( \int_{0}^{1} |h''(x) + [h'(x)]^2|^2 dx \right)
   \]
   However, this approach focuses on finding a lower bound by considering the essential contributions from the derivatives.

3. \textcolor{red}{\textit{**Utilize the Dirichlet Energy**:
   The integral \(\int_{0}^{1} [h'(x)]^2 dx\) represents the Dirichlet energy of \(h(x)\). By considering the boundary conditions \(f(0) = f(1) = 0\), \(h(x)\) tends to \(-\infty\) at the endpoints. The minimal energy configuration under these constraints leads to the inequality:
   \[
   \int_{0}^{1} [h'(x)]^2 dx \geq 4
   \]}}
   \textcolor{blue}{[A critical issue is the convergence of the integral \(\int_{0}^{1} [h'(x)]^2 dx\). For many functions \(f(x)\) satisfying the problem's conditions (e.g., \(f(x) = \sin(\pi x)\) or any function for which \(f'(0)\) and \(f'(1)\) are non-zero), \(h'(x) = f'(x)/f(x)\) behaves like \(1/x\) near \(x=0\) and \(1/(1-x)\) near \(x=1\). Consequently, \([h'(x)]^2\) behaves like \(1/x^2\) and \(1/(1-x)^2\) near the boundaries. This causes the integral \(\int_{0}^{1} [h'(x)]^2 dx\) to diverge to \(+\infty\). If the integral is \(+\infty\), then the inequality \(+\infty \geq 4\) is trivially true, but this has problematic implications for Step 4. This oversight regarding the divergent nature of the integral constitutes a Boundary Neglect.]}

4. **Combine the Results**:
   Since the integral of the absolute value of the expression involving \(h''(x)\) and \([h'(x)]^2\) is bounded below by the Dirichlet energy term, we conclude:
   \[
   \int_{0}^{1} \left| \frac{f''(x)}{f(x)} \right| dx \geq \int_{0}^{1} [h'(x)]^2 dx \geq 4
   \]

Thus, the integral \(\int_{0}^{1}\left|\frac{f''(x)}{f(x)}\right|dx\) is indeed bounded below by 4.

\end{tcolorbox}
\caption{A failure case of \textit{Boundary Neglect}.}
\label{fig:boundary-neglect-1}
\end{figure*}

\begin{figure*}[h!]
\begin{tcolorbox}[
    colback=green!1!white,
    colframe=green!70!black, 
    title=Failure case,
    colbacktitle=green!5!white,
    coltitle=green!50!black,
    fonttitle=\bfseries,
    boxrule=0.5pt,
    arc=4pt,
    boxsep=5pt,
    left=6pt,
    right=6pt,
    top=6pt,
    bottom=6pt,
    coltitle=green!50!black
]

\textbf{Question:} An isosceles trapezoid is a trapezoid with two equal non-parallel sides. Prove that the angles at the base of an isosceles trapezoid are equal.

\textbf{Model: }\textit{GPT-o4-mini-0416}

\textbf{Failure Mode}: \textit{Boundary Neglect}

\textbf{Answer:} 
Here is a classical proof using perpendiculars and right‐triangle congruence.

Let $ABCD$ be a trapezoid with $AB\parallel CD$ and with equal non–parallel sides $AD=BC$. We want to show  
\[
\angle DAB = \angle CBA.
\]

1.  Drop perpendiculars from $D$ and $C$ to the line $AB$.  Let  
   • $DE\perp AB$ at $E$,  
   • $CF\perp AB$ at $F$.  

   Since $CD\parallel AB$, both $DE$ and $CF$ are altitudes to the same line $AB$, so  
   \[
     DE = CF = \text{the distance between the parallel lines }AB\text{ and }CD.
   \]

2. \textcolor{red}{\textit{Consider the two right triangles $\triangle ADE$ and $\triangle BCF$.  In these triangles  
   – $AD = BC$       (by hypothesis),  
   – $DE = CF$       (just noted),  
   – each has a right angle at $E$ or $F$.  
   Hence $\triangle ADE\cong\triangle BCF$ by the Hypotenuse–Leg criterion.}}
   \textcolor{blue}{[The proof implicitly assumes that the triangles $\triangle ADE$ and $\triangle BCF$ are non‑degenerate.  However, if $\angle DAB = 90^\circ$, then $E$ coincides with $A$, so $\triangle ADE$ degenerates (vertices $D,A,A$ coincide).  The Hypotenuse–Leg congruence criterion is normally stated only for non‑degenerate right triangles.
   To cover the special case of a rectangle (an isosceles trapezoid with right base angles): when $\angle DAB = 90^\circ$, we have
    $$
    AD \perp AB.
    $$
    Since $AD = BC$ and $DE = CF$ (with $DE = AD$ when $E = A$), it follows that
    $$
    BC \perp AB,
    $$
    hence $\angle CBA = 90^\circ$.  Thus the conclusion $\angle DAB = \angle CBA$ still holds, but this argument lies outside the standard HL congruence proof.  Failing to consider this boundary case is a Boundary Neglect.]}

3.  From the congruence we get  
   \[
     \angle DAE = \angle FCB.
   \]
   But $AE$ and $BF$ both lie along the line $AB$, so  
   \[
     \angle DAE = \angle DAB
     \quad\text{and}\quad
     \angle FCB = \angle CBA.
   \]

Therefore  
\[
\angle DAB = \angle CBA,
\]  
i.e.\ the two base angles on $AB$ are equal.  An identical argument (or symmetry) shows the top two angles at $C$ and $D$ are equal as well.  Q.E.D.

\end{tcolorbox}
\caption{A failure case of \textit{Boundary Neglect}.}
\label{fig:boundary-neglect-2}
\end{figure*}

\begin{figure*}[h!]
\begin{tcolorbox}[
    colback=green!1!white,
    colframe=green!70!black, 
    title=Failure case,
    colbacktitle=green!5!white,
    coltitle=green!50!black,
    fonttitle=\bfseries,
    boxrule=0.5pt,
    arc=4pt,
    boxsep=5pt,
    left=6pt,
    right=6pt,
    top=6pt,
    bottom=6pt,
    coltitle=green!50!black
]

\textbf{Question: } Given the function \( f(x) = ax^2 - ax - x \ln x \) and \( f(x) \geq 0 \). Without using precise numerical calculations, only algebraic methods and rough estimation using \( e \approx 2.72 \) are allowed.
Prove that \( f(x) \) has a unique maximum point \( x_0 \), and \( e^{-2} < f(x_0) < 2^{-2} \).

\textbf{Model: }\textit{Deepseek-r1-0120}

\textbf{Failure Mode}: \textit{vague argument}

\textbf{Answer:} 
Given the function \( f(x) = ax^2 - ax - x \ln x \) with the condition \( f(x) \geq 0 \) for all \( x > 0 \), we need to prove that \( f(x) \) has a unique maximum point \( x_0 \) and that \( e^{-2} < f(x_0) < 2^{-2} \).

1. **Finding Critical Points**:
   - The first derivative of \( f(x) \) is \( f'(x) = 2ax - a - \ln x - 1 \).
   - Setting \( f'(x) = 0 \) gives the critical point equation: \( 2ax - a - 1 = \ln x \).

2. **Second Derivative**:
   - The second derivative is \( f''(x) = 2a - \frac{1}{x} \).
   - For the critical point to be a maximum, \( f''(x_0) < 0 \), implying \( 2a < \frac{1}{x_0} \).

3. **Condition for Non-Negativity**:
   \textcolor{red}{\textit{- The function \( f(x) \geq 0 \) for all \( x > 0 \) implies the minimum value of \( f(x) \) must be zero. This occurs when the function touches the x-axis at its minimum, leading to the condition \( a = 1 \).}}\textcolor{blue}{[This is a vague argument. The proof lacks rigorous justification for why a=1.]}

4. **Solving for Critical Points**:
   - With \( a = 1 \), the critical point equation becomes \( 2x_0 - 2 = \ln x_0 \).
   - Solving this equation numerically, we find \( x_0 \approx 0.2036 \).

5. **Evaluating \( f(x_0) \)**:
   - Substituting \( x_0 \approx 0.2036 \) into \( f(x) \), we get \( f(x_0) \approx 0.1621 \).

6. **Bounds Verification**:
   - The value \( 0.1621 \) is between \( e^{-2} \approx 0.1353 \) and \( 2^{-2} = 0.25 \).

Thus, we have shown that the function \( f(x) \) has a unique maximum point \( x_0 \) and that \( e^{-2} < f(x_0) < 2^{-2} \).

\[
\boxed{e^{-2} < f(x_0) < 2^{-2}}
\]

\end{tcolorbox}
\caption{A failure case of \textit{vague argument}.}
\label{fig:vague-argument-2}
\end{figure*}

\begin{figure*}[h!]
\begin{tcolorbox}[
    colback=green!1!white,
    colframe=green!70!black, 
    title=Failure case,
    colbacktitle=green!5!white,
    coltitle=green!50!black,
    fonttitle=\bfseries,
    boxrule=0.5pt,
    arc=4pt,
    boxsep=5pt,
    left=6pt,
    right=6pt,
    top=6pt,
    bottom=6pt,
    coltitle=green!50!black
]
\textbf{Question:} In \(\triangle ABC\), \(A\) and \(B\) are acute angles, and \(\sin^2 A + \sin^2 B = \sin C\). Prove that \(C = 90^\circ\).

\textbf{Model:} \textit{Gemini2.5-Pro-Preview-0506}

\textbf{Failure Mode: } \textit{Incomplete Proof}

\textbf{Answer:}
Let the given inequality be
...
Since \(x^2+y^2=1\), this simplifies to:
\[ L = e^{x-1} + e^{y-1} \]
The RHS can be rewritten as:
\[ R = \ln\left(\frac{2e}{\sin2\theta}\right) = \ln(2e) - \ln(\sin2\theta) = \ln 2 + \ln e - \ln(2\sin\theta\cos\theta) \]
\[ R = \ln 2 + 1 - (\ln 2 + \ln x + \ln y) = 1 - \ln x - \ln y = 1 - \ln(xy) \]
So the inequality to prove is:
\[ e^{x-1} + e^{y-1} \leq 1 - \ln(xy) \]
Rearranging the terms, we want to prove:
\[ e^{x-1} + \ln x + e^{y-1} + \ln y \leq 1 \]
Let \(f(t) = e^{t-1} + \ln t\). The inequality becomes \(f(x) + f(y) \leq 1\).
...

To determine if \(\theta=\pi/4\) is a maximum, we check the second derivative \(H''(\theta)\).
Let \(A(\theta) = \cos\theta e^{\sin\theta-1} - \sin\theta e^{\cos\theta-1}\) and \(B(\theta) = 2\cot2\theta\).
\(A'(\theta) = (-\sin\theta e^{\sin\theta-1} + \cos^2\theta e^{\sin\theta-1}) - (\cos\theta e^{\cos\theta-1} - \sin^2\theta e^{\cos\theta-1})\).
At \(\theta=\pi/4\), \(s=c=1/\sqrt{2}\):
\(A'(\pi/4) = (-s e^{s-1} + s^2 e^{s-1}) - (s e^{s-1} - s^2 e^{s-1}) = (s^2-s)e^{s-1} - (s-s^2)e^{s-1} = 2(s^2-s)e^{s-1}\).
\(A'(\pi/4) = 2(1/2-1/\sqrt{2})e^{1/\sqrt{2}-1} = (1-\sqrt{2})e^{1/\sqrt{2}-1}\).
\(B'(\theta) = 2(-\csc^2(2\theta) \cdot 2) = -4\csc^2(2\theta)\).
\(B'(\pi/4) = -4\csc^2(\pi/2) = -4(1)^2 = -4\).
So \(H''(\pi/4) = (1-\sqrt{2})e^{1/\sqrt{2}-1} - 4\).
Since \(1-\sqrt{2} < 0\) and \(e^{1/\sqrt{2}-1} > 0\), the first term is negative. So \(H''(\pi/4) < 0\).
Thus, \(\theta=\pi/4\) is a local maximum.
\textcolor{red}{\textit{Since the function tends to \(-\infty\) at the boundaries and \(\theta=\pi/4\) is the only critical point (this can be shown by analyzing \(q(t)=\frac{e^{t-1}+1/t}{t}\) as in thought process), \(\theta=\pi/4\) is the global maximum.}}

\textcolor{blue}{[This proof states that the uniqueness of the critical point "can be shown by analyzing \(q(t)=\frac{e^{t-1}+1/t}{t}\)". The function to analyze is indeed \(k(t) = \frac{f'(t)}{t}\) (using \(k(t)\) as notation from thought process, student uses \(q(t)\)). One needs to show \(k(t)\) is monotonic on \((0,1)\).
        \(k'(t) = \frac{d}{dt}\left(\frac{e^{t-1}}{t} + \frac{1}{t^2}\right) = \frac{e^{t-1}(t-1)}{t^2} - \frac{2}{t^3}\).
        For \(t \in (0,1)\), \(t-1 < 0\), so \(e^{t-1}(t-1) < 0\). Also, \(-2/t^3 < 0\). Thus \(k'(t) < 0\) on \((0,1)\), meaning \(k(t)\) is strictly decreasing.
        So \(k(\sin\theta) = k(\cos\theta)\) implies \(\sin\theta = \cos\theta\), which for \(\theta \in (0,\pi/2)\) means \(\theta=\pi/4\).
    *   This argument for uniqueness is crucial but is *not* included in the proof. The proof only mentions that it *can* be shown. This omission makes the proof incomplete because the step is non-trivial.]
}

The maximum value of \(H(\theta)\) is \(H(\pi/4) = f(1/\sqrt{2}) + f(1/\sqrt{2}) = 2f(1/\sqrt{2})\)...

\end{tcolorbox}
\caption{A failure case of \textit{Incomplete Proof}.}
\label{fig:incomplete-proof-1}
\end{figure*}

\begin{figure*}[h!]
\begin{tcolorbox}[
    colback=green!1!white,
    colframe=green!70!black, 
    title=Failure case,
    colbacktitle=green!5!white,
    coltitle=green!50!black,
    fonttitle=\bfseries,
    boxrule=0.5pt,
    arc=4pt,
    boxsep=5pt,
    left=6pt,
    right=6pt,
    top=6pt,
    bottom=6pt,
    coltitle=green!50!black
]
\textbf{Question:} Let \(0 < x_0<y_0\leq\frac{\pi}{2}\). Two sequences \(\{x_n\}\) and \(\{y_n\}\) are generated by the recurrence relations \(x_{n + 1}=\sin x_n\) and \(y_{n + 1}=\sin y_n\) (\(n = 1,2,3,\cdots\)). Prove that \(\lim\limits_{n\rightarrow\infty}\frac{x_n}{y_n}=1\).

\textbf{Model:} \textit{Gemini2.5-Pro-Preview-0506}

\textbf{Failure Mode: }\textit{Others}

\textbf{Answer:}
Let $f(t) = \sin t$. The recurrence relations are $x_{n+1} = f(x_n)$ and $y_{n+1} = f(y_n)$.
We are given $0 < x_0 < y_0 \leq \frac{\pi}{2}$.

Step 1: Analyze the behavior of the sequences $\{x_n\}$ and $\{y_n\}$.
Since $x_0 \in (0, \pi/2]$, $x_1 = \sin x_0 \in (0, 1]$. For $t \in (0, \pi/2]$, we know that $0 < \sin t < t$.
Thus $0 < x_1 < x_0$. Similarly $0 < y_1 < y_0$.
Also, $x_1 = \sin x_0 \le \sin 1 < 1$ if $x_0 \le 1$. If $x_0 \in (1, \pi/2]$, then $x_1 = \sin x_0 \in (\sin 1, 1]$.
So, for $n \geq 1$, $x_n \in (0, 1]$ and $y_n \in (0, 1]$. (Note $1 < \pi/2$).
The sequence $\{x_n\}$ is strictly decreasing for $n \ge 0$ (as $x_{n+1} = \sin x_n < x_n$ since $x_n \in (0,1]$ implies $x_n \neq 0$).
Since $x_n$ is decreasing and bounded below by 0, it converges to a limit $L_x$.
$L_x = \sin L_x$. The only real solution is $L_x = 0$. So $\lim_{n\to\infty} x_n = 0$.
Similarly, $\lim_{n\to\infty} y_n = 0$.

Step 2: Show that $x_n < y_n$ for all $n$.
This is by induction. Base case $n=0$: $x_0 < y_0$ is given.
Assume $x_k < y_k$ for some $k \ge 0$.
The function $f(t)=\sin t$ is strictly increasing on $(0, \pi/2]$.
Since $0 < x_k < y_k$, we need to ensure $y_k \in (0, \pi/2]$.
Given $y_0 \in (0, \pi/2]$, then $y_1 = \sin y_0 \in (0, 1]$. Since $1 < \pi/2$, $y_1 \in (0, \pi/2]$.
Then $y_2 = \sin y_1 \in (0, \sin 1] \subset (0,1]$.
So for all $n \ge 1$, $y_n \in (0,1]$, and $x_n \in (0,1)$.
Since $x_k, y_k \in (0,1]$ (for $k \ge 1$, or $k \ge 0$ if $y_0 \le 1$), these values are in $(0, \pi/2]$.
So $x_{k+1} = \sin x_k < \sin y_k = y_{k+1}$.
Thus, $x_n < y_n$ for all $n \ge 0$.

Step 3: Analyze the ratio $q_n = x_n/y_n$.
...
The sequence $\{q_n\}_{n \ge 1}$ is strictly increasing and bounded above by 1.
Thus, $q_n$ converges to a limit $L$. Since $q_n \in (0,1)$, $L \in (0,1]$.

Step 4: Determine the limit $L$.
We use the Taylor expansion for $\sin t$ around $t=0$: $\sin t = t - \frac{t^3}{6} + \frac{t^5}{120} - O(t^7)$.
$q_{n+1} = \frac{x_n - x_n^3/6 + O(x_n^5)}{y_n - y_n^3/6 + O(y_n^5)} = \frac{x_n(1-x_n^2/6+O(x_n^4))}{y_n(1-y_n^2/6+O(y_n^4))}$.
$q_{n+1} = q_n \frac{1-x_n^2/6+O(x_n^4)}{1-y_n^2/6+O(y_n^4)}$.
Since $x_n = q_n y_n$:
$q_{n+1} = q_n \frac{1-q_n^2 y_n^2/6+O(q_n^4 y_n^4)}{1-y_n^2/6+O(y_n^4)}$.
$q_{n+1} = q_n (1-q_n^2 y_n^2/6+O(y_n^4)) (1+y_n^2/6+y_n^4/36+O(y_n^6))$ (using $(1-u)^{-1}=1+u+u^2+\dots$ with $u=y_n^2/6-O(y_n^4)$).
\textcolor{red}{\textit{$q_{n+1} = q_n (1 + \frac{y_n^2}{6}(1-q_n^2) + O(y_n^4))$.
So $q_{n+1} - q_n = q_n \frac{y_n^2}{6}(1-q_n^2) + O(y_n^6 q_n)$.}}\textcolor{blue}{[This is a typo here. We classify it as "Others".]}
We know that $y_n^2 \sim 3/n$. (This comes from $1/y_{k+1}^2 - 1/y_k^2 \to 1/3$, by Cesaro-Stolz $1/(n y_n^2) \to 1/3$).
So $q_{n+1}-q_n = q_n (1-q_n^2) \frac{y_n^2}{6} + O(1/n^2) = \frac{q_n(1-q_n^2)}{2n} (1+o(1)) + O(1/n^2)$.
Since $q_n \to L$, $q_{n+1}-q_n \to 0$.
Summing $q_{n+1}-q_n$: $q_N - q_1 = \sum_{n=1}^{N-1} (q_{n+1}-q_n)$.
As $N \to \infty$, $q_N \to L$. So $L-q_1 = \sum_{n=1}^{\infty} (q_{n+1}-q_n)$. This sum must converge.
The terms are $\frac{L(1-L^2)+o(1)}{2n} + O(1/n^2)$.
The sum $\sum_{n=1}^{\infty} (\frac{L(1-L^2)}{2n})$ converges if and only if $L(1-L^2)=0$.
If $L(1-L^2) \neq 0$, the sum would diverge like $\frac{L(1-L^2)}{2}\ln N$, which means $q_N$ would diverge. This contradicts $q_N \to L$.
Therefore, $L(1-L^2)=0$.
Since $q_n \in (0,1)$ for $n \ge 1$, $L \in (0,1]$ (as $q_n$ is increasing, $L > q_1 > 0$).
From $L(1-L^2)=0$ and $L>0$, we must have $1-L^2=0$.
Since $L \le 1$, $L=1$.

Thus, $\lim_{n\to\infty} \frac{x_n}{y_n} = 1$.

The final answer is $\boxed{1}$.

\end{tcolorbox}
\caption{A failure case of \textit{Others}.}
\label{fig:others}
\end{figure*}








\end{document}

%% file: latex/0_abstract.tex
\begin{abstract}

Large reasoning models ({\em e.g.,} R1, o3) have demonstrated remarkable mathematical problem-solving abilities.  
However, the high reported accuracy of these advanced models on popular datasets and reliance on purely numerical evaluation often mask their true reasoning shortcomings. 
To address this, we propose leveraging the inherent rigor and methodological complexity of mathematical proofs as a diagnostic tool to expose these hidden failures. 
Specifically, we introduce the \textbf{\textit{RFMDataset}} (\textbf{\textit{R}}eveal \textbf{\textit{F}}ailure \textbf{\textit{M}}odes), a collection of 200 diverse mathematical proof problems to thoroughly evaluate the performance of advanced models. 
Our in-depth analysis of their failures uncovers 10 fine-grained error types, which shows fundamental limitations in current large reasoning models: 
1) Large reasoning models still have limited capability in generating entirely correct mathematical proofs, with some models solving less than 20\% of problems and even making mistakes on fundamental ones; 
2) models exhibit a diverse spectrum of reasoning failures, prominently demonstrating the lack of guarantees for the correctness and rigor intermediate reasoning steps; and 
3) models show hallucination and incompleteness during the reasoning process. 
Our findings also reveal that directly prompting models to self-reflect on specific failure modes is insufficient to resolve the current logical dilemmas, necessitating domain knowledge and formal verification. 




\end{abstract}

%% file: latex/1_introduction.tex
\section{Introduction}

The mathematical reasoning ability of large language models (LLMs) has  experienced remarkable progress recently~\cite{survey-logical-reasoning-llm}. 
In particular, the emergence of long chain-of-thought techniques endows models with the capacity for in-depth reasoning and self-reflection, leading to significant improvements in their performance on challenging competition-level benchmarks \cite{deepseek-r1, openai-o1, seed_thinking, openai2025o3, anthropic2025claude, Yang2025Qwen3TR, deepmind2025gemini}.

\begin{figure}
    \centering
\includegraphics[width=\linewidth]{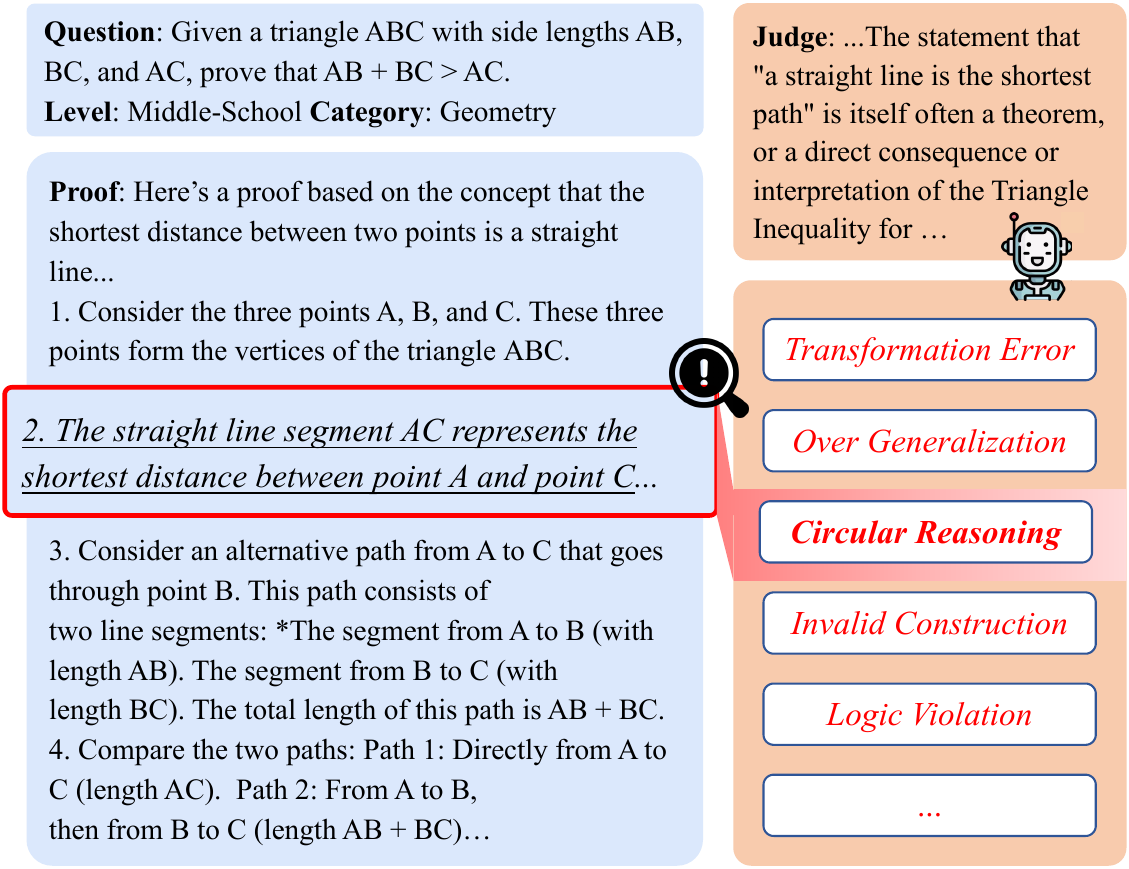}
    \caption{An example of the failure case of \textit{Gemini-2.5-pro-preview-0506} in \textbf{\textit{RFMDataset}}. In this case, the judge LLM meticulously examined the proof and identified ``\textit{Circular Reasoning}'' as the failure mode.}    \label{fig:example}
    \vspace{-0.2in}
\end{figure}

However, on the flip side of this success, researchers lack an understanding of how models fail in their reasoning. One significant obstacle to understanding the failure of large reasoning models in mathematics is the limited scale of benchmark datasets. 
Popular competition-level datasets, for example, AIME-2024~\cite{jia2025aime24} and AIME-2025~\cite{opencompass2025aime25}, consist of merely 30 problems each. 
Consequently, when a state-of-the-art model scores 90\% accuracy~\cite{openai2025o3}, it leaves only three incorrect instances for review, which is insufficient for a thorough failure analysis.
A second significant factor is the evaluation method that assess only the final numerical answer, a practice common in benchmarks such as GSM8K, MATH, and AIME~\cite{GSM8K, MathQA-dataset, MATH-dataset, jia2025aime24, opencompass2025aime25}. 
While efficient, such answer-based metrics cannot verify the logical soundness of the model's reasoning process~\cite{llm-math-failure,zheng2024processbench}. 

Addressing the issues above, mathematical proofs provide a more effective way to uncover flaws in a model’s reasoning. 
Unlike numerical answers that can obscure intermediate errors, proofs require explicit logical steps, where a single flawed or ambiguous inference can invalidate the entire argument. 
Moreover, proof-specific techniques such as constructive reasoning or proof by contradiction are often more complex than direct derivation and introduce more diverse reasoning errors. 
Existing benchmarks~\cite{miniF2F, ProofNet, FIMO-dataset, MATP-Bench} primarily evaluate models’ ability to generate proofs in formal languages (\textit{e.g.}, Lean). 
However, unlike specialized models designed for formal proof generation(\textit{e.g.}, AlphaGeometry~\cite{chervonyi2025gold}), current reasoning models are predominantly trained and evaluated on natural language reasoning tasks~\cite{wang2025let}. A huge performance gap emerges when they are required to solve the same problems using formal language~\cite{Dekoninck2025TheOP}, suggesting that the ability to generate formal proofs should not be equated with genuine reasoning ability. 

Considering these limitations, we propose to utilize mathematical proofs in natural language as an ideal diagnostic tool. 
To implement this idea, we manually construct the \textbf{\textit{RFMDataset}} to \textbf{\textit{R}}eveal the \textbf{\textit{F}}ailure \textbf{\textit{M}}ode of state-of-the-art reasoning models, which is composed of 200 mathematical proof problems manually selected from multiple sources, covering a diverse range of proof problems, as shown in Figure \ref{fig:knowledge-distribution}. 
We utilize an LLM-as-a-judge~\cite{llm-as-a-judge} framework to evaluate the overall correctness of the proofs, with human annotations provided to validate the reliability of the judge-LLM. 
Our evaluation framework also incorporates a fine-grained error classification system with over 10 distinct failure modes, enabling the judge-LLM to precisely categorize errors within proofs into one or multiple failure types.
A failure case with its failure mode analysis is shown in Figure \ref{fig:example}.


We evaluate 10 widely used, state-of-the-art reasoning models on our dataset. We find that even with impressive metrics reported on current competition-level benchmarks(\textit{e.g.}, AIME-2025), these models continue to struggle with proof tasks. Suprisingly, the widely used models like \textit{DeepSeek-R1-0120}~\cite{deepseek-r1} and \textit{Qwen3-235B}~\cite{Yang2025Qwen3TR} exhibit an overall accuracy below 20\%, even the top-performing \textit{Gemini-2.5-pro-preview-0605} model~\cite{deepmind2025gemini} achieves no more than 60\%. In addition, models exhibit unsatisfactory performance even on relatively basic proof problems, with eight models not achieving accuracy 60\% on the lowest difficulty level. Furthermore, our fine-grained failure mode analysis reveals that models always fail to ensure the correctness and rigor of single-step reasoning, while frequently rely on non-existent premises and provide incomplete proofs. We also conduct experiments that demonstrate prompting models to self-reflect on specific failure modes does not effectively resolve the issues. Based on these findings, we highlight the fundamental limitations and propose potential solutions, hoping to contribute to the future improvement of reasoning models.

%% file: latex/2_related_work.tex
\section{Related Work}
\paragraph{Natural Language Proof Benchmarks} 
A growing body of recent research has focused on the generation of natural language proofs. NaturalProofs~\cite{NaturalProofs-dataset} leverages natural language to evaluate the correctness of LLM generated proofs, but the scope is limited to over undergraduate levels.
Concurrent work OPC Dataset~\cite{Dekoninck2025TheOP} collects competition-level questions and answers from different LLMs, providing rich research materials. 
DeepTheorem~\cite{Zhang2025DeepTheoremAL} aims to improve models' ability to generate natural language proofs through the use of reinforcement learning.
IneqMath~\cite{Sheng2025SolvingIP} evaluates the ability of models to prove inequalities.
MathConstruct~\cite{Balunovic2025MathConstructCL} evaluates models' ability to generate constructive proofs.
While these valuable studies advance the field by improving and evaluating proof-generation performance, our work addresses a distinct and complementary goal, which is to systematically diagnose how reasoning fails with fine-grained taxonomy of failure modes. Unlike performance-oriented benchmarks such as the competition-focused OPC Dataset or the domain-specific IneqMath, our data collection methodology prioritizes novelty and diversity from a wide range of sources.

\paragraph{Analysis of Failure Modes}
While LLMs have achieved remarkable performance, the analysis of their reasoning failures is a growing field. In the mathematical domain, initial studies have explored error patterns~\cite{LLM-and-math-failures, USAMO-Proof-Bench}. However, these efforts are often limited by coarse-grained taxonomies that lack detail, or are constrained by a narrow scope, such as high-school level problems or very small sample sizes. 
Also distinct from our work are recent valuable studies that diagnose model weaknesses by applying external modifications to existing benchmarks. For instance, CR-MATH~\cite{Li2025OneES} introduces counterexamples to challenge the conceptual limits of models' reasoning, and MATH-Perturb~\cite{Huang2025MATHPerturbBL} applies systematic perturbations to evaluate the robustness of their solutions against superficial changes. Our approach differs in three fundamental ways. First, our \textit{\textbf{RFMDataset}} assesses reasoning on problems where the difficulty is intrinsic to the multi-step proof, rather than engineered through external tweaks. Second, our proof-centric paradigm enables a process-oriented diagnosis with a fine-grained taxonomy of logical fallacies, moving beyond the final-answer accuracy focus of robustness tests. Finally, our study is specifically designed to stress-test the latest generation of advanced models (\textit{e.g., Gemini-2.5-Pro, GPT-o3}), whose failure modes have evolved into the subtle logical inconsistencies that our new evaluation methodology is built to detect.

%% file: latex/3_method.tex
\section{Dataset Construction and Evaluation}
\label{sec:method}

\begin{figure}
    \centering
    \includegraphics[width=\linewidth]{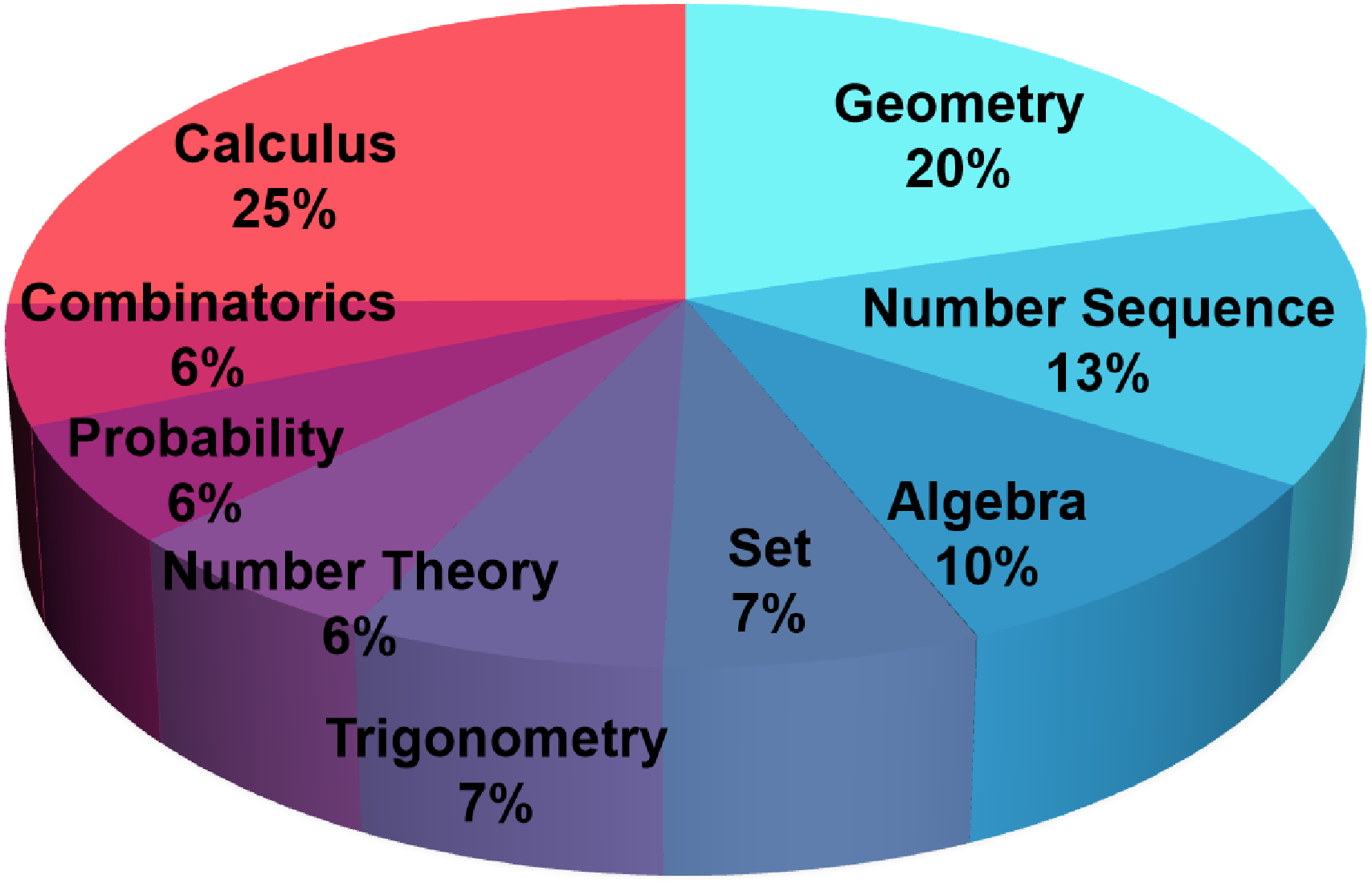}
    \caption{The knowledge distribution of \textbf{\textit{RFMDataset}}. We collect diverse questions from various domains to ensure the comprehensiveness of our evaluation.}
    \label{fig:knowledge-distribution}
    \vspace{-0.2in}
\end{figure}

In this section, we introduce the construction and key characteristics of our \textbf{\textit{RFMDataset}} and describe our evaluation methodology. Figure~\ref{fig:main} provides an overview. We begin with Section~\ref{Sec:RFMDataset}, which summarizes the dataset’s scale, coverage, difficulty design, and selection principles. We then present Section~\ref{sec:evaluation}, which outlines our LLM-as-a-judge evaluation framework and the failure mode taxonomy.

\subsection{RFMDataset}
\label{Sec:RFMDataset}
This subsection details the characteristic and construction of our dataset.

\subsubsection{Characteristics} 
Our dataset contains 200 selected mathematical proof problems, a scale comparable to established benchmarks such as miniF2F-test~\cite{miniF2F} and MathConstruct~\cite{Balunovic2025MathConstructCL}.
The problems are stratified by knowledge level, encompassing middle school (52 problems), high school (88 problems), and undergraduate curricula (60 problems).
Furthermore, the dataset covers nine distinct mathematical subjects, including but not limited to geometry, calculus, and probability, as outlined in Figure \ref{fig:knowledge-distribution}. 
Problems within each knowledge level are assigned one of four ascending difficulty levels (1 to 4) manually, with the most challenging problems approaching Olympiad difficulty.

\begin{figure*}[!h]
    \centering
    \includegraphics[width=\linewidth]{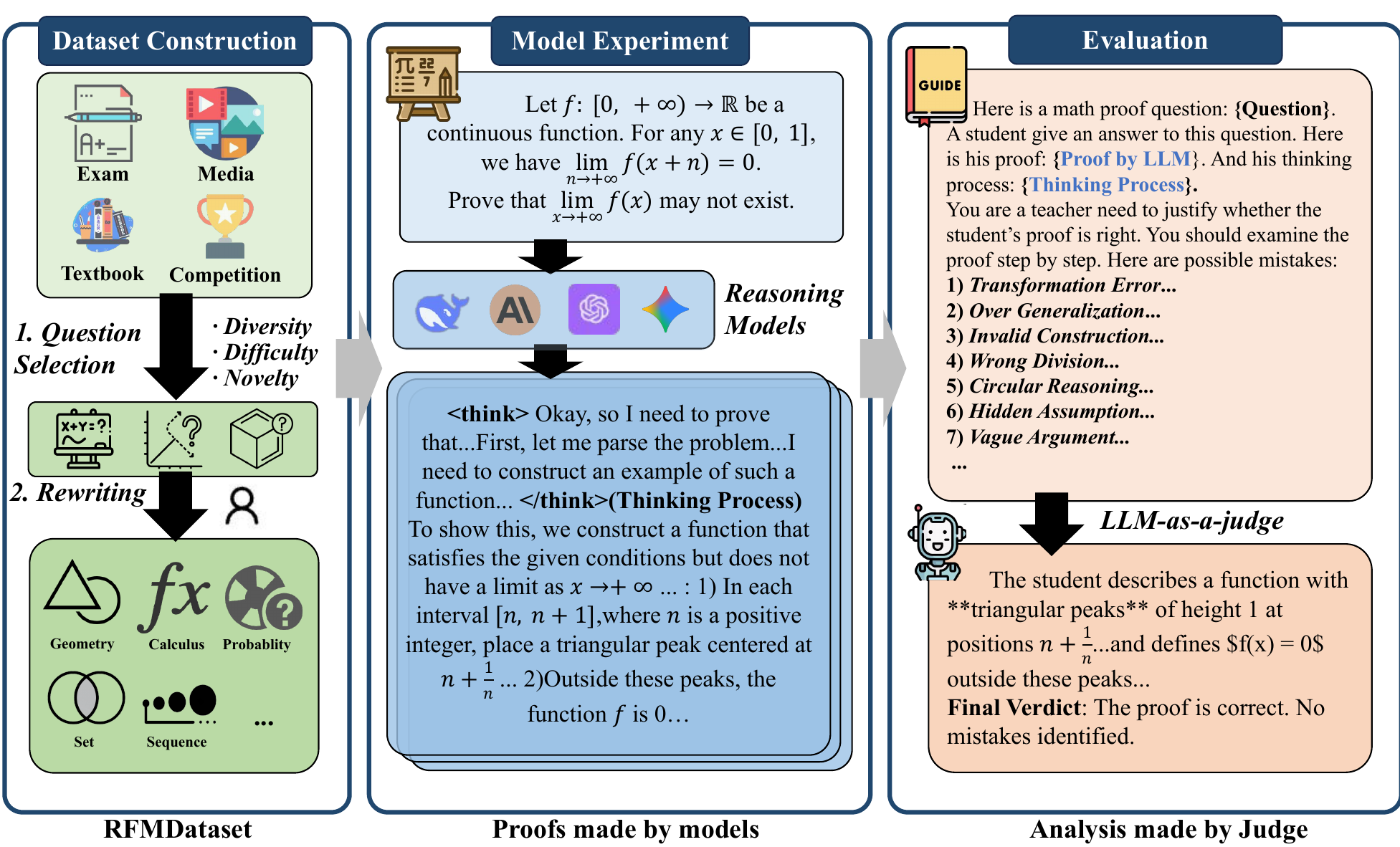}
    \caption{The construction of \textbf{\textit{RFMDataset}} and evaluation pipeline. We first manually collect challenging and diverse mathematical proof problems. Subsequently, we use an LLM-as-a-judge approach to evaluate the performance of inference models on this data, categorizing errors into various failure modes.}
    \label{fig:main}
\end{figure*}

\subsubsection{Selection Criteria} 
As depicted in Figure \ref{fig:main}, our dataset draws predominantly from four distinct sources: examinations, internet media, textbooks, and competitive contests. The data source analysis and comparison with other datasets are provided in the Appendix \ref{data_source}. To ensure the quality of our data, we manually select the data following three primary criteria:

\paragraph{Diversity.} This emphasis on diversity is crucial to prevent the benchmark from exhibiting an categorical bias towards specific problem types, thereby upholding its fairness and representativeness. 
We consider the diversity in knowledge and methodology when selecting our data. 
To ensure knowledge diversity, besides collecting questions from different subjects, we also make sure the diversity of problems by covering various fine-grained knowledge points, such as different related theorems. 
As for the methodology diversity which refers to the variety in problem-solving processes, we ensure it by reviewing each problem's solution process during problem collection. If we find a problem's solution too similar to one already collected, we retain only one of them.



\paragraph{Difficulty.} Since our objective is to identify patterns in model reasoning failures, we should make sure even the simplest problems should be solved with careful observation and critical analysis. So we exclude problems whose solution plans could be derived intuitively and deterministically. This rigorous selection process compels models to demonstrate truly in-depth problem-solving abilities. The difficulty level of each problem is determined by human annotators through studying reference solutions to understand the reasoning complexity involved, then comparing  the reasoning complexity of different problems.
Additionally, we apply special handling to geometry problems. First, we reformulated the problem descriptions to allow the geometric scene to be reconstructed purely from natural language. This adjustment accommodates models that lack vision capabilities and reduces extraneous difficulty. Second, we restrict models from using certain methods, such as parametric equations, to prevent them from reducing the geometric challenge to a more straightforward algebraic calculation.

\paragraph{Novelty.} We emphasize the novelty of the selected data, aiming for it to be distinct from commonly used mathematical benchmarks or data sources.
For questions sourced from examinations (e.g., Chinese Gaokao or Kaoyan), we prioritize modified versions crafted by mathematics enthusiasts, alongside recent regional exam problems. 
We specifically select problems exhibiting significant divergence from typical textual materials on online media as well as some published original and innovative problems. 
Textbook-derived problems are strictly limited to fundamental theorems that necessitate non-trivial proof techniques. 
Furthermore, we avoid overused sources directly, such as the International Mathematical Olympiad (IMO), which are already prevalent in existing research benchmarks. 

\begin{table*}[!ht]
\small
  \centering
  \begin{tabular}{@{}p{3.2cm} p{6.0cm} p{6.5cm}@{}}
    \toprule
    \textbf{Category} & \textbf{Definition} & \textbf{Illustrative Example} \\
    \midrule
    \textit{Transformation Error} &
      Recasting the target statement into a \emph{non-equivalent} or strictly weaker one, so a “proof” no longer addresses the real goal. &
      To show a series \(\sum a_n\) converges, one proves only \(\lim_{n\to\infty} a_n = 0\); or replaces “\(A\iff B\)” with the easier “\(A\Rightarrow B\)”.\\
      \midrule
    \textit{Over Generalization} &
      Drawing a universal conclusion from only a few special cases or situations. &
      Verifying the claim for \(n = 1, 3, 5\) and then declaring it true for \emph{all} positive integers.\\
      \midrule
    \textit{Invalid Construction} &
      Presenting an object that either cannot exist under the stated conditions or fails the required properties. &
      Claiming the existence of a function that is “everywhere linear yet nowhere differentiable”; or defining \(f(x)=1/x\) on all \(\mathbb{R}\) without addressing \(x=0\).\\
      \midrule
    \textit{Wrong Division} &
      Partitioning into cases that miss at least one legitimate possibility or overlap each other. &
      When analyzing the behavior of a function, dividing cases as “always positive,” “always zero,” and “always negative.”\\
    \midrule
    \textit{Circular Reasoning} &
      Using the desired conclusion (or an equivalent reformulation) as an explicit or hidden premise. &
      To prove \(A\Rightarrow B\), but use premises that implicitly assumes \(B\).\\
      \midrule
    \textit{Logic Violation} &
      A single deduction step contradicts logical or algebraic rules. &
      From \(a<b\) and \(c<d\) concluding \(a-c<b-d\) (false when \(c,d\) are negative).\\
      \midrule
    \textit{Hidden Assumption} &
      Applying theorems or operations whose hypotheses have not been established or stated. &
      Differentiating a function known only to be continuous; interchanging limit and integral without proving uniform convergence.\\
      \midrule
    \textit{Boundary Neglect} &
      Ignoring edge cases, endpoints, or limiting situations so the proof works only “in the middle.” &
      Declaring \(f(x)=\sqrt{x}\) differentiable on \([0,1]\) without checking \(x=0\).\\
      \midrule
    \textit{Vague Argument} &
      Relying on intuition, diagrams, or the word “obvious” rather than rigorous justification. &
      “The series obviously converges because the terms get smaller”; “One sees from the picture that the two segments are equal.”\\
      \midrule
    \textit{Incomplete Proof} &
      Omitting an essential component in proof, or even providing an unfinished proof. &
      Proving sufficiency but not necessity; in induction, writing the inductive hypothesis but never showing how \(P(k)\) implies \(P(k+1)\).\\
      
    \midrule
    \textit{Others} &
      Any error type not covered by the categories above. &
      — \\
    \bottomrule

  \end{tabular}
  \caption{Classification of Typical Proof Errors. 
  We identify 10 specific failure modes and group the rest into an “Others” category. 
  Each failure mode is clearly defined and illustrated with an example.
}
  \label{tab:proof_error_en}
\end{table*}

\subsection{Evaluation}
\label{sec:evaluation}
We employ two methods to evaluate the models' proofs: LLM-as-a-judge~\cite{llm-as-a-judge-survey, fan2025sedareval} and human evaluation.

Our LLM-as-a-judge method extends beyond holistic proof verification.
We develop a fine-grained error taxonomy comprising over 10 reasoning failure modes, including \textit{Logical Violation, Over Generalization}, and \textit{Circular Reasoning}.
This enables the precise classification of model-generated proof failures, offering a deeper understanding of their shortcomings. 
Table \ref{tab:proof_error_en} presents our complete taxonomy of failure modes, providing a detailed definition and concise, illustrative examples for each category.

Our taxonomy of failure modes is not homogeneous. It contains both classical, well-established logical fallacies (\textit{e.g., Circular Reasoning}) and patterns empirically identified from the behavior of current LLMs (\textit{e.g., Over Generalization}). Beyond this direct diagnosis, our taxonomy serves as a practical surrogate for a proof's intrinsic value. A proof's true worth often lies in a key insight (\textit{e.g.}, a clever construction), which should be highly credited even with minor flaws. However, this is a major challenge for an automated judge, because the judge often cannot solve the problem itself and therefore lacks the necessary insight to reliably identify these key ideas(providing a single gold proof as a reference is also not helpful, as multiple valid proof paths often exist). We therefore adopt a more robust and tractable approach, which is to verify the step-by-step logical integrity, making the absence of identifiable failures a reliable proxy for a proof's soundness.

The purpose of the human evaluation is to measure the reliability of the LLM-as-a-judge method. In the human evaluation process, we thoroughly assess the correctness of proofs by giving a binary pass or fail on the entire proof chain, then use correlation coefficients to show the consistency between human and LLM-generated labels.


%% file: latex/4_experiment.tex
\section{Experiments}

In this section, we detail the settings in our experiments, including the experiment setup and the dual evaluation pipeline containing LLM-as-a-judge as well as human evaluation.

\subsection{Experiment Setup}
\paragraph{Models.} We experiment with three state-of-the-art open-source and seven proprietary LLMs to test the greatest models' proving capabilities.
In the LLM-as-a-judge evaluation, we unifiedly use \textit{Gemini-2.5-pro-preview-0506}~\cite{Huang2025Gemini2P} to evaluate all the models.
For all models, the generation parameters were set to default values, with a \textit{top\_p} of 1.0 and a \textit{temperature} of 1.0. 

\begin{figure*}[ht]
\centering
\vspace{-0.0em}
\subfloat{\includegraphics[trim={0cm, 0cm 0cm 0cm},height=2in]{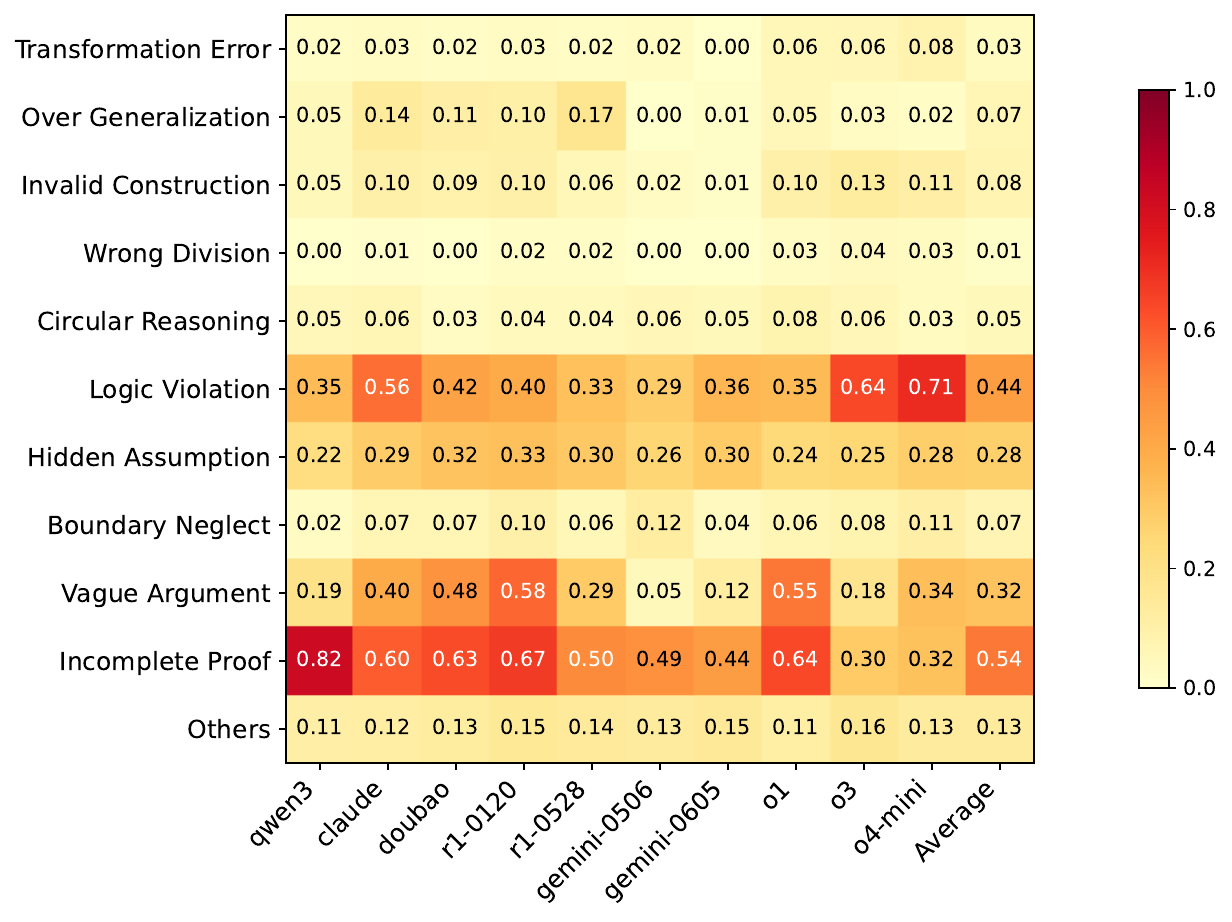}%
\label{fig:error_pattern_model}}
\hfil
\subfloat{\includegraphics[trim={0cm, 0cm 0cm 0cm},height=2.04in]{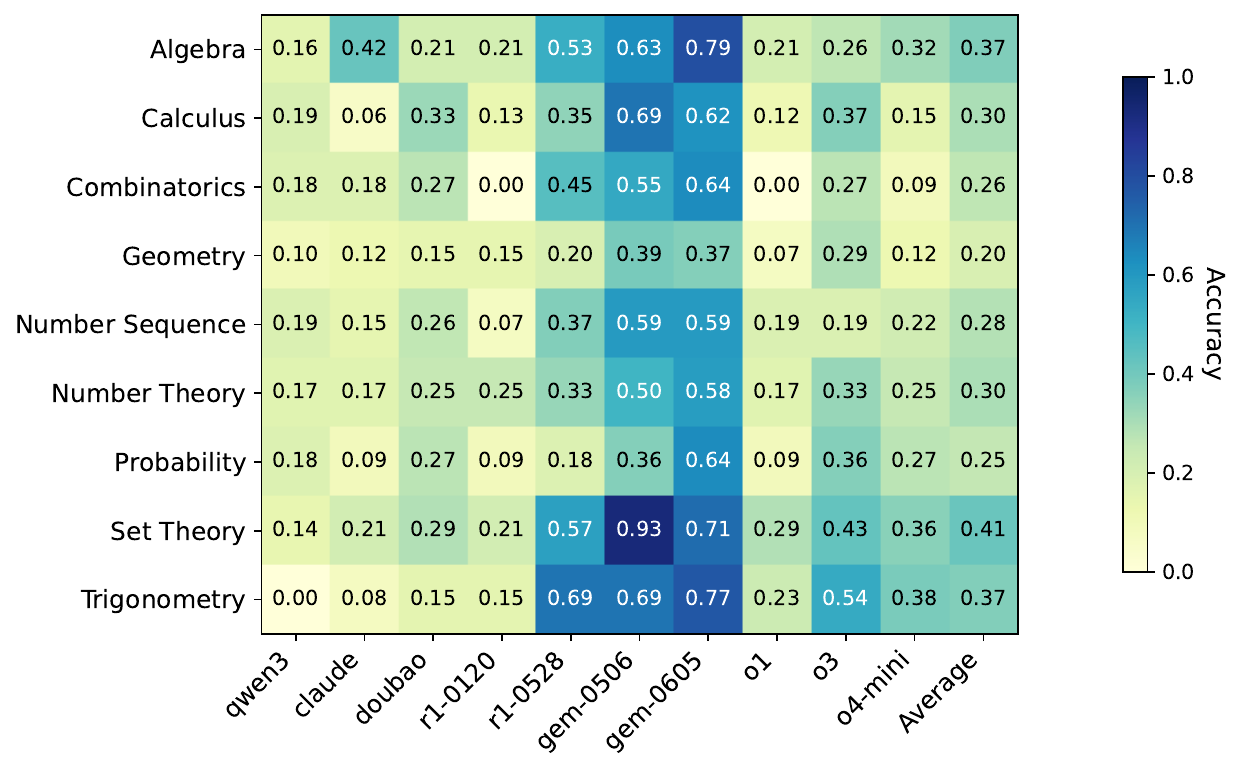}%
\label{fig:heatmap_tag_model}}
\caption{\textit{(a)} The proportion of each error pattern of each model. The \textit{Average} column summarizes the average proportion across all models for each pattern. \textit{~(b)} Accuracy of each model on every knowledge domain. The \textit{Average} column summarizes the average accuracy across all models for each domain.}
\label{fig:heapmap_all}
\end{figure*}

\paragraph{Prompts.} In the main experiments, we directly provide the question to models with no extra information. The evaluation prompt is shown in Figure \ref{fig:eval-prompt}, where we demand the judge model to analyze failure modes and provide overall correctness.

\paragraph{Metrics.} 
Contrary to prior work using the pass@k metric~\cite{ds-prover, Wang2025KiminaProverPT}, we employ pass@1 accuracy. This choice is driven by our primary objective of analyzing failure modes, in which a single sampling provides an adequate number of failure cases, as well as the prohibitive cost of generating multiple responses.


\subsection{Human Evaluation}
To validate the reliability of our LLM-as-a-judge, we randomly sampled 240 model-generated answers (60 each from \textit{Deepseek-R1-0120, Deepseek-R1-0528, Gemini-2.5-Pro-Preview-0506, and GPT-4o-mini-0416}) for human evaluation. The agreement between the judge's assignments and human labels was then measured using the Matthews Correlation Coefficient (MCC)~\cite{xu2025mpbenchcomprehensivemultimodalreasoning, 1975Matthews}.
As shown in Table~\ref{tab:mcc_comparison}, the results demonstrate high consistency between human and judge model's labels. 
\begin{table}[h]
    \small
    \centering
    \begin{tabular}{p{0.70\columnwidth} p{0.15\columnwidth}}
        \toprule
        \textbf{Models} & \textbf{MCC\%}\\
        \hline
        \textit{Deepseek-R1-0120} & 71.97 \\
        \textit{Deepseek-R1-0528} & 86.60 \\
        \textit{Gemini-2.5-pro-preview-0506} & 86.60 \\
        \textit{GPT-o4-mini-0416} & 91.09 \\
        \hline
        \textit{Overall} & 87.61 \\
        \bottomrule
    \end{tabular}
    \caption{The MCC between the manual labels of an-
swers from different models and the LLM labels.}
    \label{tab:mcc_comparison}
\end{table}

The data annotation process is reported in Appendix~\ref{app:data-annotation}. The discussion regarding the use of other LLMs as judges(\textit{e.g.}, \textit{GPT-5}~\cite{openai2025gpt5}) is included in the Appendix \ref{judge_llms}. The overall evaluation results from \textit{GPT-5} and \textit{Gemini} are highly similar, but both differ significantly from those of other models. Considering the advanced capabilities of these two models, and with \textit{Gemini}'s assessments having been validated by human evaluation, we are confident that our reported results are reliable.


\section{Experimental Findings}
For the experiment, we propose the following three research questions (RQs), and analyze and discuss them based on our experimental results.
\begin{itemize}
    \item[\(\bullet\)] \textbf{RQ1}: How reliably can large reasoning models generate rigorous proof? If not reliable, which domain do they perform poorly on?
    \item[\(\bullet\)] \textbf{RQ2}: Do different models exhibit comparable failure mode distributions, and what insights can be gleaned from analyzing them?
    \item[\(\bullet\)] \textbf{RQ3}: Can models show improvement when we explicitly prompt models to self-reflect on potential failure modes?
\end{itemize}

\begin{table*}[h!]
    \small
    \centering
    \begin{tabular}{p{5.5cm} >{\centering\arraybackslash}p{1.75cm} >{\centering\arraybackslash}p{1.75cm} >{\centering\arraybackslash}p{1.75cm} >{\centering\arraybackslash}p{1.75cm}}
        \toprule
        Model & MS & HS & UG & Overall \\
        \midrule
        \textit{Qwen3-235B-A22B} & 15.38 & \phantom{0}9.09 & 23.33 & 15.00 \\
        \textit{Doubao-1.5-thinking-pro} & 19.23 & 21.59 & 33.33 & 24.50 \\
        \textit{Deepseek-R1-0120} &  17.31&  10.23&  16.67&  14.00 \\
        \textit{Deepseek-R1-0528} &  26.92 &  42.05&  38.33 &  37.00 \\
        \textit{DeepSeek-V3.2-Speciale} &  {\colorbox{cyan!80}{63.46}} &  {\colorbox{cyan!80}{68.18}} &  {\colorbox{cyan!50}{66.67}} &  {\colorbox{cyan!80}{66.50}} \\
        \textit{Gemini-2.5-Pro-Preview-0506} &{\colorbox{cyan!50}{51.92}}&  {\colorbox{cyan!50}{53.41}}&  {\colorbox{cyan!80}{73.33}}&  {\colorbox{cyan!50}{59.00}} \\
        \textit{Gemini-2.5-Pro-Preview-0605} & 50.00&  {\colorbox{cyan!80}{60.23}}&  {\colorbox{cyan!50}{66.67}}&  {\colorbox{cyan!80}{59.50}} \\
        \textit{Gemini-3-Pro-Preview} &{\colorbox{cyan!50}{57.69}}&  48.86&  {\colorbox{cyan!50}{66.67}} &  56.50 \\
        \textit{Claude-3.7-Sonnet-Thinking} &  19.23&  13.64&  11.67&  14.50 \\
        \textit{GPT-o1} &  \phantom{0}7.69&  14.77&  18.33&  14.00 \\
        \textit{GPT-o3-0416} &  30.77&  29.55&  38.33 &  32.50 \\
        \textit{GPT-o4-mini-0416} &  15.38&  22.73&  23.33&  21.00\\
        \bottomrule
        
    \end{tabular}
    \caption{The accuracy (\%) of different state-of-the-art LLMs in \textbf{\textit{RFMDataset}}. MS, HS and UG means middle-school, high-school and undergraduate level problems correspondingly. Across all models, the scores highlighted in dark, medium, and shallow cyan indicate the highest, second-highest, and third-highest performances, respectively. 
    }
    \label{tab:main-results}
    \vspace{-0.2in}
\end{table*}

\subsection{Overall Evaluation (RQ1)}
To answer RQ1, we conduct an overall evaluation of the models' ability to generate completely correct proofs. 
The evaluation results are shown in Table~\ref{tab:main-results}.
Additionally, we also evaluate the accuracy of different models across various knowledge domains and difficulty levels, and the results are presented in Figure~\ref{fig:heatmap_tag_model} and Figure~\ref{fig:accuracy_per_level}. Our key findings are presented as follows:

\paragraph{Models exhibit limited overall effectiveness.}
Table~\ref{tab:main-results} presents the overall accuracy of LLM-as-a-judge evaluations, which shows that models generally face significant challenges. 
Surprisingly, the accuracy of models such as \textit{Qwen3-235B-A22B}, \textit{Deepseek-R1-0120}, \textit{Claude-3.7-Sonnet-Thinking}, and \textit{GPT-o1} does not exceed 20\%. Two versions of \textit{Gemini2.5-Pro-Preview} achieve a significantly higher accuracy of nearly 60\%, substantially outperforming other models. 

\paragraph{Models' performance vary significantly across different domains.}
Figure~\ref{fig:heatmap_tag_model} presents our analysis, revealing that models consistently achieve lower accuracy rates in geometry, number sequences, combinatorics, and probability. 
In stark contrast, these same models show relatively higher accuracy in algebra and number theory tasks.

\paragraph{Models may fail on easy problems.}
Our analysis revealed a surprising finding: state-of-the-art models fail even on easy problems. As depicted in Figure \ref{fig:accuracy_per_level}, model accuracy generally declines as problem difficulty increases, which aligns with our difficulty annotations. However, it's particularly noteworthy that the majority of models fall short even at the lowest difficulty level, with all models except two \textit{Gemini} variants, achieving less than 60\% accuracy.


\subsection{Failure Mode Analysis (RQ2)}

\begin{table*}[h!]
    \small
    \centering
    
    \begin{tabular}{p{5cm} >{\centering\arraybackslash}p{1.5cm} >{\centering\arraybackslash}p{1.5cm} >{\centering\arraybackslash}p{1.5cm} >{\centering\arraybackslash}p{1.5cm} >{\centering\arraybackslash}p{1.5cm}}
        \toprule
        Model & MS & HS & UG & Overall & Best \\
        
        \midrule
        \textit{Doubao-1.5-thinking-pro} & 19.23 & 21.59 & 33.33 & 24.50 &  \multirow{4}{*}{\textbf{31.50}} \\

        \textit{Doubao-1.5-thinking-pro} \scriptsize{\textit{+reflection\_l}} & 28.85 & 19.32 & 26.67 & 24.00 & \\
        
        \textit{Doubao-1.5-thinking-pro} \scriptsize{\textit{+reflection\_vi}} & 28.85 & 21.59 & \colorbox{cyan!50}{36.67} & 28.00 & \\
        
        \textit{Doubao-1.5-thinking-pro} \scriptsize{\textit{+reflection\_lvi}} & \colorbox{cyan!80}{36.54} & \colorbox{cyan!20}{25.00} & 28.33 & \colorbox{cyan!50}{29.00} & \\
        \midrule
        \textit{GPT-o4-mini} & 15.38 & 22.73 & 23.33 & 21.00 & \multirow{4}{*}{\textbf{27.00}} \\

        \textit{GPT-o4-mini} \scriptsize{\textit{+reflection\_l}} & 19.23 & 22.73 & \colorbox{cyan!20}{31.67} & \colorbox{cyan!20}{24.50} & \\
        
        \textit{GPT-o4-mini} \scriptsize{\textit{+reflection\_vi}} & 17.31 & 22.73 & 20.00 & 20.50 & \\
        
        \textit{GPT-o4-mini} \scriptsize{\textit{+reflection\_lvi}} & \colorbox{cyan!20}{21.15} & \colorbox{cyan!50}{27.27} & 20.00 & 23.50 & \\

        \midrule
        \textit{Deepseek-R1-0528} & 26.92 & \colorbox{cyan!80}{42.05} & 38.33 & 37.00 & \multirow{4}{*}{\textbf{42.00}} \\

        \textit{Deepseek-R1-0528} \scriptsize{\textit{+reflection\_l}} & 32.69 & 32.95 & 43.33 & 36.00 & \\
        
        \textit{Deepseek-R1-0528} \scriptsize{\textit{+reflection\_vi}} & 19.23 & 35.23 & 40.00 & 32.50 & \\
        
        \textit{Deepseek-R1-0528} \scriptsize{\textit{+reflection\_lvi}} & \colorbox{cyan!50}{34.62} & 31.82 & \colorbox{cyan!80}{48.33} & \colorbox{cyan!80}{37.50} & \\
        \bottomrule
    \end{tabular}
    \caption{The accuracy (\%) of the selected large reasoning models on the \textbf{\textit{RFMDataset}}. For each subset of the dataset, the highest accuracy achieved across different baselines for the same model is highlighted in blue, and the color is darker for better performance across different models. The \textit{Best} column reports the overall of these highlighted scores, providing an overall measure of the model's best performance across all subsets. 
    }
    \label{tab:analysis-self-refine}
    \vspace{-0.1in}
\end{table*}

As evaluating accuracy is insufficient to uncover the deeper root causes of errors in the model's reasoning process, we further provide a more fine-grained diagnostic of the underlying error causes by analyzing the failure modes.
Figure~\ref{fig:error_pattern_model} illustrates the distribution of failure modes across all incorrect proofs. Our main findings are as follows:

\paragraph{The failure mode distributions of different models are similar and not much relevant to difficulty.}
Different models exhibit strikingly similar failure mode distributions, predominantly concentrated in four categories: \textit{logical violation}, \textit{hidden assumption}, \textit{vague argument}, and \textit{incomplete proof}. 
While other failure modes occur, their proportions are relatively low. 
Interestingly, the dominance of these four modes remain largely unchanged when analyzing failure modes across various problem difficulty levels(shown in Appendix~\ref{app:different-difficulty-same-failure-mode}). 
This suggests that despite significant advancements in reasoning capabilities, models still possess fundamental flaws in their underlying reasoning processes. 


\paragraph{Models have significant shortcomings in the correctness and rigor of single reasoning steps.} 
\textit{Logical violation} and \textit{vague argument} are commonly observed as reasons for model failures. 
These categories directly indicate models' widespread shortcomings in the correctness and rigor of single-step reasoning. 
We can observe that the \textit{Gemini} models exhibits a significantly lower proportion of these two failure modes compared to other models, which may contribute to their superior overall performance.

\paragraph{Models struggle with severe hallucination and incompleteness issues in reasoning process.} 
According to Figure \ref{fig:error_pattern_model}, models also demonstrate prominent issues in \textit{hidden assumption} and \textit{incomplete proof}. 
\textit{Hidden assumption} can be treated as a form of hallucination, since it refers to the use of unprovided conditions during problem-solving. 
An \textit{Incomplete Proof} signifies that the model provides an insufficient demonstration, failing to independently determine when a proof is complete. 
These two aspects together reveal a fundamental lack of understanding on both the initial proposition and what should be done to prove the proposition.



\subsection{Mode-specific self-relection (RQ3)}
To improve performance, we prompt models to self-reflect on their initial proofs. Our rationale is that many errors stem not from fundamental reasoning incapacity, but from superficial flaws like oversights or presentational gaps that are amenable to self-correction. For instance, a proof flagged as a \textit{Vague Argument} might reflect a momentary lapse in rigor during a specific generation, rather than a fundamental inability to reason precisely. Similarly, a proof marked as \textit{Incomplete Proof} could be logically sound but rely on an obscure or non-trivial theorem, which the judge flags as a gap because it was not explicitly stated or proven.

In response, we design three additional baseline prompts: 
1) Target solving \textit{logical violation}: the prompt asks the model to perform step-by-step self reflection after generating answers (\textit{reflection\_l}).
2) Target solving \textit{vague argument} and \textit{incomplete proof}: based on observed phenomena, the prompt instructs the model to reason rigorously and prove any non-trivial theorems it employs (\textit{reflection\_vi}).  
3) Combination of the two kinds of reflection above   (\textit{reflection\_lvi}). Detailed prompts are reported in Appendix~\ref{app:experiment-details}.

We conduct our experiments using \textit{Doubao-1.5-thinking-pro}, \textit{o4-mini}, and \textit{Deepseek-R1-0528}. The results are shown in Table~\ref{tab:analysis-self-refine}.
The \textit{Best} column reports a synthetic score, calculated as the weighted average of the highest score achieved within each of the MS, HS, and UG categories respectively, across all tested reflection strategies for a given model. This aggregated score synthesizes the model's upper-bound potential, showing its performance if one could perfectly select the optimal reflection strategy for each category.

Our experimental results demonstrate that prompts designed to address specific failure modes can enhance the model's performance in at least one of the MS, HS, or UG problem categories, and even lead to an improvement in overall performance. 
Furthermore, the \textit{Best} results suggest that these models possess untapped potential for improvement when guided by failure mode prompting. 
However, these gains are modest when compared to the substantial number of problems the model still fails to solve. We discuss possible solutions for improving the future models in Appendix \ref{app:possible_solutions}.

%% file: latex/5_conclusion.tex
\section{Conclusion}

In this paper, we introduce \textbf{\textit{RFMDataset}}, a novel and high-quality dataset comprised of 200 carefully selected mathematical proof problems designed to revealing the reasoning failure modes of large reasoning models. We utilize LLM-as-a-judge to evaluate the efficiency of advanced models on our dataset and discover the failure mode distributions. Our findings demonstrate that models struggle on generating entirely correct proofs. Specifically, they frequently fall into single-step reasoning errors, exhibit vague argumentation, utilize non-existent premises, and produce incomplete proofs. Furthermore, our experiments reveal that explicitly prompting models to self-reflect on these specific failure modes is insufficient to resolve these issues, indicating fundamental drawbacks in current large reasoning models. For future work, we propose to take advantage of formally-verifiable environment and high-quality domain-specific data to endow models the mathematician's way of thinking.

%% file: custom.bib
@article{CostBench,
  author       = {Jiayu Liu and
                  Cheng Qian and
                  Zhaochen Su and
                  Qing Zong and
                  Shijue Huang and
                  Bingxiang He and
                  Yi R. Fung},
  title        = {CostBench: Evaluating Multi-Turn Cost-Optimal Planning and Adaptation
                  in Dynamic Environments for {LLM} Tool-Use Agents},
  journal      = {CoRR},
  volume       = {abs/2511.02734},
  year         = {2025},
  url          = {https://doi.org/10.48550/arXiv.2511.02734},
  doi          = {10.48550/ARXIV.2511.02734},
  eprinttype    = {arXiv},
  eprint       = {2511.02734},
  bibsource    = {dblp computer science bibliography, https://dblp.org}
}

@misc{deepseek-r1,
      title={DeepSeek-R1: Incentivizing Reasoning Capability in LLMs via Reinforcement Learning}, 
      author={DeepSeek-AI and Daya Guo and Dejian Yang and Haowei Zhang and Junxiao Song and Ruoyu Zhang and Runxin Xu and Qihao Zhu and Shirong Ma and Peiyi Wang and Xiao Bi and Xiaokang Zhang and Xingkai Yu and Yu Wu and Z. F. Wu and Zhibin Gou and Zhihong Shao and Zhuoshu Li and Ziyi Gao and Aixin Liu and Bing Xue and Bingxuan Wang and Bochao Wu and Bei Feng and Chengda Lu and Chenggang Zhao and Chengqi Deng and Chenyu Zhang and Chong Ruan and Damai Dai and Deli Chen and Dongjie Ji and Erhang Li and Fangyun Lin and Fucong Dai and Fuli Luo and Guangbo Hao and Guanting Chen and Guowei Li and H. Zhang and Han Bao and Hanwei Xu and Haocheng Wang and Honghui Ding and Huajian Xin and Huazuo Gao and Hui Qu and Hui Li and Jianzhong Guo and Jiashi Li and Jiawei Wang and Jingchang Chen and Jingyang Yuan and Junjie Qiu and Junlong Li and J. L. Cai and Jiaqi Ni and Jian Liang and Jin Chen and Kai Dong and Kai Hu and Kaige Gao and Kang Guan and Kexin Huang and Kuai Yu and Lean Wang and Lecong Zhang and Liang Zhao and Litong Wang and Liyue Zhang and Lei Xu and Leyi Xia and Mingchuan Zhang and Minghua Zhang and Minghui Tang and Meng Li and Miaojun Wang and Mingming Li and Ning Tian and Panpan Huang and Peng Zhang and Qiancheng Wang and Qinyu Chen and Qiushi Du and Ruiqi Ge and Ruisong Zhang and Ruizhe Pan and Runji Wang and R. J. Chen and R. L. Jin and Ruyi Chen and Shanghao Lu and Shangyan Zhou and Shanhuang Chen and Shengfeng Ye and Shiyu Wang and Shuiping Yu and Shunfeng Zhou and Shuting Pan and S. S. Li and Shuang Zhou and Shaoqing Wu and Shengfeng Ye and Tao Yun and Tian Pei and Tianyu Sun and T. Wang and Wangding Zeng and Wanjia Zhao and Wen Liu and Wenfeng Liang and Wenjun Gao and Wenqin Yu and Wentao Zhang and W. L. Xiao and Wei An and Xiaodong Liu and Xiaohan Wang and Xiaokang Chen and Xiaotao Nie and Xin Cheng and Xin Liu and Xin Xie and Xingchao Liu and Xinyu Yang and Xinyuan Li and Xuecheng Su and Xuheng Lin and X. Q. Li and Xiangyue Jin and Xiaojin Shen and Xiaosha Chen and Xiaowen Sun and Xiaoxiang Wang and Xinnan Song and Xinyi Zhou and Xianzu Wang and Xinxia Shan and Y. K. Li and Y. Q. Wang and Y. X. Wei and Yang Zhang and Yanhong Xu and Yao Li and Yao Zhao and Yaofeng Sun and Yaohui Wang and Yi Yu and Yichao Zhang and Yifan Shi and Yiliang Xiong and Ying He and Yishi Piao and Yisong Wang and Yixuan Tan and Yiyang Ma and Yiyuan Liu and Yongqiang Guo and Yuan Ou and Yuduan Wang and Yue Gong and Yuheng Zou and Yujia He and Yunfan Xiong and Yuxiang Luo and Yuxiang You and Yuxuan Liu and Yuyang Zhou and Y. X. Zhu and Yanhong Xu and Yanping Huang and Yaohui Li and Yi Zheng and Yuchen Zhu and Yunxian Ma and Ying Tang and Yukun Zha and Yuting Yan and Z. Z. Ren and Zehui Ren and Zhangli Sha and Zhe Fu and Zhean Xu and Zhenda Xie and Zhengyan Zhang and Zhewen Hao and Zhicheng Ma and Zhigang Yan and Zhiyu Wu and Zihui Gu and Zijia Zhu and Zijun Liu and Zilin Li and Ziwei Xie and Ziyang Song and Zizheng Pan and Zhen Huang and Zhipeng Xu and Zhongyu Zhang and Zhen Zhang},
      year={2025},
      eprint={2501.12948},
      archivePrefix={arXiv},
      primaryClass={cs.CL},
      url={https://arxiv.org/abs/2501.12948}, 
}

@inproceedings{MATH-dataset,
  author       = {Dan Hendrycks and
                  Collin Burns and
                  Saurav Kadavath and
                  Akul Arora and
                  Steven Basart and
                  Eric Tang and
                  Dawn Song and
                  Jacob Steinhardt},
  editor       = {Joaquin Vanschoren and
                  Sai{-}Kit Yeung},
  title        = {Measuring Mathematical Problem Solving With the {MATH} Dataset},
  booktitle    = {Proceedings of the Neural Information Processing Systems Track on
                  Datasets and Benchmarks 1, NeurIPS Datasets and Benchmarks 2021, December
                  2021, virtual},
  year         = {2021},
  url          = {https://datasets-benchmarks-proceedings.neurips.cc/paper/2021/hash/be83ab3ecd0db773eb2dc1b0a17836a1-Abstract-round2.html},
  bibsource    = {dblp computer science bibliography, https://dblp.org}
}

@article{GSM8K,
  author       = {Karl Cobbe and
                  Vineet Kosaraju and
                  Mohammad Bavarian and
                  Mark Chen and
                  Heewoo Jun and
                  Lukasz Kaiser and
                  Matthias Plappert and
                  Jerry Tworek and
                  Jacob Hilton and
                  Reiichiro Nakano and
                  Christopher Hesse and
                  John Schulman},
  title        = {Training Verifiers to Solve Math Word Problems},
  journal      = {CoRR},
  volume       = {abs/2110.14168},
  year         = {2021},
  url          = {https://arxiv.org/abs/2110.14168},
  eprinttype    = {arXiv},
  eprint       = {2110.14168},
  bibsource    = {dblp computer science bibliography, https://dblp.org}
}

@misc{llm-as-a-judge,
      title={Judging LLM-as-a-Judge with MT-Bench and Chatbot Arena}, 
      author={Lianmin Zheng and Wei-Lin Chiang and Ying Sheng and Siyuan Zhuang and Zhanghao Wu and Yonghao Zhuang and Zi Lin and Zhuohan Li and Dacheng Li and Eric P. Xing and Hao Zhang and Joseph E. Gonzalez and Ion Stoica},
      year={2023},
      eprint={2306.05685},
      archivePrefix={arXiv},
      primaryClass={cs.CL},
      url={https://arxiv.org/abs/2306.05685}, 
}

@misc{openai-o1,
      title={OpenAI o1 System Card}, 
      author={OpenAI and : and Aaron Jaech and Adam Kalai and Adam Lerer and Adam Richardson and Ahmed El-Kishky and Aiden Low and Alec Helyar and Aleksander Madry and Alex Beutel and Alex Carney and Alex Iftimie and Alex Karpenko and Alex Tachard Passos and Alexander Neitz and Alexander Prokofiev and Alexander Wei and Allison Tam and Ally Bennett and Ananya Kumar and Andre Saraiva and Andrea Vallone and Andrew Duberstein and Andrew Kondrich and Andrey Mishchenko and Andy Applebaum and Angela Jiang and Ashvin Nair and Barret Zoph and Behrooz Ghorbani and Ben Rossen and Benjamin Sokolowsky and Boaz Barak and Bob McGrew and Borys Minaiev and Botao Hao and Bowen Baker and Brandon Houghton and Brandon McKinzie and Brydon Eastman and Camillo Lugaresi and Cary Bassin and Cary Hudson and Chak Ming Li and Charles de Bourcy and Chelsea Voss and Chen Shen and Chong Zhang and Chris Koch and Chris Orsinger and Christopher Hesse and Claudia Fischer and Clive Chan and Dan Roberts and Daniel Kappler and Daniel Levy and Daniel Selsam and David Dohan and David Farhi and David Mely and David Robinson and Dimitris Tsipras and Doug Li and Dragos Oprica and Eben Freeman and Eddie Zhang and Edmund Wong and Elizabeth Proehl and Enoch Cheung and Eric Mitchell and Eric Wallace and Erik Ritter and Evan Mays and Fan Wang and Felipe Petroski Such and Filippo Raso and Florencia Leoni and Foivos Tsimpourlas and Francis Song and Fred von Lohmann and Freddie Sulit and Geoff Salmon and Giambattista Parascandolo and Gildas Chabot and Grace Zhao and Greg Brockman and Guillaume Leclerc and Hadi Salman and Haiming Bao and Hao Sheng and Hart Andrin and Hessam Bagherinezhad and Hongyu Ren and Hunter Lightman and Hyung Won Chung and Ian Kivlichan and Ian O'Connell and Ian Osband and Ignasi Clavera Gilaberte and Ilge Akkaya and Ilya Kostrikov and Ilya Sutskever and Irina Kofman and Jakub Pachocki and James Lennon and Jason Wei and Jean Harb and Jerry Twore and Jiacheng Feng and Jiahui Yu and Jiayi Weng and Jie Tang and Jieqi Yu and Joaquin Quiñonero Candela and Joe Palermo and Joel Parish and Johannes Heidecke and John Hallman and John Rizzo and Jonathan Gordon and Jonathan Uesato and Jonathan Ward and Joost Huizinga and Julie Wang and Kai Chen and Kai Xiao and Karan Singhal and Karina Nguyen and Karl Cobbe and Katy Shi and Kayla Wood and Kendra Rimbach and Keren Gu-Lemberg and Kevin Liu and Kevin Lu and Kevin Stone and Kevin Yu and Lama Ahmad and Lauren Yang and Leo Liu and Leon Maksin and Leyton Ho and Liam Fedus and Lilian Weng and Linden Li and Lindsay McCallum and Lindsey Held and Lorenz Kuhn and Lukas Kondraciuk and Lukasz Kaiser and Luke Metz and Madelaine Boyd and Maja Trebacz and Manas Joglekar and Mark Chen and Marko Tintor and Mason Meyer and Matt Jones and Matt Kaufer and Max Schwarzer and Meghan Shah and Mehmet Yatbaz and Melody Y. Guan and Mengyuan Xu and Mengyuan Yan and Mia Glaese and Mianna Chen and Michael Lampe and Michael Malek and Michele Wang and Michelle Fradin and Mike McClay and Mikhail Pavlov and Miles Wang and Mingxuan Wang and Mira Murati and Mo Bavarian and Mostafa Rohaninejad and Nat McAleese and Neil Chowdhury and Neil Chowdhury and Nick Ryder and Nikolas Tezak and Noam Brown and Ofir Nachum and Oleg Boiko and Oleg Murk and Olivia Watkins and Patrick Chao and Paul Ashbourne and Pavel Izmailov and Peter Zhokhov and Rachel Dias and Rahul Arora and Randall Lin and Rapha Gontijo Lopes and Raz Gaon and Reah Miyara and Reimar Leike and Renny Hwang and Rhythm Garg and Robin Brown and Roshan James and Rui Shu and Ryan Cheu and Ryan Greene and Saachi Jain and Sam Altman and Sam Toizer and Sam Toyer and Samuel Miserendino and Sandhini Agarwal and Santiago Hernandez and Sasha Baker and Scott McKinney and Scottie Yan and Shengjia Zhao and Shengli Hu and Shibani Santurkar and Shraman Ray Chaudhuri and Shuyuan Zhang and Siyuan Fu and Spencer Papay and Steph Lin and Suchir Balaji and Suvansh Sanjeev and Szymon Sidor and Tal Broda and Aidan Clark and Tao Wang and Taylor Gordon and Ted Sanders and Tejal Patwardhan and Thibault Sottiaux and Thomas Degry and Thomas Dimson and Tianhao Zheng and Timur Garipov and Tom Stasi and Trapit Bansal and Trevor Creech and Troy Peterson and Tyna Eloundou and Valerie Qi and Vineet Kosaraju and Vinnie Monaco and Vitchyr Pong and Vlad Fomenko and Weiyi Zheng and Wenda Zhou and Wes McCabe and Wojciech Zaremba and Yann Dubois and Yinghai Lu and Yining Chen and Young Cha and Yu Bai and Yuchen He and Yuchen Zhang and Yunyun Wang and Zheng Shao and Zhuohan Li},
      year={2024},
      eprint={2412.16720},
      archivePrefix={arXiv},
      primaryClass={cs.AI},
      url={https://arxiv.org/abs/2412.16720}, 
}

@article{Critical,
  author       = {Qing Zong and
                  Jiayu Liu and
                  Tianshi Zheng and
                  Chunyang Li and
                  Baixuan Xu and
                  Haochen Shi and
                  Weiqi Wang and
                  Zhaowei Wang and
                  Chunkit Chan and
                  Yangqiu Song},
  title        = {CritiCal: Can Critique Help {LLM} Uncertainty or Confidence Calibration?},
  journal      = {CoRR},
  volume       = {abs/2510.24505},
  year         = {2025},
  url          = {https://doi.org/10.48550/arXiv.2510.24505},
  doi          = {10.48550/ARXIV.2510.24505},
  eprinttype    = {arXiv},
  eprint       = {2510.24505},
  bibsource    = {dblp computer science bibliography, https://dblp.org}
}

@article{seed_thinking,
  title={Seed-thinking-v1. 5: Advancing superb reasoning models with reinforcement learning},
  author={Seed, ByteDance and Yuan, Yufeng and Yue, Yu and Wang, Mingxuan and Zuo, Xiaochen and Chen, Jiaze and Yan, Lin and Xu, Wenyuan and Zhang, Chi and Liu, Xin and others},
  journal={arXiv preprint arXiv:2504.13914},
  year={2025},
  url={https://arxiv.org/abs/2504.13914}
}

@techreport{openai2025o3,
  author      = {OpenAI},
  title       = {OpenAI o3 and o4-mini System Card},
  institution = {OpenAI},
  year        = {2025},
  month       = {April},
  url         = {https://cdn.openai.com/pdf/2221c875-02dc-4789-800b-e7758f3722c1/o3-and-o4-mini-system-card.pdf}
}

@inproceedings{Yang2025Qwen3TR,
  title={Qwen3 Technical Report},
  author={An Yang and Anfeng Li and Baosong Yang and Beichen Zhang and Binyuan Hui and Bo Zheng and Bowen Yu and Chang Gao and Chengen Huang and Chenxu Lv and Chujie Zheng and Dayiheng Liu and Fan Zhou and Fei Huang and Feng Hu and Hao Ge and Haoran Wei and Huan Lin and Jialong Tang and Jian Yang and Jianhong Tu and Jianwei Zhang and Jianxin Yang and Jiaxi Yang and Jing Zhou and Jingren Zhou and Junyang Lin and Kai Dang and Keqin Bao and Kexin Yang and Le Yu and Lianghao Deng and Mei Li and Min Xue and Mingze Li and Pei Zhang and Peng Wang and Qin Zhu and Rui Men and Ruize Gao and Shixuan Liu and Shuang Luo and Tianhao Li and Tianyi Tang and Wenbiao Yin and Xingzhang Ren and Xinyu Wang and Xinyu Zhang and Xuancheng Ren and Yang Fan and Yang Su and Yi-Chao Zhang and Yinger Zhang and Yu Wan and Yuqiong Liu and Zekun Wang and Zeyu Cui and Zhenru Zhang and Zhipeng Zhou and Zihan Qiu},
  year={2025},
  url={https://api.semanticscholar.org/CorpusID:278602855}
}

@techreport{anthropic2025claude,
    title={Claude 3.7 Sonnet System Card},
    author={Anthropic},
    year={2025},
    institution={Anthropic},
    url={https://www.anthropic.com/claude-3-7-sonnet-system-card}
}

@misc{deepmind2025gemini,
    title={Gemini 2.5 Pro Overview},
    author={{Google DeepMind}},
    year={2025},
    howpublished={\url{https://deepmind.google/technologies/gemini/pro/}}
}

@inproceedings{MarCon,
    title = "Revisiting Epistemic Markers in Confidence Estimation: Can Markers Accurately Reflect Large Language Models' Uncertainty?",
    author = "Liu, Jiayu  and
      Zong, Qing  and
      Wang, Weiqi  and
      Song, Yangqiu",
    editor = "Che, Wanxiang  and
      Nabende, Joyce  and
      Shutova, Ekaterina  and
      Pilehvar, Mohammad Taher",
    booktitle = "Proceedings of the 63rd Annual Meeting of the Association for Computational Linguistics (Volume 2: Short Papers)",
    month = jul,
    year = "2025",
    address = "Vienna, Austria",
    publisher = "Association for Computational Linguistics",
    url = "https://aclanthology.org/2025.acl-short.18/",
    doi = "10.18653/v1/2025.acl-short.18",
    pages = "206--221",
    ISBN = "979-8-89176-252-7"
}

@misc{MATP-Bench,
      title={MATP-BENCH: Can MLLM Be a Good Automated Theorem Prover for Multimodal Problems?}, 
      author={Zhitao He and Zongwei Lyu and Dazhong Chen and Dadi Guo and Yi R. Fung},
      year={2025},
      eprint={2506.06034},
      archivePrefix={arXiv},
      primaryClass={cs.CL},
      url={https://arxiv.org/abs/2506.06034}, 
}

@misc{kimi-k1.5,
      title={Kimi k1.5: Scaling Reinforcement Learning with LLMs}, 
      author={Kimi Team and Angang Du and Bofei Gao and Bowei Xing and Changjiu Jiang and Cheng Chen and Cheng Li and Chenjun Xiao and Chenzhuang Du and Chonghua Liao and Chuning Tang and Congcong Wang and Dehao Zhang and Enming Yuan and Enzhe Lu and Fengxiang Tang and Flood Sung and Guangda Wei and Guokun Lai and Haiqing Guo and Han Zhu and Hao Ding and Hao Hu and Hao Yang and Hao Zhang and Haotian Yao and Haotian Zhao and Haoyu Lu and Haoze Li and Haozhen Yu and Hongcheng Gao and Huabin Zheng and Huan Yuan and Jia Chen and Jianhang Guo and Jianlin Su and Jianzhou Wang and Jie Zhao and Jin Zhang and Jingyuan Liu and Junjie Yan and Junyan Wu and Lidong Shi and Ling Ye and Longhui Yu and Mengnan Dong and Neo Zhang and Ningchen Ma and Qiwei Pan and Qucheng Gong and Shaowei Liu and Shengling Ma and Shupeng Wei and Sihan Cao and Siying Huang and Tao Jiang and Weihao Gao and Weimin Xiong and Weiran He and Weixiao Huang and Wenhao Wu and Wenyang He and Xianghui Wei and Xianqing Jia and Xingzhe Wu and Xinran Xu and Xinxing Zu and Xinyu Zhou and Xuehai Pan and Y. Charles and Yang Li and Yangyang Hu and Yangyang Liu and Yanru Chen and Yejie Wang and Yibo Liu and Yidao Qin and Yifeng Liu and Ying Yang and Yiping Bao and Yulun Du and Yuxin Wu and Yuzhi Wang and Zaida Zhou and Zhaoji Wang and Zhaowei Li and Zhen Zhu and Zheng Zhang and Zhexu Wang and Zhilin Yang and Zhiqi Huang and Zihao Huang and Ziyao Xu and Zonghan Yang},
      year={2025},
      eprint={2501.12599},
      archivePrefix={arXiv},
      primaryClass={cs.AI},
      url={https://arxiv.org/abs/2501.12599}, 
}

@misc{survey-logical-reasoning-llm,
      title={Logical Reasoning in Large Language Models: A Survey}, 
      author={Hanmeng Liu and Zhizhang Fu and Mengru Ding and Ruoxi Ning and Chaoli Zhang and Xiaozhang Liu and Yue Zhang},
      year={2025},
      eprint={2502.09100},
      archivePrefix={arXiv},
      primaryClass={cs.AI},
      url={https://arxiv.org/abs/2502.09100}, 
}

@dataset{jia2025aime24,
  author       = {Maxwell Jia},
  title        = {AIME 2024 Dataset},
  year         = {2025},
  month        = {feb},
  howpublished = {\url{https://huggingface.co/datasets/Maxwell-Jia/AIME_2024}},
  version      = {8d88b28},
  note         = {American Invitational Mathematics Examination 2024, 30 items.}
}

@dataset{opencompass2025aime25,
  author       = {{OpenCompass Team}},
  title        = {AIME 2025 Dataset},
  year         = {2025},
  month        = {feb},
  howpublished = {\url{https://huggingface.co/datasets/opencompass/AIME2025}},
  version      = {a6ad95f},
  note         = {American Invitational Mathematics Examination 2025, 30 items.}
}

@inproceedings{Wang2025KiminaProverPT,
  title={Kimina-Prover Preview: Towards Large Formal Reasoning Models with Reinforcement Learning},
  author={Haiming Wang and Mert Unsal and Xiaohan Lin and Mantas Baksys and Junqi Liu and Marco Dos Santos and Flood Sung and Marina Vinyes and Zhenzhe Ying and Zekai Zhu and Jianqiao Lu and Hugues de Saxc'e and Bolton Bailey and Chendong Song and Chenjun Xiao and Dehao Zhang and Ebony Zhang and Frederick Pu and Han Zhu and Jiawei Liu and Jonas Bayer and Julien Michel and Long Yu and L{\'e}o Dreyfus-Schmidt and Lewis Tunstall and Luigi Pagani and Moreira Machado and Pauline Bourigault and Ran Wang and Stanislas Polu and Thibaut Barroyer and Wen-Ding Li and Yazhe Niu and Yann Fleureau and Yang Hu and Zhouliang Yu and Zihan Wang and Zhilin Yang and Zhengying Liu and Jia Li},
  year={2025},
  url={https://api.semanticscholar.org/CorpusID:277786851}
}

@misc{llm-math-failure,
      title={Large Language Models and Mathematical Reasoning Failures}, 
      author={Johan Boye and Birger Moell},
      year={2025},
      eprint={2502.11574},
      archivePrefix={arXiv},
      primaryClass={cs.AI},
      url={https://arxiv.org/abs/2502.11574}, 
}

@misc{ds-prover,
      title={DeepSeek-Prover-V2: Advancing Formal Mathematical Reasoning via Reinforcement Learning for Subgoal Decomposition}, 
      author={Z. Z. Ren and Zhihong Shao and Junxiao Song and Huajian Xin and Haocheng Wang and Wanjia Zhao and Liyue Zhang and Zhe Fu and Qihao Zhu and Dejian Yang and Z. F. Wu and Zhibin Gou and Shirong Ma and Hongxuan Tang and Yuxuan Liu and Wenjun Gao and Daya Guo and Chong Ruan},
      year={2025},
      eprint={2504.21801},
      archivePrefix={arXiv},
      primaryClass={cs.CL},
      url={https://arxiv.org/abs/2504.21801}, 
}

@inproceedings{miniF2F,
  author       = {Kunhao Zheng and
                  Jesse Michael Han and
                  Stanislas Polu},
  title        = {miniF2F: a cross-system benchmark for formal Olympiad-level mathematics},
  booktitle    = {The Tenth International Conference on Learning Representations, {ICLR}
                  2022, Virtual Event, April 25-29, 2022},
  publisher    = {OpenReview.net},
  year         = {2022},
  url          = {https://openreview.net/forum?id=9ZPegFuFTFv},
  bibsource    = {dblp computer science bibliography, https://dblp.org}
}

@article{ProofNet,
  author       = {Zhangir Azerbayev and
                  Bartosz Piotrowski and
                  Hailey Schoelkopf and
                  Edward W. Ayers and
                  Dragomir Radev and
                  Jeremy Avigad},
  title        = {ProofNet: Autoformalizing and Formally Proving Undergraduate-Level
                  Mathematics},
  journal      = {CoRR},
  volume       = {abs/2302.12433},
  year         = {2023},
  url          = {https://doi.org/10.48550/arXiv.2302.12433},
  doi          = {10.48550/ARXIV.2302.12433},
  eprinttype    = {arXiv},
  eprint       = {2302.12433},
  bibsource    = {dblp computer science bibliography, https://dblp.org}
}

@misc{USAMO-Proof-Bench,
      title={Proof or Bluff? Evaluating LLMs on 2025 USA Math Olympiad}, 
      author={Ivo Petrov and Jasper Dekoninck and Lyuben Baltadzhiev and Maria Drencheva and Kristian Minchev and Mislav Balunović and Nikola Jovanović and Martin Vechev},
      year={2025},
      eprint={2503.21934},
      archivePrefix={arXiv},
      primaryClass={cs.CL},
      url={https://arxiv.org/abs/2503.21934}, 
}

@inproceedings{MathQA-dataset,
  author       = {Aida Amini and
                  Saadia Gabriel and
                  Shanchuan Lin and
                  Rik Koncel{-}Kedziorski and
                  Yejin Choi and
                  Hannaneh Hajishirzi},
  editor       = {Jill Burstein and
                  Christy Doran and
                  Thamar Solorio},
  title        = {MathQA: Towards Interpretable Math Word Problem Solving with Operation-Based
                  Formalisms},
  booktitle    = {Proceedings of the 2019 Conference of the North American Chapter of
                  the Association for Computational Linguistics: Human Language Technologies,
                  {NAACL-HLT} 2019, Minneapolis, MN, USA, June 2-7, 2019, Volume 1 (Long
                  and Short Papers)},
  pages        = {2357--2367},
  publisher    = {Association for Computational Linguistics},
  year         = {2019},
  url          = {https://doi.org/10.18653/v1/n19-1245},
  doi          = {10.18653/V1/N19-1245},
  bibsource    = {dblp computer science bibliography, https://dblp.org}
}

@misc{Veri-R1,
      title={Veri-R1: Toward Precise and Faithful Claim Verification via Online Reinforcement Learning}, 
      author={Qi He and Cheng Qian and Xiusi Chen and Bingxiang He and Yi R. Fung and Heng Ji},
      year={2025},
      eprint={2510.01932},
      archivePrefix={arXiv},
      primaryClass={cs.CL},
      url={https://arxiv.org/abs/2510.01932}, 
}

@inproceedings{llm-retrieval-failure-mode,
  title={GProofT: A Multi-dimension Multi-round Fact Checking Framework Based on Claim Fact Extraction},
  author={Liu, Jiayu and Tang, Junhao and Wang, Hanwen and Xu, Baixuan and Shi, Haochen and Wang, Weiqi and Song, Yangqiu},
  booktitle={The Seventh Fact Extraction and VERification Workshop},
  pages={118},
  year={2024}
}

@article{FIMO-dataset,
  author       = {Chengwu Liu and
                  Jianhao Shen and
                  Huajian Xin and
                  Zhengying Liu and
                  Ye Yuan and
                  Haiming Wang and
                  Wei Ju and
                  Chuanyang Zheng and
                  Yichun Yin and
                  Lin Li and
                  Ming Zhang and
                  Qun Liu},
  title        = {{FIMO:} {A} Challenge Formal Dataset for Automated Theorem Proving},
  journal      = {CoRR},
  volume       = {abs/2309.04295},
  year         = {2023},
  url          = {https://doi.org/10.48550/arXiv.2309.04295},
  doi          = {10.48550/ARXIV.2309.04295},
  eprinttype    = {arXiv},
  eprint       = {2309.04295},
  bibsource    = {dblp computer science bibliography, https://dblp.org}
}

@inproceedings{NaturalProofs-dataset,
  author       = {Sean Welleck and
                  Jiacheng Liu and
                  Ronan Le Bras and
                  Hanna Hajishirzi and
                  Yejin Choi and
                  Kyunghyun Cho},
  editor       = {Joaquin Vanschoren and
                  Sai{-}Kit Yeung},
  title        = {NaturalProofs: Mathematical Theorem Proving in Natural Language},
  booktitle    = {Proceedings of the Neural Information Processing Systems Track on
                  Datasets and Benchmarks 1, NeurIPS Datasets and Benchmarks 2021, December
                  2021, virtual},
  year         = {2021},
  url          = {https://datasets-benchmarks-proceedings.neurips.cc/paper/2021/hash/d9d4f495e875a2e075a1a4a6e1b9770f-Abstract-round1.html},
  bibsource    = {dblp computer science bibliography, https://dblp.org}
}

@misc{LLM-and-math-failures,
      title={Large Language Models and Mathematical Reasoning Failures}, 
      author={Johan Boye and Birger Moell},
      year={2025},
      eprint={2502.11574},
      archivePrefix={arXiv},
      primaryClass={cs.AI},
      url={https://arxiv.org/abs/2502.11574}, 
}

@article{chervonyi2025gold,
  title={Gold-medalist Performance in Solving Olympiad Geometry with AlphaGeometry2},
  author={Chervonyi, Yuri and Trinh, Trieu H and Ol{\v{s}}{\'a}k, Miroslav and Yang, Xiaomeng and Nguyen, Hoang and Menegali, Marcelo and Jung, Junehyuk and Verma, Vikas and Le, Quoc V and Luong, Thang},
  journal={arXiv preprint arXiv:2502.03544},
  year={2025},
  url={https://arxiv.org/abs/2502.03544}
}

@misc{wang2025let,
      title={Let's Reason Formally: Natural-Formal Hybrid Reasoning Enhances LLM's Math Capability}, 
      author={Ruida Wang and Yuxin Li and Yi R. Fung and Tong Zhang},
      year={2025},
      eprint={2505.23703},
      archivePrefix={arXiv},
      primaryClass={cs.AI},
      url={https://arxiv.org/abs/2505.23703}, 
}

@article{fan2025sedareval,
  title={SedarEval: Automated Evaluation using Self-Adaptive Rubrics},
  author={Fan, Zhiyuan and Wang, Weinong and Wu, Xing and Zhang, Debing},
  journal={arXiv preprint arXiv:2501.15595},
  year={2025},
  url={https://arxiv.org/abs/2501.15595}
}

@misc{llm-as-a-judge-survey,
      title={A Survey on LLM-as-a-Judge}, 
      author={Jiawei Gu and Xuhui Jiang and Zhichao Shi and Hexiang Tan and Xuehao Zhai and Chengjin Xu and Wei Li and Yinghan Shen and Shengjie Ma and Honghao Liu and Saizhuo Wang and Kun Zhang and Yuanzhuo Wang and Wen Gao and Lionel Ni and Jian Guo},
      year={2025},
      eprint={2411.15594},
      archivePrefix={arXiv},
      primaryClass={cs.CL},
      url={https://arxiv.org/abs/2411.15594}, 
}

@misc{MarPT,
      title={Prospect Theory Fails for LLMs: Revealing Instability of Decision-Making under Epistemic Uncertainty}, 
      author={Rui Wang and Qihan Lin and Jiayu Liu and Qing Zong and Tianshi Zheng and Weiqi Wang and Yangqiu Song},
      year={2025},
      eprint={2508.08992},
      archivePrefix={arXiv},
      primaryClass={cs.AI},
      url={https://arxiv.org/abs/2508.08992}, 
}

@misc{xu2025mpbenchcomprehensivemultimodalreasoning,
      title={MPBench: A Comprehensive Multimodal Reasoning Benchmark for Process Errors Identification}, 
      author={Zhaopan Xu and Pengfei Zhou and Jiaxin Ai and Wangbo Zhao and Kai Wang and Xiaojiang Peng and Wenqi Shao and Hongxun Yao and Kaipeng Zhang},
      year={2025},
      eprint={2503.12505},
      archivePrefix={arXiv},
      primaryClass={cs.AI},
      url={https://arxiv.org/abs/2503.12505}, 
}

@article{1975Matthews,
  title={Matthews, B.: Comparison of the predicted and observed secondary structure of T4 phage lysozyme. Biochimica et Biophysica Acta (BBA)-Protein Structure 405(2), 442-451},
  author={ Matthews, B W },
  journal={Biochimica et Biophysica Acta},
  volume={405},
  number={2},
  pages={442-451},
  year={1975},
}

@online{gowers2025llmproofs,
  author       = {Gowers, Timothy},
  title        = {Why are {LLMs} not Better at Finding Proofs},
  date         = {2025-05-22},
  url          = {https://youtu.be/5D3x_Ygv3No},
  urldate      = {2025-06-20},
  publisher    = {YouTube},
  type         = {Video}
}

@article{Hu2024miniCTXNT,
  title={miniCTX: Neural Theorem Proving with (Long-)Contexts},
  author={Jiewen Hu and Thomas (Hanwen) Zhu and Sean Welleck},
  journal={ArXiv},
  year={2024},
  volume={abs/2408.03350},
  url={https://api.semanticscholar.org/CorpusID:271745051}
}

@inproceedings{Huang2025Gemini2P,
  title={Gemini 2.5 Pro Capable of Winning Gold at IMO 2025},
  author={Yichen Huang and Lin F. Yang},
  year={2025},
  url={https://api.semanticscholar.org/CorpusID:280045605}
}

@article{wang2025ma,
  title={MA-LoT: Model-Collaboration Lean-based Long Chain-of-Thought Reasoning enhances Formal Theorem Proving},
  author={Wang, Ruida and Pan, Rui and Li, Yuxin and Zhang, Jipeng and Jia, Yizhen and Diao, Shizhe and Pi, Renjie and Hu, Junjie and Zhang, Tong},
  journal={arXiv preprint arXiv:2503.03205},
  year={2025},
  url={https://arxiv.org/abs/2503.03205}
}

@article{zheng2024processbench,
  title={Processbench: Identifying process errors in mathematical reasoning},
  author={Zheng, Chujie and Zhang, Zhenru and Zhang, Beichen and Lin, Runji and Lu, Keming and Yu, Bowen and Liu, Dayiheng and Zhou, Jingren and Lin, Junyang},
  journal={arXiv preprint arXiv:2412.06559},
  year={2024},
  url={https://arxiv.org/abs/2412.06559}
}

@article{jiang2025pag,
  title={PAG: Multi-Turn Reinforced LLM Self-Correction with Policy as Generative Verifier},
  author={Jiang, Yuhua and Xiong, Yuwen and Yuan, Yufeng and Xin, Chao and Xu, Wenyuan and Yue, Yu and Zhao, Qianchuan and Yan, Lin},
  journal={arXiv preprint arXiv:2506.10406},
  year={2025},
  url={https://arxiv.org/abs/2506.10406}
}

@article{Dekoninck2025TheOP,
  title={The Open Proof Corpus: A Large-Scale Study of LLM-Generated Mathematical Proofs},
  author={Jasper Dekoninck and Ivo Petrov and Kristian Minchev and Mislav Balunovic and Martin T. Vechev and Miroslav Marinov and Maria Drencheva and Lyuba Konova and Milen Shumanov and Kaloyan Tsvetkov and Nikolay Drenchev and Lazar Todorov and Kalina Nikolova and Nikolay Georgiev and Vanesa Kalinkova and Margulan Ismoldayev},
  journal={ArXiv},
  year={2025},
  volume={abs/2506.21621},
  url={https://api.semanticscholar.org/CorpusID:280011115}
}

@article{onda2025leanconjecturer,
  title={LeanConjecturer: Automatic Generation of Mathematical Conjectures for Theorem Proving},
  author={Onda, Naoto and Kasaura, Kazumi and Oriike, Yuta and Taniguchi, Masaya and Sannai, Akiyoshi and Sonoda, Sho},
  journal={arXiv preprint arXiv:2506.22005},
  year={2025},
  url={https://arxiv.org/abs/2506.22005}
}

@article{zhang2025realmath,
  title={RealMath: A Continuous Benchmark for Evaluating Language Models on Research-Level Mathematics},
  author={Zhang, Jie and Petrui, Cezara and Nikoli{\'c}, Kristina and Tram{\`e}r, Florian},
  journal={arXiv preprint arXiv:2505.12575},
  year={2025},
  url={https://arxiv.org/abs/2505.12575}
}

@inproceedings{Wang2025DiversityEnhancedRF,
  title={Diversity-Enhanced Reasoning for Subjective Questions},
  author={Yumeng Wang and Zhiyuan Fan and Jiayu Liu and Yi R. Fung},
  year={2025},
  url={https://api.semanticscholar.org/CorpusID:280323285}
}

@article{Zhang2025DeepTheoremAL,
  title={DeepTheorem: Advancing LLM Reasoning for Theorem Proving Through Natural Language and Reinforcement Learning},
  author={Ziyin Zhang and Jiahao Xu and Zhiwei He and Tian Liang and Qiuzhi Liu and Yansi Li and Linfeng Song and Zhen-Pu Liang and Zhuosheng Zhang and Rui Wang and Zhaopeng Tu and Haitao Mi and Dong Yu},
  journal={ArXiv},
  year={2025},
  volume={abs/2505.23754},
  url={https://api.semanticscholar.org/CorpusID:278996747}
}

@article{Sheng2025SolvingIP,
  title={Solving Inequality Proofs with Large Language Models},
  author={Jiayi Sheng and Luna Lyu and Jikai Jin and Tony Xia and Alex Gu and James Zou and Pan Lu},
  journal={ArXiv},
  year={2025},
  volume={abs/2506.07927},
  url={https://api.semanticscholar.org/CorpusID:279250536}
}

@misc{openai2025gpt5,
  author       = {OpenAI},
  title        = {GPT-5 System Card},
  year         = {2025},
  url          = {https://cdn.openai.com/gpt-5-system-card.pdf},
  note         = {Accessed: 2025-10-07}
}

@article{Balunovic2025MathConstructCL,
  title={MathConstruct: Challenging LLM Reasoning with Constructive Proofs},
  author={Mislav Balunovi'c and Jasper Dekoninck and Nikola Jovanovi'c and Ivo Petrov and Martin T. Vechev},
  journal={ArXiv},
  year={2025},
  volume={abs/2502.10197},
  url={https://api.semanticscholar.org/CorpusID:276395130}
}

@article{Li2025OneES,
  title={One Example Shown, Many Concepts Known! Counterexample-Driven Conceptual Reasoning in Mathematical LLMs},
  author={Yinghui Li and Jiayi Kuang and Haojing Huang and Zhikun Xu and Xinnian Liang and Yi Yu and Wenlian Lu and Yangning Li and Xiaoyu Tan and Chao Qu and Ying Shen and Hai-Tao Zheng and Philip S. Yu},
  journal={ArXiv},
  year={2025},
  volume={abs/2502.10454},
  url={https://api.semanticscholar.org/CorpusID:276409210}
}

@article{Huang2025MATHPerturbBL,
  title={MATH-Perturb: Benchmarking LLMs' Math Reasoning Abilities against Hard Perturbations},
  author={Kaixuan Huang and Jiacheng Guo and Zihao Li and Xiang Ji and Jiawei Ge and Wenzhe Li and Yingqing Guo and Tianle Cai and Hui Yuan and Runzhe Wang and Yue Wu and Ming Yin and Shange Tang and Yangsibo Huang and Chi Jin and Xinyun Chen and Chiyuan Zhang and Mengdi Wang},
  journal={ArXiv},
  year={2025},
  volume={abs/2502.06453},
  url={https://api.semanticscholar.org/CorpusID:276249117}
}
